\documentclass[11pt,a4paper]{article}
\usepackage[hyperref]{acl2021}
\usepackage{times}
\usepackage{latexsym}

\usepackage{microtype}

\usepackage{graphicx}
\usepackage{tabularx} 
\usepackage{booktabs} 
\usepackage{mathtools}
\usepackage{siunitx}
\usepackage{amsmath}
\usepackage{amssymb}
\usepackage{euscript}
\usepackage{float}
\usepackage{caption}
\usepackage{subcaption}
\usepackage{multirow}
\usepackage{arydshln}
\usepackage{adjustbox}
\usepackage{tablefootnote}

\newcommand{\phillip}[1]{\textcolor{orange}{\bf\small [#1 --Phillip]}}

\newcommand\blfootnote[1]{%
  \begingroup
  \renewcommand\thefootnote{}\footnote{#1}%
  \addtocounter{footnote}{-1}%
  \endgroup
}

\aclfinalcopy
\def\aclpaperid{59}

\title{How Good is Your Tokenizer? \\ On the Monolingual Performance of Multilingual Language Models}

\author{
Phillip Rust\thanks{{ } Both authors contributed equally to this work.}$^{\,\,\,1}$\thanks{{ } PR is now affiliated with the University of Copenhagen.}$^{\,\,\,}$,  Jonas Pfeiffer$^{*1}$, \\
{\bf Ivan Vuli\'{c}$^{2}$,  Sebastian Ruder$^{3}$,  Iryna Gurevych$^{1}$ } \\
$^1$Ubiquitous Knowledge Processing Lab, 
  Technical University of Darmstadt \\
$^2$Language Technology Lab, University of Cambridge \hspace{0.5em} \\
$^3$DeepMind \\
{\url{www.ukp.tu-darmstadt.de}}\\
}

\date{}

\begin{document}
\maketitle
\begin{abstract}
In this work, we provide a \textit{systematic and comprehensive empirical comparison} of pretrained multilingual language models versus their monolingual counterparts with regard to their monolingual task performance. We study a set of nine typologically diverse languages with readily available pretrained monolingual models on a set of five diverse monolingual downstream tasks.
We first aim to establish, via fair and controlled comparisons, if a gap between the multilingual and the corresponding monolingual representation model of that language exists, and subsequently investigate the reason for any performance difference. To disentangle conflating factors, we train new monolingual models on the same data, with monolingually and multilingually trained tokenizers.
We find that while the pretraining data size is an important factor, a designated monolingual tokenizer plays an equally important role in the downstream performance. Our results show that languages that are adequately represented in the multilingual model's vocabulary exhibit negligible performance decreases over their monolingual counterparts. We further find that replacing the original multilingual tokenizer with the specialized monolingual tokenizer improves the downstream performance of the multilingual model for almost every task and language. \blfootnote{\\Our code is available at \href{https://github.com/Adapter-Hub/hgiyt}{https://github.com/Adapter-Hub/hgiyt}.}
\end{abstract}

\section{Introduction}

Following large transformer-based language models \cite[LMs,][]{Vaswani:2017} pretrained on large English corpora  \cite[e.g., BERT, RoBERTa, T5;][]{devlin:2019,liu:2019,Raffel:2020t5}, similar monolingual language models have been introduced for other languages \cite[\textit{inter alia}]{virtanen:2019, antoun:2020, martin:2020}, offering previously unmatched performance in all NLP tasks. Concurrently, massively multilingual models with the same architectures and training procedures, covering more than 100 languages, have been proposed  \cite[e.g., mBERT, XLM-R, mT5;][]{devlin:2019, conneau:2020, xue2020mt5}. 

The ``industry'' of pretraining and releasing new monolingual BERT models continues its operations despite the fact that the corresponding languages are already covered by multilingual models.
The common argument justifying the need for monolingual variants is the assumption that multilingual models---due to suffering from the so-called curse of multilinguality \cite[i.e., the lack of capacity to represent all languages in an equitable way]{conneau:2020}---underperform monolingual models when applied to monolingual tasks \cite[\textit{inter alia}]{virtanen:2019,antoun:2020,ronnqvist:2019}. However, little to no compelling empirical evidence with rigorous experiments and fair comparisons have been presented so far to support or invalidate this strong claim. In this regard, much of the work proposing and releasing new monolingual models is grounded in anecdotal evidence, pointing to the positive results reported for other monolingual BERT models \cite{vries:2019, virtanen:2019, antoun:2020}.

Monolingual BERT models are typically evaluated on downstream NLP tasks to demonstrate their effectiveness in comparison to previous monolingual models or mBERT \cite[\textit{inter alia}]{virtanen:2019, antoun:2020, martin:2020}. While these results do show that \textit{certain} monolingual models \textit{can} outperform mBERT in \textit{certain} tasks, we hypothesize that this may substantially vary across different languages and language properties, tasks, pretrained models and their pretraining data, domain, and size. We further argue that conclusive evidence, either supporting or refuting the key hypothesis that monolingual models currently outperform multilingual models, necessitates an independent and controlled empirical comparison on a diverse set of languages and tasks.

While recent work has argued and validated that mBERT is under-trained \cite{ronnqvist:2019, wu-dredze-2020-languages}, providing evidence of improved performance when training monolingual models on more data, it is unclear if this is the only factor relevant for the performance of monolingual models. 
Another so far under-studied factor is  the limited vocabulary size of multilingual models compared to the sum of tokens of all corresponding monolingual models. 
Our analyses  investigating dedicated (i.e., language-specific)  tokenizers reveal the importance of high-quality tokenizers for the performance of both model variants. We also shed light on the interplay of tokenization with other factors such as pretraining data size.

\vspace{1.6mm}
\noindent \textbf{Contributions.} \textbf{1)} We systematically compare monolingual with multilingual pretrained language models for 9 typologically diverse languages on 5 structurally different tasks. \textbf{2)} We train new monolingual models on equally sized datasets with different tokenizers (i.e., shared multilingual versus dedicated language-specific tokenizers) to disentangle the impact of pretraining data size from the vocabulary of the tokenizer. \textbf{3)} We isolate factors that contribute to a performance difference (e.g., tokenizers' ``fertility'', the number of unseen (sub)words, data size) and provide an in-depth analysis of the impact of these  factors on task performance. \textbf{4)} Our results suggest that monolingually adapted tokenizers can robustly improve monolingual performance of multilingual models.

\section{Background and Related Work}

\noindent \textbf{Multilingual LMs.}
The widespread usage of pretrained multilingual Transformer-based LMs has been instigated by the release of multilingual BERT \cite{devlin:2019}, which followed on the success of the monolingual English BERT model. 
mBERT adopted the same pretraining regime as monolingual BERT by concatenating the 104 largest Wikipedias. Exponential smoothing was used when creating the subword vocabulary based on WordPieces \cite{Wu:2016wp} and a pretraining corpus. By oversampling underrepresented languages and undersampling overrepresented ones, it aims  to counteract the imbalance of pretraining data sizes. The final shared mBERT vocabulary comprises a total of 119,547 subword tokens.

Other multilingual models followed mBERT, such as XLM-R \cite{conneau:2020}. Concurrently, many studies  analyzed mBERT's and XLM-R's capabilities and limitations, finding that the multilingual models work surprisingly well for cross-lingual tasks, despite the fact that they do not rely on  direct cross-lingual supervision  \cite[e.g., parallel or comparable data, translation dictionaries;][]{pires:2019, wu-dredze-2019-beto,artetxe-etal-2020-cross,hu:2020, k:2020}.

However, recent work has also pointed to some fundamental limitations of  multilingual LMs. \citet{conneau:2020} observe that, for a fixed model capacity, adding new languages increases cross-lingual performance up to a certain point, after which adding more languages results in  performance drops. This phenomenon, termed the \textit{curse of multilinguality}, can be attenuated by increasing the model capacity \cite{artetxe-etal-2020-cross,pfeiffer:2020b,ChauLS20Parsing} or through additional training for particular language pairs \cite{pfeiffer:2020b,ponti-etal-2020-xcopa}. Another observation concerns substantially reduced cross-lingual and monolingual abilities of the models for resource-poor languages with smaller pretraining data \cite{wu-dredze-2020-languages,hu:2020,Lauscher:2020zerohero}. Those languages remain underrepresented in the subword vocabulary and the model's shared representation space despite oversampling. 
Despite recent efforts to mitigate this issue (e.g., \citet{ChungGTR20} propose to cluster and merge the vocabularies of similar languages, before defining a joint vocabulary across all languages), the multilingual LMs still struggle with balancing their parameters across many languages.

\vspace{1.6mm}
\noindent \textbf{Monolingual versus Multilingual LMs.}
\label{sec:mono_vs_multi}
New monolingual language-specific models also emerged for many languages, following BERT's architecture and pretraining procedure. There are monolingual BERT variants for Arabic \cite{antoun:2020}, French \cite{martin:2020}, Finnish \cite{virtanen:2019}, Dutch \cite{vries:2019}, to name only a few. \citet{pyysalo2020wikibert} released 44 monolingual WikiBERT models trained on Wikipedia. However, only a few studies have thus far, either explicitly or implicitly, attempted to understand how monolingual and multilingual LMs compare across languages. \\

\citet{nozza2020mask} extracted task results from the respective papers on monolingual BERTs to facilitate an overview of monolingual models and their comparison to mBERT.\footnote{\href{https://bertlang.unibocconi.it/}{https://bertlang.unibocconi.it/}}
However, they have not verified the scores, nor have they performed a controlled impartial comparison.

\citet{vulic-etal-2020-probing} probed mBERT and monolingual BERT models across six typologically diverse languages for lexical semantics. 
They show that pretrained monolingual BERT models encode significantly more lexical information than mBERT.

\citet{zhang2020need} investigated the role of pretraining data size with RoBERTa, finding that the model  learns most syntactic and semantic features on corpora spanning 10M--100M word tokens, but still requires massive datasets to learn higher-level semantic and commonsense knowledge. 

\citet{mulcaire-etal-2019-polyglot} compared monolingual and bilingual ELMo \cite{peters:2018} LMs across three downstream tasks, finding that contextualized representations from the bilingual models can improve monolingual task performance relative to their monolingual counterparts.\footnote{\citet{mulcaire-etal-2019-polyglot} clearly differentiate between \emph{multilingual} and \emph{polyglot} models. Their definition of polyglot models is in line with what we term multilingual models.} However, it is unclear how their findings extend to \emph{massively} multilingual LMs potentially suffering from the curse of multilinguality. 

\citet{ronnqvist:2019} compared mBERT to monolingual BERT models for six languages (German, English, Swedish, Danish, Norwegian, Finnish) on three different tasks. 
They find that mBERT lags behind its monolingual counterparts in terms of performance on
cloze and generation tasks. 
They also identified clear differences among the six languages in terms of this performance gap. 
They speculate that mBERT is under-trained with respect to individual languages. 
However, their set of tasks is limited, and their language sample is typologically narrow; it remains unclear whether these findings extend to different language families and to structurally different tasks. 

Despite recent efforts, a careful, systematic study within a \textit{controlled} experimental setup, a diverse language sample and set of tasks is still lacking. We aim to address this gap in this work.

\section{Controlled Experimental Setup}
\label{sec:experimental_setup}

We compare multilingual BERT with its monolingual counterparts in a spectrum of typologically diverse languages and across a variety of downstream tasks. By isolating and analyzing crucial factors contributing to downstream performance, such as  tokenizers and pretraining data, we can conduct unbiased and fair comparisons.

\subsection{Language and Task Selection}
\label{sec:tasks}

Our selection of languages has been guided by several (sometimes competing) criteria: \textbf{C1)} typological diversity; \textbf{C2)} availability of pretrained monolingual BERT models; \textbf{C3)} representation of the languages in standard evaluation benchmarks for a sufficient number of tasks. 

Regarding C1, most high-resource languages belong to the same language families, thus sharing a majority of their linguistic features. Neglecting typological diversity inevitably leads to poor generalizability and  language-specific biases \cite{gerz-etal-2018-relation, ponti-etal-2019-modeling, joshi-etal-2020-state}. Following recent work in multilingual NLP that pays particular attention to typological diversity \cite[\textit{inter alia}]{clark:2020,hu:2020, ponti-etal-2020-xcopa}, we experiment with a language sample covering a broad spectrum of language properties.

Regarding C2, for computational tractability, we only select languages with readily available BERT models. Unlike prior work, which typically lacks either language \cite{ronnqvist:2019, zhang2020need} or task diversity \cite{wu-dredze-2020-languages, vulic-etal-2020-probing}, we ensure that our experimental framework takes both into account, thus also satisfying C3. We achieve task diversity and generalizability by selecting a combination of tasks driven by lower-level syntactic and higher-level semantic features \cite{Lauscher:2020zerohero}.

Finally, we select a set of 9 languages from 8 language families, as listed in Table~\ref{tab:language_selection}.\footnote{Note that, since we evaluate monolingual performance and not cross-lingual transfer performance, we require \textit{training data} in the target language. Therefore, we are unable to leverage many of the available multilingual evaluation data such as XQuAD \cite{artetxe-etal-2020-cross}, MLQA \cite{lewis-etal-2020-mlqa}, or XNLI \cite{conneau-etal-2018-xnli}. These evaluation sets do not provide any training portions for languages other than English. Additional information regarding our selection of pretrained models is available in Appendix~\ref{a:model_selection}.}  We evaluate mBERT and monolingual BERT models on five downstream NLP tasks: named entity recognition (NER), sentiment analysis (SA), question answering (QA), universal dependency parsing (UDP), and part-of-speech tagging (POS).\footnote{Information on which datasets are associated with which language and the dataset sizes (examples per split) are provided in Appendix~\ref{a:data_preprocessing}.}

\begin{table}[H]
\centering
\resizebox{0.49\textwidth}{!}{%
\begin{tabular}{l ll l}
\toprule
\textbf{Language} & \textbf{ISO} & \textbf{Language Family} & \textbf{Pretrained BERT Model} \\ 
\midrule
Arabic & \textsc{ar} & Afroasiatic & AraBERT \cite{antoun:2020} \\ 
English & \textsc{en} & Indo-European   & BERT \cite{devlin:2019} \\ 
Finnish & \textsc{fi} & Uralic & FinBERT \cite{virtanen:2019} \\ 
Indonesian & \textsc{id} & Austronesian & IndoBERT \cite{wilie-etal-2020-indonlu} \\
Japanese & \textsc{ja} & Japonic & Japanese-char BERT\tablefootnote{\label{fn:jbert}\href{https://github.com/cl-tohoku/bert-japanese}{https://github.com/cl-tohoku/bert-japanese}} \\ 
Korean & \textsc{ko} & Koreanic & KR-BERT \cite{lee2020krbert} \\ 
Russian & \textsc{ru} & Indo-European   & RuBERT \cite{kuratov:2019} \\ 
Turkish & \textsc{tr} & Turkic & BERTurk \cite{schweter:2020} \\
Chinese & \textsc{zh} & Sino-Tibetan & Chinese BERT \cite{devlin:2019} \\
\bottomrule
\end{tabular}
}%
\caption{Overview of selected languages and their respective pretrained monolingual BERT models.}
\label{tab:language_selection}
\end{table}

\vspace{1.6mm}
\noindent \textbf{Named Entity Recognition (NER).} We rely on: CoNLL-2003 \cite{sang:2003}, FiNER \cite{ruokolainen:2019}, Chinese Literature \cite{xu:2017}, KMOU NER,\footnote{\label{fn:kmou}\href{https://github.com/kmounlp/NER}{https://github.com/kmounlp/NER}} WikiAnn \cite{pan:2017, rahimi:2019}.

\vspace{1mm}
\noindent \textbf{Sentiment Analysis (SA).}
We employ: HARD \cite{elnagar:2018}, IMDb Movie Reviews \cite{maas:2011}, Indonesian Prosa \cite{purwarianti:2019}, Yahoo Movie Reviews,\footnote{\label{fn:yahoo} \href{https://github.com/dennybritz/sentiment-analysis/tree/master/data}{https://github.com/dennybritz/sentiment-analysis}} NSMC,\footnote{\label{note:nsmc} \href{https://www.lucypark.kr/docs/2015-pyconkr/\#39}{https://www.lucypark.kr/docs/2015-pyconkr/\#39}} RuReviews \cite{smetanin:2019}, Turkish Movie and Product Reviews \cite{demirtas:2013}, ChnSentiCorp.\footnote{\label{note:chnsenticorp}\href{https://github.com/pengming617/bert_classification/tree/master/data}{https://github.com/pengming617/bert\_classification}}

\vspace{1mm}
\noindent \textbf{Question Answering (QA).} 
We use: SQuADv1.1 \cite{rajpurkar:2016}, KorQuAD 1.0 \cite{lim:2019}, SberQuAD \cite{efimov:2020}, TQuAD,\footnote{\label{fn:tquad}\href{https://tquad.github.io/turkish-nlp-qa-dataset/}{https://tquad.github.io/turkish-nlp-qa-dataset/}} DRCD \cite{shao:2019}, TyDiQA-GoldP \cite{clark:2020}. 

\vspace{1mm}
\noindent \textbf{Dependency Parsing (UDP).} 
We rely on Universal Dependencies \cite{nivre:2016, nivre:2020} v2.6 \cite{zeman:2020} for all languages.

\vspace{1mm}
\noindent \textbf{Part-of-Speech Tagging (POS).} 
We again utilize Universal Dependencies v2.6.

\subsection{Task-Based Fine-Tuning}

\textbf{Fine-Tuning Setup.} For all tasks besides UDP, we use the standard fine-tuning setup of \citet{devlin:2019}. For UDP, we use a transformer-based variant \cite{glavas:2020} of the standard deep biaffine attention dependency parser \cite{dozat:2017}. 
We distinguish between  fully fine-tuning a monolingual BERT model and fully fine-tuning mBERT on the task.
For both settings, we average scores over three random initializations on the development set. On the test set, we report the results of the initialization that achieved the highest score on the development set.

\vspace{1.6mm}
\noindent \textbf{Evaluation Measures.} We report $F_1$ scores for NER, accuracy scores for SA and POS, unlabeled and labeled attachment scores (UAS \& LAS) for UDP, and exact match and $F_1$ scores for QA.

\vspace{1.6mm}
\noindent\textbf{Hyper-Parameters and Technical Details.} We use AdamW \cite{kingma:2015} in all experiments, with a learning rate of $3e-5$.\footnote{ Preliminary experiments indicated this to be a well performing learning rate.
 Due to the large volume of our experiments, we were unable to tune all the hyper-parameters for each setting. We found that a higher learning rate of $5e-4$ works best for adapter-based fine-tuning (see later) since the task adapter parameters are learned from scratch (i.e., they are randomly initialized).}
We  train for 10 epochs with early stopping \cite{prechelt:1998}.\footnote{We evaluate a model every 500 gradient steps on the development set, saving the best-performing model based on the respective evaluation measures. We terminate training if no performance gains are observed within five consecutive evaluation runs ($=$ 2,500 steps). For QA and UDP, we use the $F_1$ scores and LAS, respectively. For \textsc{FI} and \textsc{ID} QA, we train for 20 epochs due to slower convergence. We train with batch size 32 and max sequence length 256 for all tasks except QA. In QA, the batch size is 24, max sequence length 384, query length 64, and document stride is set to 128.}

\begin{table}[!t]
\centering
\resizebox{0.48\textwidth}{!}{%
\begin{tabular}{@{}llccccccccc@{}}
\toprule
\multirow{3}{*}{\textbf{Lg}} & \multirow{3}{*}{\textbf{Model}} & \multicolumn{1}{l}{\textbf{NER}} &  & \textbf{SA} &  & \textbf{QA} &  & \textbf{UDP} &  & \textbf{POS} \\
& & Test &  & Test &  & Dev &  & Test &  & Test \\
& & $F_1$ &  & Acc &  & EM / $F_1$ &  & UAS / LAS &  & Acc \\ \midrule
\multirow{2}{*}{\textsc{ar}} & Monolingual & \textbf{91.1} &  & \textbf{95.9} &  & \textbf{68.3} / \textbf{82.4} &  & \textbf{90.1} / \textbf{85.6} &  & \textbf{96.8} \\
 & mBERT & 90.0 &  & 95.4 &  & 66.1 / 80.6 &  & 88.8 / 83.8 &  & \textbf{96.8} \\
\midrule
\multirow{2}{*}{\textsc{en}} & Monolingual & \textbf{91.5} &  & \textbf{91.6} &  & 80.5 / 88.0 &  & \textbf{92.1} / \textbf{89.7} &  & \textbf{97.0} \\
 & mBERT & 91.2 &  & 89.8 &  & \textbf{80.9} / \textbf{88.4} &  & 91.6 / 89.1 &  & 96.9 \\
\midrule
\multirow{2}{*}{\textsc{fi}} & Monolingual & \textbf{92.0} &  & ----- &  & \textbf{69.9} / \textbf{81.6} &  & \textbf{95.9} / \textbf{94.4} &  & \textbf{98.4} \\
 & mBERT & 88.2 &  & ----- &  & 66.6 / 77.6 &  & 91.9 / 88.7 &  & 96.2 \\
\midrule
\multirow{2}{*}{\textsc{id}} & Monolingual & 91.0 &  & \textbf{96.0} &  & 66.8 / 78.1 &  & 85.3 / 78.1 &  & 92.1 \\
 & mBERT & \textbf{93.5} &  & 91.4 &  & \textbf{71.2} / \textbf{82.1} &  & \textbf{85.9} / \textbf{79.3} &  & \textbf{93.5} \\
\midrule
\multirow{2}{*}{\textsc{ja}} & Monolingual & 72.4 &  & \textbf{88.0} &  & ----- / ----- &  & \textbf{94.7} / \textbf{93.0} &  & \textbf{98.1} \\
 & mBERT & \textbf{73.4} &  & 87.8 &  & ----- / ----- &  & 94.0 / 92.3 &  & 97.8 \\
\midrule
\multirow{2}{*}{\textsc{ko}} & Monolingual & \textbf{88.8} &  & \textbf{89.7} &  & \textbf{74.2} / \textbf{91.1} &  & \textbf{90.3} / \textbf{87.2} &  & \textbf{97.0} \\
 & mBERT & 86.6 &  & 86.7 &  & 69.7 / 89.5 &  & 89.2 / 85.7 &  & 96.0 \\
\midrule
\multirow{2}{*}{\textsc{ru}} & Monolingual & \textbf{91.0} &  & \textbf{95.2} &  & \textbf{64.3} / \textbf{83.7} &  & \textbf{93.1} / \textbf{89.9} &  & \textbf{98.4} \\
 & mBERT & 90.0 &  & 95.0 &  & 63.3 / 82.6 &  & 91.9 / 88.5 &  & 98.2 \\
\midrule
\multirow{2}{*}{\textsc{tr}} & Monolingual & 92.8 &  & \textbf{88.8} &  & \textbf{60.6} / \textbf{78.1} &  & \textbf{79.8} / \textbf{73.2} &  & \textbf{96.9} \\
 & mBERT & \textbf{93.8} &  & 86.4 &  & 57.9 / 76.4 &  & 74.5 / 67.4 &  & 95.7 \\
\midrule
\multirow{2}{*}{\textsc{zh}} & Monolingual & \textbf{76.5} &  & \textbf{95.3} &  & \textbf{82.3} / \textbf{89.3} &  & \textbf{88.6} / \textbf{85.6} &  & \textbf{97.2} \\
 & mBERT & 76.1 &  & 93.8 &  & 82.0 / \textbf{89.3} &  & 88.1 / 85.0 &  & 96.7 \\ \midrule \midrule
\multirow{2}{*}{\textsc{avg}} & Monolingual & \textbf{87.4} & & \textbf{92.4} & & \textbf{70.8} / \textbf{84.0} & & \textbf{90.0} / \textbf{86.3} & & \textbf{96.9} \\
 & mBERT & 87.0 & & 91.0 & & 69.7 / 83.3 & & 88.4 / 84.4 & & 96.4 \\
\bottomrule
\end{tabular}%
}
\caption{Performance on Named Entity Recognition (NER), Sentiment Analysis (SA), Question Answering (QA), Universal Dependency Parsing (UDP), and Part-of-Speech Tagging (POS). We use development (dev) sets only for QA. Finnish (\textsc{fi}) SA and Japanese (\textsc{ja}) QA lack respective datasets.}
\label{tab:results_table_test}
\end{table}

\subsection{Initial Results}
\label{sec:results}
We report our first set of results in Table~\ref{tab:results_table_test}.\footnote{See Appendix Table~\ref{tab:full_results} for the results on development sets.} We find that the performance gap between monolingual models and mBERT does exist to a large extent, confirming anecdotal evidence from prior work. However, we also notice that the score differences are largely dependent on the language and task at hand. The largest performance gains of monolingual models over mBERT are found for \textsc{fi}, \textsc{tr}, \textsc{ko}, and \textsc{ar}. In contrast, mBERT outperforms the IndoBERT (\textsc{id}) model in all tasks except SA, and performs competitively with the \textsc{ja} and \textsc{zh} monolingual models on most datasets. In general, the gap is particularly narrow for POS tagging, where all models tend to score high (in most cases north of 95\% accuracy). \textsc{id} aside, we also see a clear trend for UDP, with monolingual models outperforming fully fine-tuned mBERT models, most notably for \textsc{fi} and \textsc{tr}.
In what follows, we seek to understand the causes of this behavior in relation to different factors such as  tokenizers, corpora sizes, as well as languages and tasks in consideration.

\begin{figure*}[!t]
    \centering
    \begin{subfigure}[b]{0.328\textwidth}
        \centering
        \includegraphics[width=\textwidth]{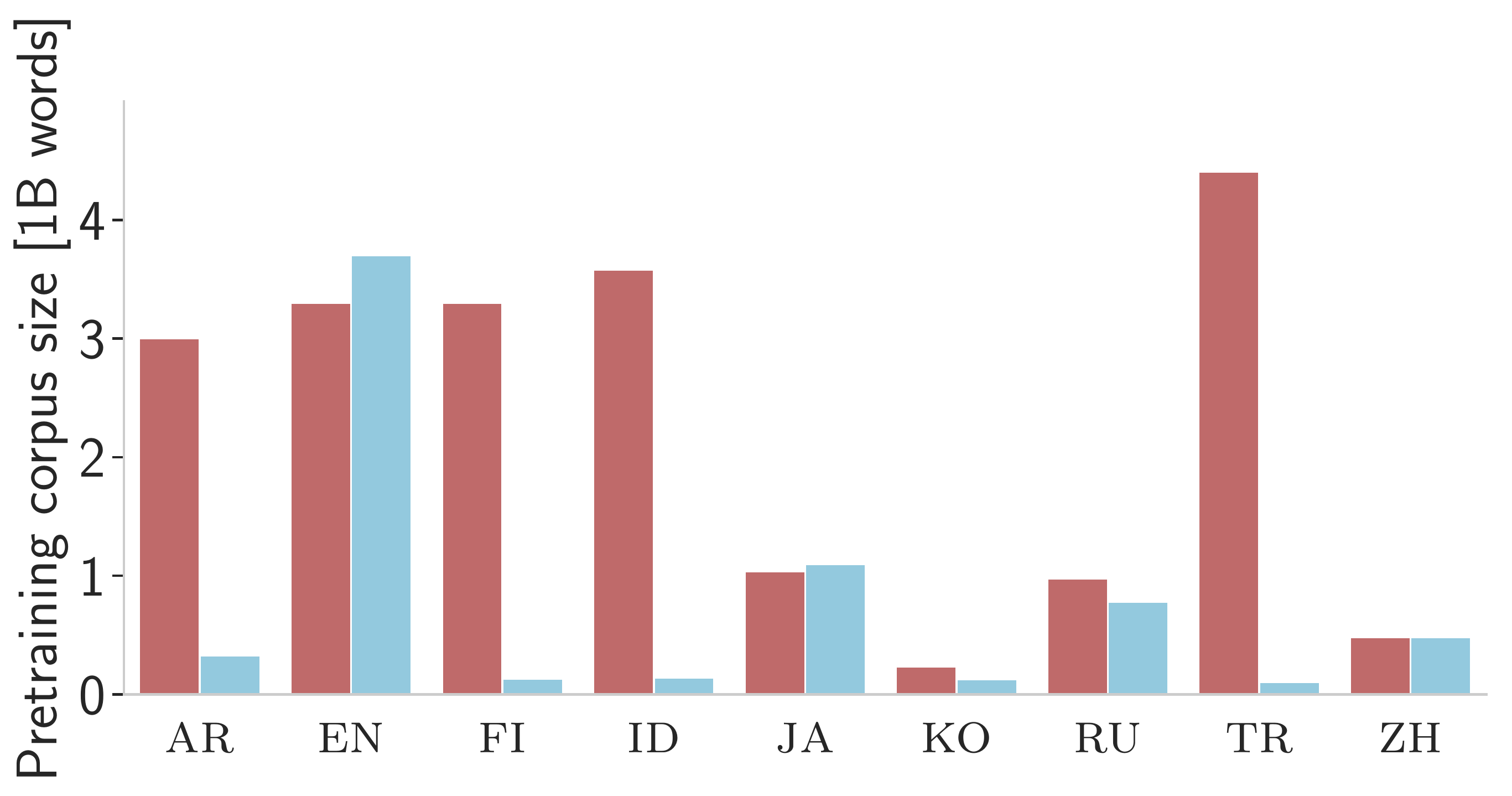}
        \caption{Pretraining corpus size}
        \label{fig:size_comparison}
    \end{subfigure}
    \hspace{-1em}
    \begin{subfigure}[b]{0.328\textwidth}
        \centering
        \includegraphics[width=\textwidth]{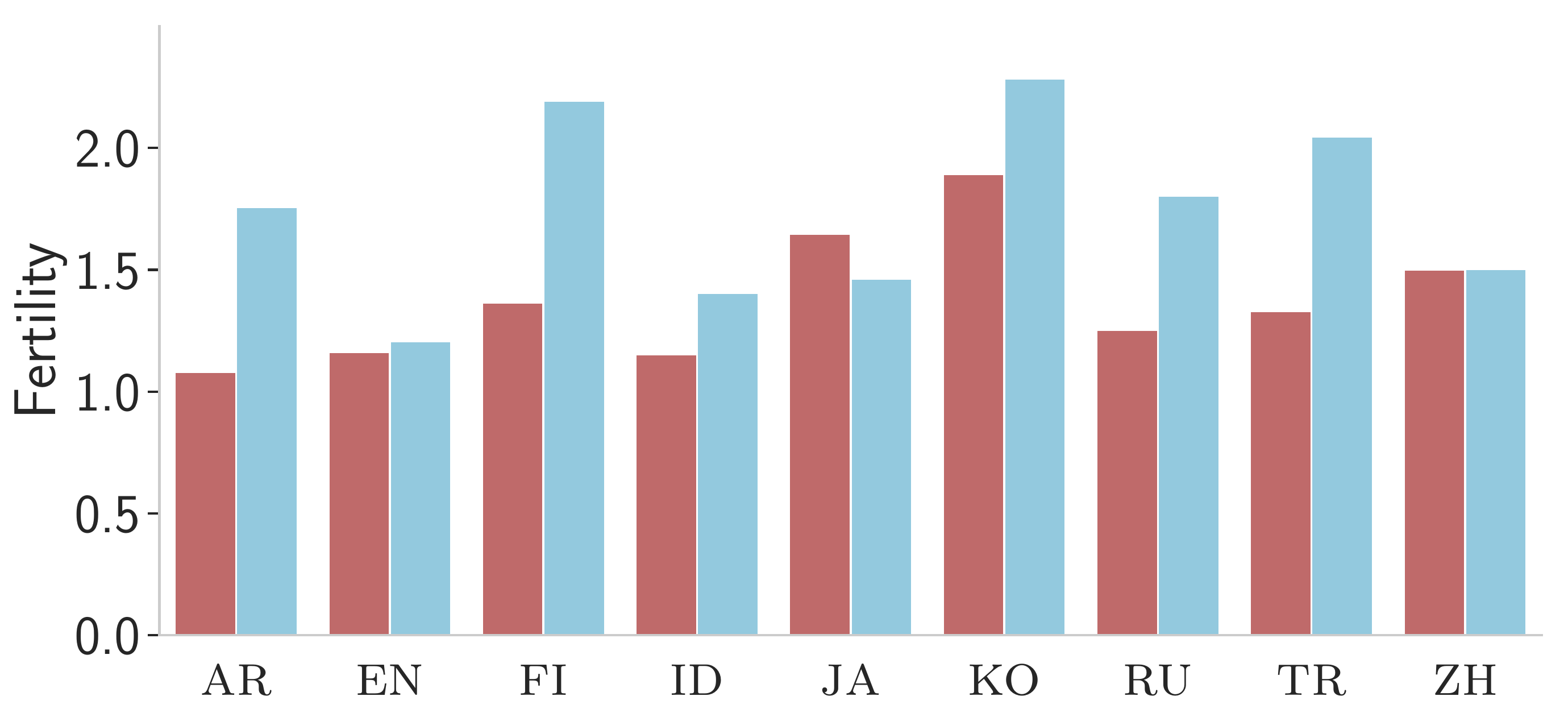}
        \caption{Subword fertility}
        \label{fig:fertility}
    \end{subfigure}
    \hspace{-0.5em}
    \begin{subfigure}[b]{0.328\textwidth}
        \centering
        \includegraphics[width=\textwidth]{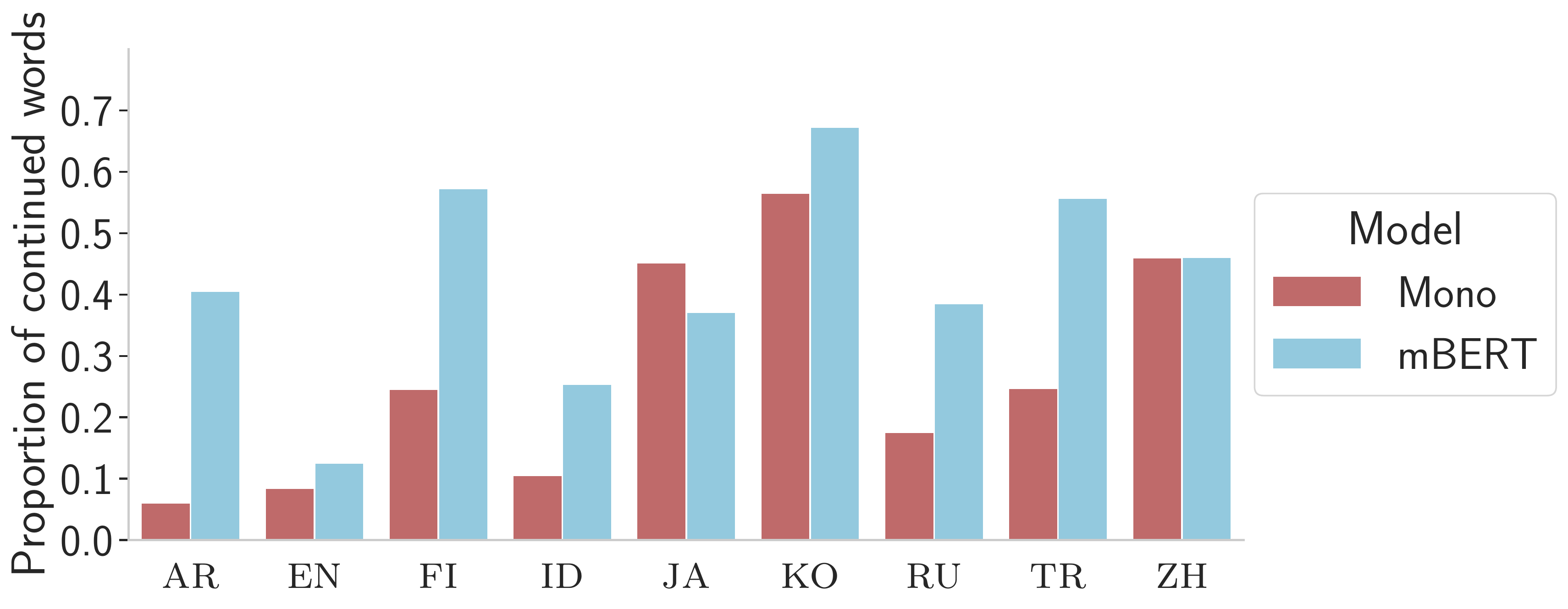}
        \caption{Proportion of continued words}
        \label{fig:continuation}
    \end{subfigure}
    \caption{Comparison of monolingual models with mBERT w.r.t. pretraining corpus size (measured in billions of words), subword fertility (i.e., the average number of subword tokens produced per tokenized word \cite{acs:2019}), and proportion of continued words (i.e., words split into multiple subword tokens \cite{acs:2019}).}
    \label{fig:metrics}
\end{figure*}

\section{Tokenizer versus Corpus Size}
\label{sec:analysis}

\subsection{Pretraining Corpus Size}

The size of the pretraining corpora plays an important role in the performance of transformers \cite[\textit{inter alia}]{liu:2019, conneau:2020, zhang2020need}. Therefore, we compare how much data each monolingual model was trained on with the amount of data in the respective language that mBERT has seen during training. Given that mBERT was trained on entire Wikipedia dumps, we estimate the latter by the total number of words across all articles listed for each Wiki.\footnote{Based on the numbers from \\ \href{https://meta.m.wikimedia.org/wiki/List\_of\_Wikipedias}{https://meta.m.wikimedia.org/wiki/List\_of\_Wikipedias}} For the monolingual LMs, we extract information on pretraining data from the model documentation. If no exact numbers are explicitly stated, and the pretraining corpora are unavailable, we make estimations based on the information provided by the authors.\footnote{We provide further details in Appendix~\ref{a:data_size_estimation}.} The statistics are provided in~Figure \ref{fig:size_comparison}.
For \textsc{en}, \textsc{ja}, \textsc{ru}, and \textsc{zh}, both the respective monolingual BERT and mBERT were trained on similar amounts of monolingual data. On the other hand, monolingual BERTs of \textsc{ar}, \textsc{id}, \textsc{fi}, \textsc{ko}, and \textsc{tr} were trained on  about twice (\textsc{ko}) up to more than 40 times (\textsc{tr}) as much data in their language than mBERT.

\subsection{Tokenizer} 
\label{s:tokenizer_analysis}

Compared to monolingual models, mBERT is substantially more limited in terms of the parameter budget that it can allocate for each of its 104 languages in its vocabulary. In addition, monolingual tokenizers are typically trained by native-speaking experts who are aware of relevant linguistic phenomena exhibited by their target language. We thus inspect how this affects the tokenizations of monolingual data produced by our sample of monolingual models and mBERT. We tokenize examples from Universal Dependencies v2.6 treebanks  and compute two metrics \cite{acs:2019}.\footnote{We provide further details  in Appendix~\ref{a:ud_tokenizer_data}.} First, the subword \textit{fertility} measures the average number of subwords produced per tokenized word. A minimum fertility of 1 means that the tokenizer's vocabulary contains every single word in the text. We plot the fertility scores in Figure~\ref{fig:fertility}.
We find that mBERT has similar fertility values as its monolingual counterparts for \textsc{en}, \textsc{id}, \textsc{ja}, and \textsc{zh}. In contrast, mBERT has a much higher fertility for \textsc{ar}, \textsc{fi}, \textsc{ko}, \textsc{ru}, and \textsc{tr}, indicating that such languages are over-segmented. mBERT's fertility is the lowest for \textsc{en}; this is  due to mBERT having seen the most data in this language during training, as well as English being morphologically poor in contrast to languages such as \textsc{ar}, \textsc{fi}, \textsc{ru}, or \textsc{tr}.\footnote{The \textsc{ja} model is the only monolingual BERT with a fertility score higher than mBERT; its tokenizer is character-based and thus by design produces the maximum number of subwords.}

The second metric we employ is the proportion of words where the tokenized word is continued across at least two sub-tokens (denoted by continuation symbols \#\#). Whereas the fertility is concerned with how aggressively a tokenizer splits, this metric measures how often it splits words. Intuitively, low scores are preferable for both metrics as they indicate that the tokenizer is well suited to the language.
The plots in Figure~\ref{fig:continuation} show similar trends as with the fertility statistic. In addition to \textsc{ar}, \textsc{fi}, \textsc{ko}, \textsc{ru}, and \textsc{tr}, which already displayed differences in fertility, mBERT also produces a proportion of continued words more than twice as high as the monolingual model for \textsc{id}.\footnote{We discuss additional tokenization statistics, further highlighting the differences (or lack thereof) between the individual monolingual tokenizers and the mBERT tokenizer, in Appendix~\ref{a:tokenization_analysis}.}

\subsection{New Pretrained Models}

The differences in pretraining corpora and tokenizer statistics seem to align with the variations in downstream performance across languages. In particular, it appears that the performance gains of monolingual models over mBERT are larger for languages where the differences between the respective tokenizers and pretraining corpora sizes are also larger (\textsc{ar}, \textsc{fi}, \textsc{ko}, \textsc{ru}, \textsc{tr} vs. \textsc{en}, \textsc{ja}, \textsc{zh}).\footnote{The only exception is \textsc{id}, where the monolingual model has seen significantly more data and also scores lower on the tokenizer metrics, yet underperforms mBERT in most tasks. We suspect this exception is because IndoBERT is uncased, whereas the remaining models are cased.} 
This implies that both the data size and the tokenizer are among the main driving forces of downstream task performance. To disentangle the effects of these two factors, we pretrain new models for \textsc{ar}, \textsc{fi}, \textsc{id}, \textsc{ko}, and \textsc{tr} (the languages that exhibited the largest discrepancies in tokenization and pretraining data size) on Wikipedia data. 

We train four model variants for each language. First, we train two new monolingual BERT models on the same data, one with the original monolingual tokenizer ({\footnotesize\textsc{MonoModel-MonoTok}}) and one with the mBERT tokenizer ({\footnotesize\textsc{MonoModel-mBERTTok}}).\footnote{The only exception is \textsc{id}; instead of relying on the uncased IndoBERT tokenizer by \citet{wilie-etal-2020-indonlu}, we introduce a new \textit{cased} tokenizer with identical vocabulary size (30,521).}
Second, similar to \citet{artetxe-etal-2020-cross}, we retrain the embedding layer of mBERT, once with the respective monolingual tokenizer ({\footnotesize\textsc{mBERTModel-MonoTok}}) and once with the mBERT tokenizer ({\footnotesize\textsc{mBERTModel-mBERTTok}}). We freeze the transformer and only retrain the embedding weights, thus largely preserving mBERT's multilinguality. The reason we retrain mBERT's embedding layer with its own tokenizer is to further eliminate confounding factors when comparing to the version of mBERT with monolingually retrained embeddings.
By comparing models trained on the same amount of data, but with different tokenizers ({\footnotesize\textsc{MonoModel-MonoTok}} vs. {\footnotesize\textsc{MonoModel-mBERTTok}}, {\footnotesize\textsc{mBERTModel-mBERTTok}} vs. {\footnotesize\textsc{mBERTModel-MonoTok}}), we  disentangle the effect of the dataset size from the tokenizer, both with monolingual and multilingual LM variants. 

\vspace{1.6mm}
\noindent\textbf{Pretraining Setup.} We pretrain new BERT models for each language on its respective Wikipedia dump.\footnote{We use Wiki dumps from June 20, 2020 (e.g., fiwiki-20200720-pages-articles.xml.bz2 for \textsc{fi}).} 
We apply two preprocessing steps to obtain clean data for pretraining. First, we use WikiExtractor \cite{Wikiextractor2015} to extract text passages from the raw dumps. Next, we follow \citet{pyysalo2020wikibert} and utilize UDPipe \cite{straka-etal-2016-udpipe} parsers pretrained on UD data to segment the extracted text passages into texts with document, sentence, and word boundaries.

Following \citet{liu:2019,wu-dredze-2020-languages}, we only use the masked language modeling (MLM) objective and omit the next sentence prediction task. Besides that, we largely follow the default pretraining procedure by \citet{devlin:2019}.
We pretrain the new monolingual LMs ({\footnotesize$\textsc{MonoModel-*}$}) from scratch for 1M steps.\footnote{The batch size is 64; the sequence length is 128 for the first 900,000 steps, and 512 for the remaining 100,000 steps.}
We enable whole word masking \cite{devlin:2019} for the \textsc{fi} monolingual models, following the pretraining procedure for FinBERT \cite{virtanen:2019}. 
For the retrained mBERT models ({\footnotesize\textsc{mBERTModel-*}}), we train for 250,000 steps following \citet{artetxe-etal-2020-cross}.\footnote{We train with batch size 64 and sequence length 512, otherwise using the same hyper-parameters as for the monolingual models.} 
We freeze all parameters outside the embedding layer.\footnote{
For more details see Appendix \ref{sec:app:training_new_models}.}

\vspace{1.6mm}
\noindent\textbf{Results.}
We perform the same evaluations on downstream tasks for our new models as described in \S\ref{sec:experimental_setup}, and report the results in Table~\ref{tab:new_models_results}.\footnote{Full results including development set scores are available in Table~\ref{tab:full_results_new_models} of the Appendix.}

\begin{table}[!t]
\centering
\resizebox{0.48\textwidth}{!}{%
\begin{tabular}{@{}lllllccccccc@{}}
\toprule
\multirow{3}{*}{\textbf{Lg}}  & \multicolumn{2}{l}{\multirow{3}{*}{\textbf{Model}}}       & \textbf{NER}  &  & \textbf{SA}   &  & \textbf{QA}                   &  & \textbf{UDP}                  &  & \textbf{POS}  \\ 
                     & \multicolumn{2}{c}{}                             & Test          &  & Test          &  & Dev                           &  & Test                          &  & Test          \\
                     & \multicolumn{2}{c}{}                             & $F_1$         &  & Acc           &  & EM / $F_1$                    &  & UAS / LAS                     &  & Acc           \\ \addlinespace[0.4em]\midrule\addlinespace[0.4em]

\multirow{6}{*}{\vspace{-2.5em}\textsc{ar}}  & 

                     \multicolumn{2}{l}{Monolingual} & 91.1 &  & \textbf{95.9} &  & \textbf{68.3} / \textbf{82.4} &  & \textbf{90.1} / \textbf{85.6} &  & 96.8 \\ \addlinespace[0.4em]\addlinespace[0.4em]
                     & \multicolumn{2}{l}{\footnotesize\textsc{MonoModel-MonoTok}}   & \underline{\textbf{91.7}} &  & \underline{95.6} &  & \underline{67.7} / \underline{81.6} &  & \underline{89.2} / \underline{84.4} &  & 96.6          \\
                     & \multicolumn{2}{l}{\footnotesize\textsc{MonoModel-mBERTTok}}  & 90.0          &  & 95.5          &  & 64.1 / 79.4                   &  & 88.8 / 84.0                   &  & \underline{\textbf{97.0}} \\ \addlinespace[0.4em]\addlinespace[0.4em]
                     & \multicolumn{2}{l}{\footnotesize\textsc{mBERTModel-MonoTok}}  & \underline{91.2} &  & 95.4          &  & \underline{66.9} / \underline{81.8} &  & \underline{89.3} / \underline{84.5} &  & 96.4          \\
                     & \multicolumn{2}{l}{\footnotesize\textsc{mBERTModel-mBERTTok}} & 89.7          &  & \underline{95.6} &  & 66.3 / 80.7                   &  & 89.1 / 84.2                   &  & \underline{96.8} \\ \addlinespace[0.4em]\addlinespace[0.4em]
                     & \multicolumn{2}{l}{mBERT} & 90.0 &  & 95.4 &  & 66.1 / 80.6 &  & 88.8 / 83.8 &  & 96.8 \\\addlinespace[0.4em]\midrule\addlinespace[0.4em]

\multirow{6}{*}{\vspace{-2.5em}\textsc{fi}}  & 

                     \multicolumn{2}{l}{Monolingual} & \textbf{92.0} &  & ----- &  & \textbf{69.9} / \textbf{81.6} &  & \textbf{95.9} / \textbf{94.4} &  & \textbf{98.4} \\ \addlinespace[0.4em]\addlinespace[0.4em]
                     & \multicolumn{2}{l}{\footnotesize\textsc{MonoModel-MonoTok}}   & 89.1          &  & -----         &  & \underline{66.9} / \underline{79.5} &  & \underline{93.7} / \underline{91.5} &  & \underline{97.3} \\
                     & \multicolumn{2}{l}{\footnotesize\textsc{MonoModel-mBERTTok}}  & \underline{90.0} &  & -----         &  & 65.1 / 77.0                   &  & 93.6 / \underline{91.5}          &  & 97.0          \\ \addlinespace[0.4em]\addlinespace[0.4em]
                     & \multicolumn{2}{l}{\footnotesize\textsc{mBERTModel-MonoTok}}  & \underline{88.1} &  & -----         &  & \underline{66.4} / \underline{78.3} &  & \underline{92.4} / \underline{89.6} &  & 96.6          \\
                     & \multicolumn{2}{l}{\footnotesize\textsc{mBERTModel-mBERTTok}} & \underline{88.1} &  & -----         &  & 65.9 / 77.3                   &  & 92.2 / 89.4                   &  & \underline{96.7} \\ \addlinespace[0.4em]\addlinespace[0.4em]
                     & \multicolumn{2}{l}{mBERT} & 88.2 &  & ----- &  & 66.6 / 77.6 &  & 91.9 / 88.7 &  & 96.2 \\ \addlinespace[0.4em]\midrule\addlinespace[0.4em]

\multirow{6}{*}{\vspace{-2.5em}\textsc{id}} & 
                     \multicolumn{2}{l}{Monolingual} & 91.0 &  & \textbf{96.0} &  & 66.8 / 78.1 &  & 85.3 / 78.1 &  & 92.1 \\ \addlinespace[0.4em]\addlinespace[0.4em]
                     & \multicolumn{2}{l}{\footnotesize\textsc{MonoModel-MonoTok}}   & 92.5          &  & \underline{\textbf{96.0}} &  & \underline{73.1} / \underline{83.6} &  & \underline{85.0} / 78.5          &  & \underline{\textbf{93.9}} \\
                     & \multicolumn{2}{l}{\footnotesize\textsc{MonoModel-mBERTTok}}  & \underline{93.2} &  & 94.8          &  & 67.0 / 79.2                   &  & 84.9 / \underline{78.6}          &  & 93.6          \\ \addlinespace[0.4em]\addlinespace[0.4em]
                     & \multicolumn{2}{l}{\footnotesize\textsc{mBERTModel-MonoTok}}  & \underline{\textbf{93.9}} &  & \underline{94.6} &  & \underline{\textbf{74.1}} / \underline{\textbf{83.8}} &  & \underline{\textbf{86.4}} / \underline{\textbf{80.2}} &  & \underline{93.8} \\
                     & \multicolumn{2}{l}{\footnotesize\textsc{mBERTModel-mBERTTok}} & \underline{\textbf{93.9}} &  & \underline{94.6} &  & 71.9 / 82.7                   &  & 86.2 / 79.6                   &  & 93.7          \\ \addlinespace[0.4em]\addlinespace[0.4em]
                     & \multicolumn{2}{l}{mBERT} & 93.5 &  & 91.4 &  & 71.2 / 82.1 &  & 85.9 / 79.3 &  & 93.5 \\ \addlinespace[0.4em]\midrule\addlinespace[0.4em]

\multirow{6}{*}{\vspace{-2.5em}\textsc{ko}}  & 
                     \multicolumn{2}{l}{Monolingual} & \textbf{88.8} &  & \textbf{89.7} &  & \textbf{74.2} / \textbf{91.1} &  & \textbf{90.3} / \textbf{87.2} &  & \textbf{97.0} \\ \addlinespace[0.4em]\addlinespace[0.4em]
                     & \multicolumn{2}{l}{\footnotesize\textsc{MonoModel-MonoTok}}   & \underline{87.1} &  & \underline{88.8} &  & \underline{72.8} / \underline{90.3} &  & \underline{89.8} / \underline{86.6} &  & \underline{96.7} \\
                     & \multicolumn{2}{l}{\footnotesize\textsc{MonoModel-mBERTTok}}  & 85.8          &  & 87.2          &  & 68.9 / 88.7                   &  & 88.9 / 85.6                   &  & 96.4          \\ \addlinespace[0.4em]\addlinespace[0.4em]
                     & \multicolumn{2}{l}{\footnotesize\textsc{mBERTModel-MonoTok}}  & \underline{86.6} &  & \underline{88.1} &  & \underline{72.9} / \underline{90.2} &  & \underline{90.1} / \underline{87.0} &  & \underline{96.5} \\
                     & \multicolumn{2}{l}{\footnotesize\textsc{mBERTModel-mBERTTok}} & 86.2          &  & 86.6          &  & 69.3 / 89.3                   &  & 89.2 / 85.9                   &  & 96.2          \\ \addlinespace[0.4em]\addlinespace[0.4em]
                     & \multicolumn{2}{l}{mBERT} & 86.6 &  & 86.7 &  & 69.7 / 89.5 &  & 89.2 / 85.7 &  & 96.0 \\ \addlinespace[0.4em]\midrule\addlinespace[0.4em]

\multirow{6}{*}{\vspace{-2.5em}\textsc{tr}} &  
                     \multicolumn{2}{l}{Monolingual} & 92.8 & & \textbf{88.8} & & \textbf{60.6} / \textbf{78.1} & & \textbf{79.8} / \textbf{73.2} & & \textbf{96.9} \\ \addlinespace[0.4em]\addlinespace[0.4em]
                     & \multicolumn{2}{l}{\footnotesize\textsc{MonoModel-MonoTok}} & \underline{93.4} &  & \underline{87.0} &  & \underline{56.2} / \underline{73.7} &  & \underline{76.1} / \underline{68.9} &  & 96.3 \\
                     & \multicolumn{2}{l}{\footnotesize\textsc{MonoModel-mBERTTok}}  & 93.3          &  & 84.8          &  & 55.3 / 72.5                   &  & 75.3 / 68.3                   &  & \underline{96.5} \\ \addlinespace[0.4em]\addlinespace[0.4em]
                     & \multicolumn{2}{l}{\footnotesize\textsc{mBERTModel-MonoTok}}  & 93.7          &  & 85.3          &  & \underline{59.4} / \underline{76.7} &  & \underline{77.1} / \underline{70.2} &  & \underline{96.3} \\
                     & \multicolumn{2}{l}{\footnotesize\textsc{mBERTModel-mBERTTok}} & \underline{\textbf{93.8}} &  & \underline{86.1} &  & 58.7 / 76.6                   &  & 76.2 / 69.2                   &  & \underline{96.3} \\ \addlinespace[0.4em]\addlinespace[0.4em]
                     & \multicolumn{2}{l}{mBERT} & \textbf{93.8} &  & 86.4 &  & 57.9 / 76.4 &  & 74.5 / 67.4 &  & 95.7 \\ \addlinespace[0.4em]\midrule\midrule\addlinespace[0.4em]

\multirow{6}{*}{\vspace{-2.5em}\textsc{avg}} &
                     \multicolumn{2}{l}{Monolingual} & \textbf{91.1} && \textbf{92.6} && \textbf{68.0} / \textbf{82.3} && \textbf{88.3} / \textbf{83.7} && \textbf{96.2} \\ \addlinespace[0.4em]\addlinespace[0.4em]
                     & \multicolumn{2}{l}{\footnotesize\textsc{MonoModel-MonoTok}} & \underline{90.8} && \underline{91.9} && \underline{67.3} / \underline{81.7} && \underline{86.8} / \underline{82.0}  && \underline{\textbf{96.2}} \\ 
                     & \multicolumn{2}{l}{\footnotesize\textsc{MonoModel-mBERTTok}}  & 90.5 && 90.6 && 64.1 / 79.4 && 86.3 / 81.6 && 96.1    \\ \addlinespace[0.4em]\addlinespace[0.4em]
                     & \multicolumn{2}{l}{\footnotesize\textsc{mBERTModel-MonoTok}}  & \underline{90.7} && \underline{90.9} && \underline{\textbf{68.0}} / \underline{82.2} && \underline{87.1} / \underline{82.3} && \underline{95.9} \\
                     & \multicolumn{2}{l}{\footnotesize\textsc{mBERTModel-mBERTTok}} & 90.3 && 90.7 && 66.4 / 81.3 && 86.6 / 81.7 && \underline{95.9} \\ \addlinespace[0.4em]\addlinespace[0.4em]
                     & \multicolumn{2}{l}{mBERT} & 90.4 && 90.0 && 66.3 / 81.2 && 86.1 / 81.0 && 95.6 \\ \addlinespace[0.4em]\bottomrule

\end{tabular}
}
\caption{Performance of our new {\footnotesize\textsc{MonoModel-*}} and {\footnotesize\textsc{mBERTModel-*}} models (see \S\ref{sec:app:training_new_models}) fine-tuned for the NER, SA, QA, UDP, and POS tasks (see \S\ref{sec:tasks}), compared to the monolingual models from prior work and fully fine-tuned mBERT. We group model counterparts w.r.t. tokenizer choice to facilitate a direct comparison between respective counterparts. We use development sets only for QA. \textbf{Bold} denotes best score across all models for a given language and task. \underline{Underlined} denotes best score compared to its respective counterpart.}
\label{tab:new_models_results}
\vspace{-2.5mm}
\end{table}

The results indicate that the models trained with dedicated monolingual tokenizers outperform their counterparts with multilingual tokenizers in most tasks, with particular consistency for QA, UDP, and SA. In NER, the models trained with multilingual tokenizers score competitively or higher than the monolingual ones in half of the cases. Overall, the performance gap is the smallest for POS tagging (at most 0.4\% accuracy). We observe the largest gaps for QA (6.1 EM / 4.4 $F_1$ in \textsc{id}), SA (2.2\% accuracy in \textsc{tr}), and NER (1.7 $F_1$ in \textsc{ar}). Although the only language in which the monolingual counterpart always comes out on top is \textsc{ko}, the multilingual counterpart comes out on top at most 3/10 times (for \textsc{ar} and \textsc{tr}) in the other languages. The largest decrease in performance of a monolingual tokenizer relative to its multilingual counterpart is found for SA in \textsc{tr} (0.8\% accuracy).

Overall, we find that for 38 out of 48 task, model, and language combinations, the monolingual tokenizer outperforms the mBERT counterpart. We were able to improve the monolingual performance of the original mBERT for 20 out of 24 languages and tasks by only replacing the tokenizer and, thus, leveraging a specialized monolingual version.
Similar to how the chosen method of tokenization affects neural machine translation quality \cite{domingo:2019}, these results establish that, in fact, the designated pretrained tokenizer plays a fundamental role in the monolingual downstream task performance of contemporary LMs. 

In 18/24 language and task settings, the monolingual model from prior work (trained on more data) outperforms its corresponding  {\footnotesize\textsc{MonoModel-MonoTok}} model. 4/6 settings in which our  {\footnotesize\textsc{MonoModel-MonoTok}} model performs better are found for \textsc{id}, where IndoBERT uses an uncased tokenizer and our model uses a cased one, which may affect the comparison. Expectedly, these results strongly indicate that data size plays a major role in downstream performance and corroborate prior research findings \cite[\textit{inter alia}]{liu:2019, conneau:2020, zhang2020need}.

\subsection{Adapter-Based Training}
\label{sec:adapter_based_training}
Another way to provide more language-specific capacity to a multilingual LM beyond a dedicated tokenizer, thereby potentially making gains in monolingual downstream performance, is to introduce adapters \cite{pfeiffer:2020b, Pfeiffer20Gib, ustun-etal-2020-udapter}, a small number of additional parameters at every layer of a pretrained model.
To train adapters, usually all pretrained weights are frozen, while only the adapter weights are fine-tuned.\footnote{\citet{pfeiffer:2020b} propose to stack task-specific adapters on top of language adapters and extend this approach in \citet{Pfeiffer20Gib} by additionally training new embeddings for the target language.} The adapter-based approaches thus offer increased efficiency and modularity; it is crucial to verify to which extent our findings extend to the more efficient and more versatile adapter-based fine-tuning setup.

\begin{table}[!t]
\centering
\resizebox{0.48\textwidth}{!}{%
\begin{tabular}{@{}llccccc@{}}
\toprule
\multirow{3}{*}{\textbf{Lg}} & \multirow{3}{*}{\textbf{Model}} & \multicolumn{1}{l}{\textbf{NER}} &   \textbf{SA} &   \textbf{QA} &   \textbf{UDP} &   \textbf{POS} \\
& & Test &  Test &   Dev &   Test &   Test \\
& & $F_1$ &   Acc &  EM / $F_1$ &  UAS / LAS &   Acc \\ \midrule
 \multirow{4}{*}{\textsc{ar}} & mBERT & 90.0 &   95.4 &   66.1 / 80.6 &   \textbf{88.8} / \textbf{83.8} &   \textbf{96.8} \\
  & \, \, + A\textsuperscript{Task} & 89.6 &  95.6 &  66.7 / 81.1 &   87.8 / 82.6 &   \textbf{96.8} \\
 & \, \, + A\textsuperscript{Task} + A\textsuperscript{Lang} & 89.7 &   \textbf{95.7}   & 66.9 / 81.0 &   88.0 / 82.8 &   \textbf{96.8} \\
 & \, \, + A\textsuperscript{Task} + A\textsuperscript{Lang} + {\footnotesize\textsc{MonoTok}} & \textbf{91.1} &  \textbf{95.7}  & \textbf{67.7} / \textbf{82.1} &  88.5 / 83.4 &  96.5 \\ \midrule
\multirow{4}{*}{\textsc{fi}}  & mBERT & 88.2 &   ----- &   66.6 / 77.6 &   91.9 / 88.7 &   96.2 \\
  & \, \, + A\textsuperscript{Task} & \textbf{88.5} &   ----- &   65.2 / 77.3 &   90.8 / 87.0 &   95.7 \\
 & \, \, +  A\textsuperscript{Task} + A\textsuperscript{Lang} & 88.4 &   ----- &  65.7 / 77.1 &   91.8 / 88.5 &   96.6 \\
 & \, \,  + A\textsuperscript{Task} + A\textsuperscript{Lang} + {\footnotesize\textsc{MonoTok}} & 88.1 &  ----- &  \textbf{66.7} / \textbf{79.0} &  \textbf{92.8} / \textbf{90.1} &  \textbf{97.3} \\ \midrule
\multirow{4}{*}{\textsc{id}}  & mBERT & \textbf{93.5} &   91.4 &   71.2 / 82.1 &   \textbf{85.9} / \textbf{79.3} &  \textbf{93.5}\\
  & \, \, + A\textsuperscript{Task} & \textbf{93.5} &   90.6 &   70.6 / 82.5 &   84.8 / 77.4 &   93.4 \\
 & \, \, +  A\textsuperscript{Task} + A\textsuperscript{Lang}& \textbf{93.5} &   93.6 &   70.8 / 82.2 &   85.4 / 78.1 &   93.4 \\
 &\, \,  + A\textsuperscript{Task} + A\textsuperscript{Lang} + {\footnotesize\textsc{MonoTok}} & 93.4 &  \textbf{93.8} & \textbf{74.4} / \textbf{84.4} &  85.1 / 78.3 &  \textbf{93.5} \\ \midrule
\multirow{4}{*}{\textsc{ko}} & mBERT & \textbf{86.6} &   86.7 &   69.7 / 89.5 &   \textbf{89.2} / \textbf{85.7} &   96.0 \\
  & \, \, + A\textsuperscript{Task} & 86.2 &   86.5 &  69.8 / 89.7 &   87.8 / 83.9 &   96.2 \\
 & \, \, +  A\textsuperscript{Task} + A\textsuperscript{Lang} & 86.2 &   86.3 &   70.0 / 89.8 &   88.3 / 84.3 &   96.2 \\
 & \, \, + A\textsuperscript{Task} + A\textsuperscript{Lang} +  {\footnotesize\textsc{MonoTok}} & 86.5 &  \textbf{87.9} &  \textbf{73.1} / \textbf{90.4} &  88.9 / 85.2 &  \textbf{96.5} \\ \midrule
\multirow{4}{*}{\textsc{tr}} & mBERT & \textbf{93.8} &   \textbf{86.4} &   57.9 / 76.4 &   74.5 / 67.4 &   95.7 \\
  & \, \, + A\textsuperscript{Task} & 93.0 &   83.9 &  55.3 / 75.1 &   72.4 / 64.1 &   95.7 \\
 & \, \, +  A\textsuperscript{Task} + A\textsuperscript{Lang} & 93.5 &   84.8 &   56.9 / 75.8 &  73.0 / 64.7 &   95.9 \\
 & \, \,  +  A\textsuperscript{Task} + A\textsuperscript{Lang} + {\footnotesize\textsc{MonoTok}} & 92.7 & 85.3 &  \textbf{60.0} / \textbf{77.0} &  \textbf{75.7} / \textbf{68.1} &  \textbf{96.3} \\ \midrule
 \midrule
\multirow{4}{*}{\textsc{avg}} & mBERT & \textbf{90.4} & 90.0 & 66.3 / 81.2 & 86.0 / \textbf{81.0} & 95.6 \\
  & \, \, + A\textsuperscript{Task} & 90.2 & 89.2 & 65.5 / 81.1 & 84.7 / 79.0 & 95.6 \\
 & \, \, +  A\textsuperscript{Task} + A\textsuperscript{Lang} & 90.3 & 90.1 & 66.1 / 81.2 & 85.3 / 79.7 & 95.8  \\
 & \, \,  +  A\textsuperscript{Task} + A\textsuperscript{Lang} + {\footnotesize\textsc{MonoTok}} & \textbf{90.4} & \textbf{90.7} & \textbf{68.4} / \textbf{82.6} & \textbf{86.2} / \textbf{81.0} & \textbf{96.0}  \\
\bottomrule
\end{tabular}%
}
\caption{
Performance on the different  tasks leveraging mBERT with different adapter components (see \S\ref{sec:adapter_based_training}). 
}
\label{tab:adapter_results_table_test}
\vspace{-1.5mm}
\end{table}

We evaluate the impact of different adapter components on the downstream task performance  and their complementarity with monolingual tokenizers in Table \ref{tab:adapter_results_table_test}.\footnote{See Appendix Table~\ref{tab:full_results_adapters} for the results on dev sets.} Here, $+A^{Task}$ and $+A^{Lang}$ implies adding task- and language-adapters respectively, whereas  $+${\footnotesize\textsc{MonoTok}} additionally includes a new embedding layer. As mentioned, we only fine-tune adapter weights on the downstream task, leveraging the adapter architecture proposed by \citet{pfeiffer:2020a}. For the $+A^{Task} +A^{Lang}$ setting we leverage  pretrained language adapter weights available at \href{https://AdapterHub.ml}{AdapterHub.ml} \cite{pfeiffer:2020c}. Language adapters are added to the model and frozen while only task adapters are trained on the target task. For the $+A^{Task} +A^{Lang} + $ {\footnotesize\textsc{MonoTok}} we train language adapters and new embeddings with the corresponding monolingual tokenizer equally as described in the previous section (e.g. {\footnotesize\textsc{mBERTModel-MonoTok}}), task adapters are trained with a learning rate of $5e-4$ and 30 epochs with early stopping.

\begin{figure}[!t]
    \centering
    \begin{subfigure}[!t]{0.561\columnwidth}
        \centering
        \includegraphics[width=0.975\linewidth]{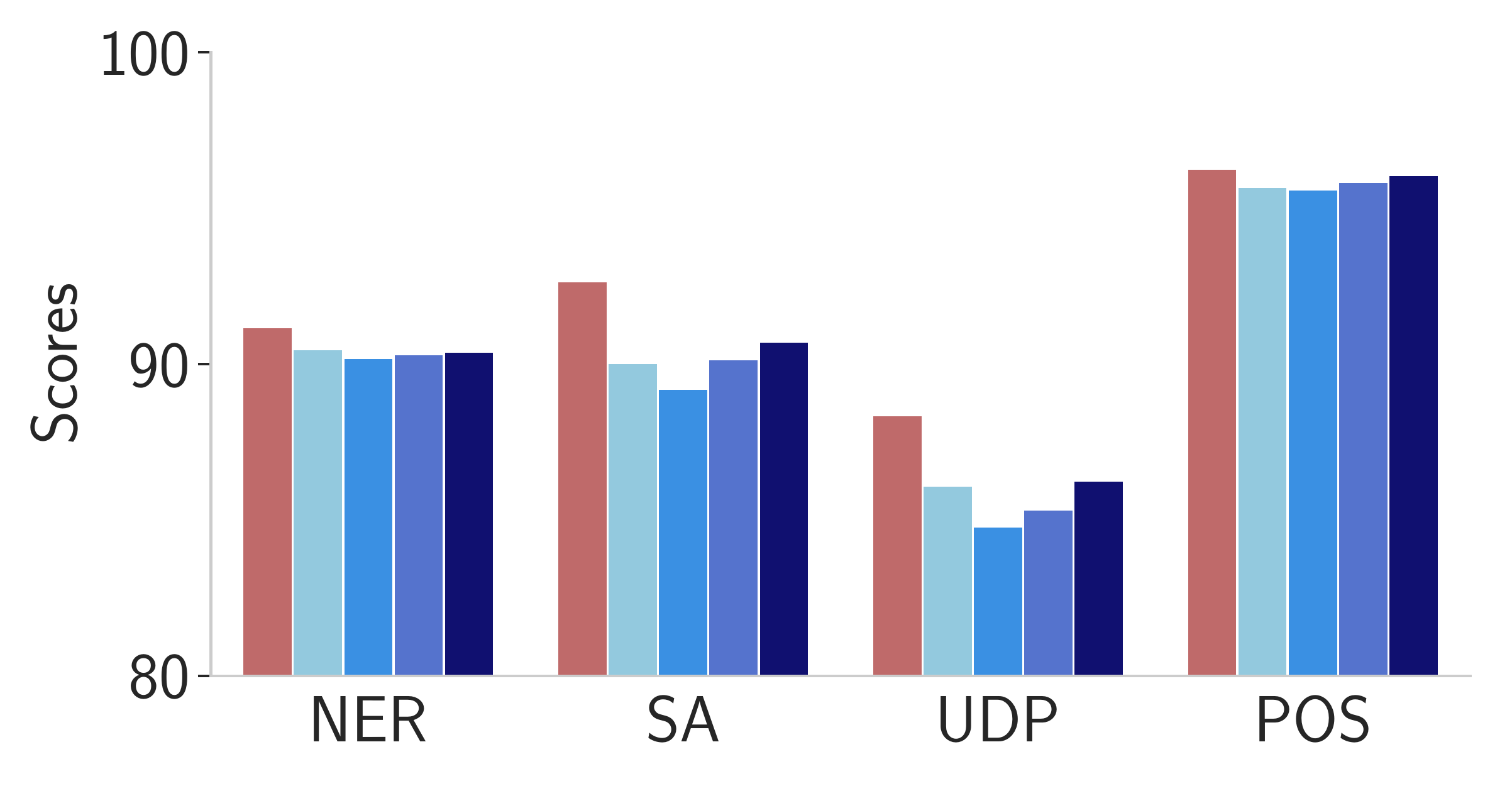}
    \end{subfigure}
    \hspace{-1em}
    \begin{subfigure}[!t]{0.464\columnwidth}
        \centering
        \includegraphics[width=0.98\linewidth]{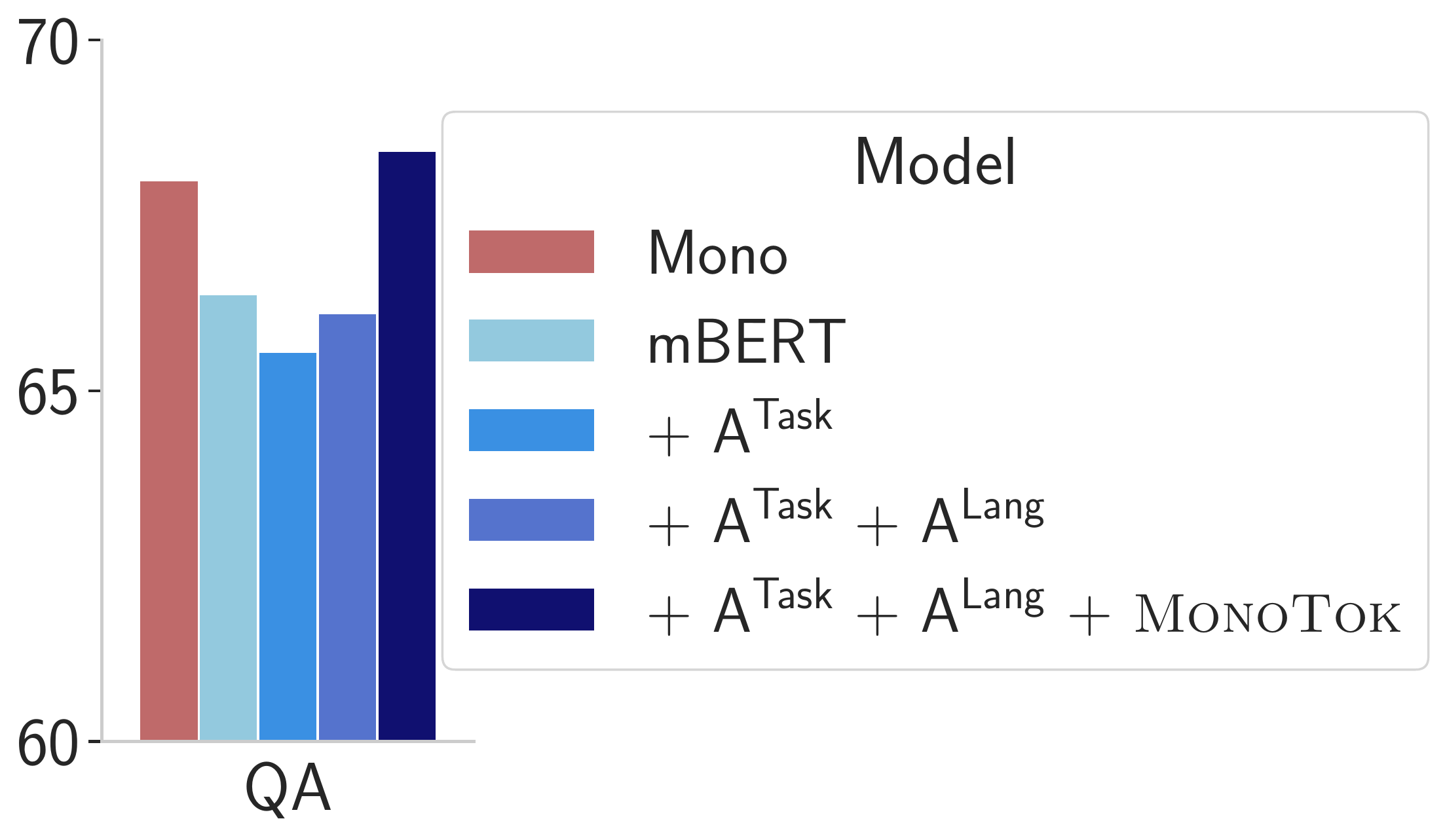}
    \end{subfigure}
    \caption{Task performance averaged over all languages for different models: fully fine-tuned monolingual (\textbf{Mono}), fully fine-tuned mBERT (\textbf{mBERT}), mBERT with task adapter ($\mathbf{+A^{Task}}$), with task and language adapter ($\mathbf{+A^{Task}+A^{Lang}}$), with task and language adapter and embedding layer retraining ($\mathbf{+A^{Task}+A^{Lang} +  }$ {\footnotesize\textsc{\textbf{MonoTok}}}).}
\label{fig:adapterplot}
\vspace{-1.5mm}
\end{figure}

\vspace{1.6mm}
\noindent\textbf{Results.} Similar to previous findings, adapters improve upon mBERT in 18/24 language, and task settings, 13 of which can be attributed to the improved {\footnotesize\textsc{mBERTModel-MonoTok}} tokenizer. Figure~\ref{fig:adapterplot} illustrates the average performance of the different adapter components in comparison to the monolingual models. We find that adapters with dedicated tokenizers  reduce the performance gap considerably without leveraging more training data, and even outperform the monolingual models in QA. This finding shows that adding additional language-specific capacity to existing multilingual LMs, which can be achieved with adapters in a portable and efficient way, is a viable alternative to monolingual pretraining.

\section{Further Analysis}
At first glance, our results displayed in Table~\ref{tab:results_table_test} seem to confirm the prevailing view that monolingual models are more effective than multilingual models \cite[\textit{inter alia}]{ronnqvist:2019, antoun:2020, vries:2019}. However, the broad scope of our experiments reveals certain nuances that were previously undiscovered. Unlike prior work, which primarily attributes gaps in performance to mBERT being under-trained \cite{ronnqvist:2019, wu-dredze-2020-languages}, our disentangled results (Table~\ref{tab:new_models_results}) suggest that a large portion of 
existing performance gaps can be attributed to the capability of the tokenizer.

With monolingual tokenizers with lower fertility and proportion-of-continued-words values than the mBERT tokenizer (such as for \textsc{ar}, \textsc{fi}, \textsc{id}, \textsc{ko}, \textsc{tr}), consistent gains can be achieved, irrespective of whether the LMs are monolingual (the {\footnotesize\textsc{MonoModel-*}} comparison) or multilingual (a comparison of {\footnotesize\textsc{mBERTModel-*}} variants).

Whenever the differences between monolingual models and mBERT with respect to the tokenizer properties
and the pretraining corpus size are small (e.g., for \textsc{en}, \textsc{ja}, and \textsc{zh}), the performance gap is typically negligible. In QA, we even find mBERT to be favorable for these languages. Therefore, we conclude that monolingual models are not superior to multilingual ones per se, but gain
advantage in direct comparisons by incorporating more pretraining data and using language-adapted tokenizers.

\begin{figure}[!t]
    \centering
    \includegraphics[width=\columnwidth]{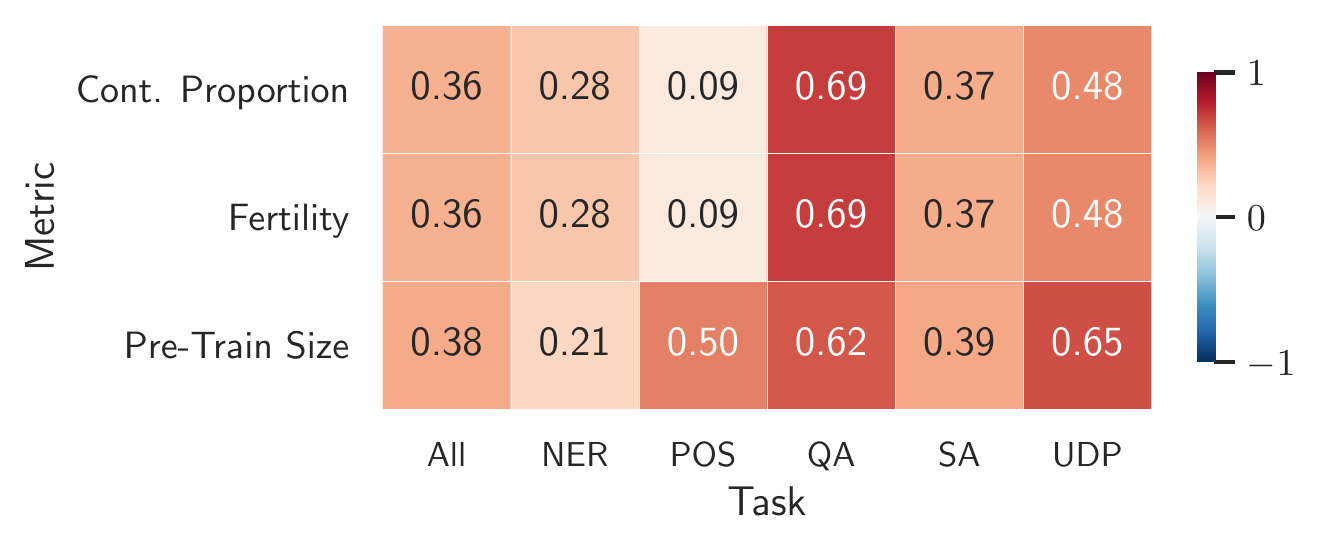}
    \caption[Spearman's $\rho$ correlation of proportion of continued words, fertility, and pretraining corpus size with downstream performance (Counterparts, Indonesian excluded)]{Spearman's $\rho$ correlation of a relative decrease in the proportion of continued words (Cont. Proportion), a relative decrease in fertility, and a relative increase in pretraining corpus size with a relative increase in downstream performance over fully fine-tuned mBERT. For the proportion of continued words and the fertility, we consider fully fine-tuned mBERT, the {\footnotesize\textsc{MonoModel-*}} models, and the {\footnotesize\textsc{mBERTModel-*}} models. For the pretraining corpus size, we consider the original monolingual models and the {\footnotesize\textsc{MonoModel-MonoTok}} models. We exclude the \textsc{ID} models (see Appendix~\ref{sec:app:correlation_analysis} for the clarification).}
    \label{fig:new_models_correlation_no_id}
\end{figure}

\vspace{1.5mm}
\noindent \textbf{Correlation Analysis.}
\label{sec:correlation_analysis}
To uncover additional patterns in our results (Tables~\ref{tab:results_table_test}, \ref{tab:new_models_results}, \ref{tab:adapter_results_table_test}), we perform a statistical analysis assessing the correlation between the individual factors (pretraining data size, subword fertility, proportion of continued words) and the downstream performance. Although our framework may not provide enough data points to be statistically representative, we argue that the correlation coefficient can still provide reasonable indications and reveal relations not immediately evident by looking at the tables.

Figure~\ref{fig:new_models_correlation_no_id} shows that both decreases in the proportion of continued words and the fertility correlate with an increase in downstream performance relative to fully fine-tuned mBERT across all tasks. The correlation is stronger for UDP and QA, where we find models with monolingual tokenizers to outperform their counterparts with the mBERT tokenizer consistently. The correlation is weaker for NER and POS tagging, which is also expected, considering the inconsistency of the results.\footnote{For further information, see Appendix \ref{sec:app:correlation_analysis}.}

Overall, we find that the fertility and the proportion of continued words have a similar effect on the monolingual downstream performance as the corpus size for pretraining; This indicates that the tokenizer's ability of representing a language plays a crucial role; Consequently, choosing a sub-optimal tokenizer typically results in deteriorated downstream performance.

\section{Conclusion}
We have conducted the first comprehensive empirical investigation concerning the monolingual performance of monolingual and multilingual language models (LMs). While our results support the existence of a performance gap in most but not all languages and tasks, further analyses revealed that the gaps are often substantially smaller than what was previously assumed. The gaps exist in certain languages due to the discrepancies in \textbf{1)} pretraining data size, and \textbf{2)} chosen tokenizers, and the level of their adaptation to the target language.

Further, we have disentangled the impact of pretrained corpora size from the influence of the tokenizers on the downstream task performance. We have trained new monolingual LMs on the same data, but with two different tokenizers;  one being the dedicated tokenizer of the monolingual LM provided by native speakers; the other being the automatically generated multilingual mBERT tokenizer. We have found that for (almost) every task and language, the use of monolingual tokenizers outperforms the mBERT tokenizer.
 
Consequently, in line with recent work by \citet{ChungGTR20}, our results suggest that investing more effort into \textbf{1)} improving the balance of individual languages' representations in the vocabulary of multilingual LMs, and \textbf{2)} providing language-specific adaptations and extensions of multilingual tokenizers \cite{Pfeiffer20Gib} can reduce the gap between monolingual and multilingual LMs. Another promising future research direction is completely disposing of any (language-specific or multilingual) tokenizers during pretraining \cite{canine}.

Our code, pretrained models, and adapters are available at \href{https://github.com/Adapter-Hub/hgiyt}{https://github.com/Adapter-Hub/hgiyt}.
\section*{Acknowledgments}

Jonas Pfeiffer is supported by the LOEWE initiative (Hesse, Germany) within the emergenCITY center. The work of Ivan Vuli\'{c} is supported by the ERC Consolidator Grant LEXICAL: Lexical Acquisition Across Languages (no 648909). 

We thank Nils Reimers, Prasetya Ajie Utama, and Adhiguna Kuncoro for insightful feedback and suggestions on a draft of this paper.

\bibliographystyle{acl_natbib}
\bibliography{anthology, acl2021}

\begin{thebibliography}{71}
\expandafter\ifx\csname natexlab\endcsname\relax\def\natexlab#1{#1}\fi

\bibitem[{\'Acs(2019)}]{acs:2019}
Judit \'Acs. 2019.
\newblock \href
  {http://juditacs.github.io/2019/02/19/bert-tokenization-stats.html}
  {Exploring {BERT}'s {V}ocabulary}.
\newblock \emph{Blog Post}.

\bibitem[{Antoun et~al.(2020)Antoun, Baly, and Hajj}]{antoun:2020}
Wissam Antoun, Fady Baly, and Hazem Hajj. 2020.
\newblock \href {https://www.aclweb.org/anthology/2020.osact-1.2} {{A}ra{BERT}:
  Transformer-based model for {A}rabic language understanding}.
\newblock In \emph{Proceedings of the 4th Workshop on Open-Source Arabic
  Corpora and Processing Tools, with a Shared Task on Offensive Language
  Detection}, pages 9--15, Marseille, France. European Language Resource
  Association.

\bibitem[{Artetxe et~al.(2020)Artetxe, Ruder, and
  Yogatama}]{artetxe-etal-2020-cross}
Mikel Artetxe, Sebastian Ruder, and Dani Yogatama. 2020.
\newblock \href {https://doi.org/10.18653/v1/2020.acl-main.421} {On the
  cross-lingual transferability of monolingual representations}.
\newblock In \emph{Proceedings of the 58th Annual Meeting of the Association
  for Computational Linguistics}, pages 4623--4637, Online. Association for
  Computational Linguistics.

\bibitem[{Attardi(2015)}]{Wikiextractor2015}
Giusepppe Attardi. 2015.
\newblock \href {https://github.com/attardi/wikiextractor} {Wikiextractor}.
\newblock \emph{GitHub Repository}.

\bibitem[{Chau et~al.(2020)Chau, Lin, and Smith}]{ChauLS20Parsing}
Ethan~C. Chau, Lucy~H. Lin, and Noah~A. Smith. 2020.
\newblock \href {https://doi.org/10.18653/v1/2020.findings-emnlp.118} {Parsing
  with multilingual {BERT}, a small corpus, and a small treebank}.
\newblock In \emph{Findings of the Association for Computational Linguistics:
  EMNLP 2020}, pages 1324--1334, Online. Association for Computational
  Linguistics.

\bibitem[{Chung et~al.(2020)Chung, Garrette, Tan, and Riesa}]{ChungGTR20}
Hyung~Won Chung, Dan Garrette, Kiat~Chuan Tan, and Jason Riesa. 2020.
\newblock \href {https://doi.org/10.18653/v1/2020.emnlp-main.367} {Improving
  multilingual models with language-clustered vocabularies}.
\newblock In \emph{Proceedings of the 2020 Conference on Empirical Methods in
  Natural Language Processing (EMNLP)}, pages 4536--4546, Online. Association
  for Computational Linguistics.

\bibitem[{Clark et~al.(2020)Clark, Choi, Collins, Garrette, Kwiatkowski,
  Nikolaev, and Palomaki}]{clark:2020}
Jonathan~H. Clark, Eunsol Choi, Michael Collins, Dan Garrette, Tom Kwiatkowski,
  Vitaly Nikolaev, and Jennimaria Palomaki. 2020.
\newblock \href {https://doi.org/10.1162/tacl_a_00317} {{T}y{D}i {QA}: A
  benchmark for information-seeking question answering in typologically diverse
  languages}.
\newblock \emph{Transactions of the Association for Computational Linguistics},
  8:454--470.

\bibitem[{Clark et~al.(2021)Clark, Garrette, Turc, and Wieting}]{canine}
Jonathan~H. Clark, Dan Garrette, Iulia Turc, and John Wieting. 2021.
\newblock \href {http://arxiv.org/abs/2103.06874} {{CANINE: P}re-training an
  efficient tokenization-free encoder for language representation}.
\newblock \emph{arXiv preprint}.

\bibitem[{Conneau et~al.(2020)Conneau, Khandelwal, Goyal, Chaudhary, Wenzek,
  Guzm{\'a}n, Grave, Ott, Zettlemoyer, and Stoyanov}]{conneau:2020}
Alexis Conneau, Kartikay Khandelwal, Naman Goyal, Vishrav Chaudhary, Guillaume
  Wenzek, Francisco Guzm{\'a}n, Edouard Grave, Myle Ott, Luke Zettlemoyer, and
  Veselin Stoyanov. 2020.
\newblock \href {https://doi.org/10.18653/v1/2020.acl-main.747} {Unsupervised
  cross-lingual representation learning at scale}.
\newblock In \emph{Proceedings of the 58th Annual Meeting of the Association
  for Computational Linguistics}, pages 8440--8451, Online. Association for
  Computational Linguistics.

\bibitem[{Conneau et~al.(2018)Conneau, Rinott, Lample, Williams, Bowman,
  Schwenk, and Stoyanov}]{conneau-etal-2018-xnli}
Alexis Conneau, Ruty Rinott, Guillaume Lample, Adina Williams, Samuel Bowman,
  Holger Schwenk, and Veselin Stoyanov. 2018.
\newblock \href {https://doi.org/10.18653/v1/D18-1269} {{XNLI}: Evaluating
  cross-lingual sentence representations}.
\newblock In \emph{Proceedings of the 2018 Conference on Empirical Methods in
  Natural Language Processing}, pages 2475--2485, Brussels, Belgium.
  Association for Computational Linguistics.

\bibitem[{Demirtas and Pechenizkiy(2013)}]{demirtas:2013}
Erkin Demirtas and Mykola Pechenizkiy. 2013.
\newblock \href {https://doi.org/10.1145/2502069.2502078} {Cross-lingual
  polarity detection with machine translation}.
\newblock In \emph{Proceedings of the Second International Workshop on Issues
  of Sentiment Discovery and Opinion Mining (WISDOM '13)}, pages 9:1--8,
  Chicago, USA. Association for Computing Machinery.

\bibitem[{Devlin et~al.(2019)Devlin, Chang, Lee, and Toutanova}]{devlin:2019}
Jacob Devlin, Ming-Wei Chang, Kenton Lee, and Kristina Toutanova. 2019.
\newblock \href {https://doi.org/10.18653/v1/N19-1423} {{BERT}: Pre-training of
  deep bidirectional transformers for language understanding}.
\newblock In \emph{Proceedings of the 2019 Conference of the North {A}merican
  Chapter of the Association for Computational Linguistics: Human Language
  Technologies, Volume 1 (Long and Short Papers)}, pages 4171--4186,
  Minneapolis, Minnesota. Association for Computational Linguistics.

\bibitem[{Domingo et~al.(2019)Domingo, Garcıa-Martınez, Helle, Casacuberta,
  and Herranz}]{domingo:2019}
Miguel Domingo, Mercedes Garcıa-Martınez, Alexandre Helle, Francisco
  Casacuberta, and Manuel Herranz. 2019.
\newblock \href {http://arxiv.org/abs/1812.08621} {How much does tokenization
  affect neural machine translation?}
\newblock \emph{arXiv preprint}.

\bibitem[{Dozat and Manning(2017)}]{dozat:2017}
Timothy Dozat and Christopher~D. Manning. 2017.
\newblock \href {https://openreview.net/forum?id=Hk95PK9le} {Deep biaffine
  attention for neural dependency parsing}.
\newblock In \emph{Proceedings of the 5th International Conference on Learning
  Representations ({ICLR})}, Toulon, France. OpenReview.net.

\bibitem[{Efimov et~al.(2020)Efimov, Chertok, Boytsov, and
  Braslavski}]{efimov:2020}
Pavel Efimov, Andrey Chertok, Leonid Boytsov, and Pavel Braslavski. 2020.
\newblock \href {http://dx.doi.org/10.1007/978-3-030-58219-7_1} {{S}ber{Q}u{AD}
  – {R}ussian {R}eading {C}omprehension {D}ataset: Description and analysis}.
\newblock In \emph{CLEF 2020: Experimental IR Meets Multilinguality,
  Multimodality, and Interaction}, pages 3--15. Springer, Cham, Switzerland.

\bibitem[{Elnagar et~al.(2018)Elnagar, Khalifa, and Einea}]{elnagar:2018}
Ashraf Elnagar, Yasmin~S. Khalifa, and Anas Einea. 2018.
\newblock \href {https://doi.org/10.1007/978-3-319-67056-0_3} {{H}otel
  {A}rabic-{R}eviews {D}ataset {C}onstruction for {S}entiment {A}nalysis
  {A}pplications}.
\newblock In \emph{Intelligent Natural Language Processing: Trends and
  Applications}, pages 35--52. Springer, Cham, Switzerland.

\bibitem[{Gerz et~al.(2018)Gerz, Vuli{\'c}, Ponti, Reichart, and
  Korhonen}]{gerz-etal-2018-relation}
Daniela Gerz, Ivan Vuli{\'c}, Edoardo~Maria Ponti, Roi Reichart, and Anna
  Korhonen. 2018.
\newblock \href {https://doi.org/10.18653/v1/D18-1029} {On the relation between
  linguistic typology and (limitations of) multilingual language modeling}.
\newblock In \emph{Proceedings of the 2018 Conference on Empirical Methods in
  Natural Language Processing}, pages 316--327, Brussels, Belgium. Association
  for Computational Linguistics.

\bibitem[{Glava{\v{s}} and Vuli{\'c}(2021)}]{glavas:2020}
Goran Glava{\v{s}} and Ivan Vuli{\'c}. 2021.
\newblock \href {https://www.aclweb.org/anthology/2021.eacl-main.270} {Is
  supervised syntactic parsing beneficial for language understanding tasks? an
  empirical investigation}.
\newblock In \emph{Proceedings of the 16th Conference of the European Chapter
  of the Association for Computational Linguistics: Main Volume}, pages
  3090--3104, Online. Association for Computational Linguistics.

\bibitem[{Hu et~al.(2020)Hu, Ruder, Siddhant, Neubig, Firat, and
  Johnson}]{hu:2020}
Junjie Hu, Sebastian Ruder, Aditya Siddhant, Graham Neubig, Orhan Firat, and
  Melvin Johnson. 2020.
\newblock \href {http://proceedings.mlr.press/v119/hu20b.html} {{XTREME}: A
  massively multilingual multi-task benchmark for evaluating cross-lingual
  generalisation}.
\newblock In \emph{Proceedings of the 37th International Conference on Machine
  Learning}, pages 4411--4421, Virtual. PMLR.

\bibitem[{Joshi et~al.(2020)Joshi, Santy, Budhiraja, Bali, and
  Choudhury}]{joshi-etal-2020-state}
Pratik Joshi, Sebastin Santy, Amar Budhiraja, Kalika Bali, and Monojit
  Choudhury. 2020.
\newblock \href {https://doi.org/10.18653/v1/2020.acl-main.560} {The state and
  fate of linguistic diversity and inclusion in the {NLP} world}.
\newblock In \emph{Proceedings of the 58th Annual Meeting of the Association
  for Computational Linguistics}, pages 6282--6293, Online. Association for
  Computational Linguistics.

\bibitem[{K et~al.(2020)K, Wang, Mayhew, and Roth}]{k:2020}
Karthikeyan K, Zihan Wang, Stephen Mayhew, and Dan Roth. 2020.
\newblock \href {https://openreview.net/forum?id=HJeT3yrtDr} {Cross-lingual
  ability of multilingual {BERT:} an empirical study}.
\newblock In \emph{Proceedings of the 8th International Conference on Learning
  Representations ({ICLR})}, Addis Ababa, Ethiopia. OpenReview.net.

\bibitem[{Kingma and Ba(2015)}]{kingma:2015}
Diederik~P. Kingma and Jimmy Ba. 2015.
\newblock \href {http://arxiv.org/abs/1412.6980} {Adam: {A} method for
  stochastic optimization}.
\newblock In \emph{Proceedings of the 3rd International Conference on Learning
  Representations ({ICLR})}, San Diego, CA, USA.

\bibitem[{Kuratov and Arkhipov(2019)}]{kuratov:2019}
Yuri Kuratov and Mikhail Arkhipov. 2019.
\newblock \href {http://arxiv.org/abs/1905.07213} {Adaptation of deep
  bidirectional multilingual transformers for russian language}.
\newblock \emph{arXiv preprint}.

\bibitem[{Lauscher et~al.(2020)Lauscher, Ravishankar, Vuli{\'c}, and
  Glava{\v{s}}}]{Lauscher:2020zerohero}
Anne Lauscher, Vinit Ravishankar, Ivan Vuli{\'c}, and Goran Glava{\v{s}}. 2020.
\newblock \href {https://doi.org/10.18653/v1/2020.emnlp-main.363} {From zero to
  hero: {O}n the limitations of zero-shot language transfer with multilingual
  {T}ransformers}.
\newblock In \emph{Proceedings of the 2020 Conference on Empirical Methods in
  Natural Language Processing (EMNLP)}, pages 4483--4499, Online. Association
  for Computational Linguistics.

\bibitem[{Lee et~al.(2020)Lee, Jang, Baik, Park, and Shin}]{lee2020krbert}
Sangah Lee, Hansol Jang, Yunmee Baik, Suzi Park, and Hyopil Shin. 2020.
\newblock \href {http://arxiv.org/abs/2008.03979} {{KR}-{BERT}: A small-scale
  {K}orean-specific language model}.
\newblock \emph{arXiv preprint}.

\bibitem[{Lewis et~al.(2020)Lewis, Oguz, Rinott, Riedel, and
  Schwenk}]{lewis-etal-2020-mlqa}
Patrick Lewis, Barlas Oguz, Ruty Rinott, Sebastian Riedel, and Holger Schwenk.
  2020.
\newblock \href {https://doi.org/10.18653/v1/2020.acl-main.653} {{MLQA}:
  Evaluating cross-lingual extractive question answering}.
\newblock In \emph{Proceedings of the 58th Annual Meeting of the Association
  for Computational Linguistics}, pages 7315--7330, Online. Association for
  Computational Linguistics.

\bibitem[{Lim et~al.(2019)Lim, Kim, and Lee}]{lim:2019}
Seungyoung Lim, Myungji Kim, and Jooyoul Lee. 2019.
\newblock \href {http://arxiv.org/abs/1909.07005} {Kor{Q}u{AD}1.0: {K}orean
  {QA} dataset for machine reading comprehension}.
\newblock \emph{arXiv preprint}.

\bibitem[{Liu et~al.(2019)Liu, Ott, Goyal, Du, Joshi, Chen, Levy, Lewis,
  Zettlemoyer, and Stoyanov}]{liu:2019}
Yinhan Liu, Myle Ott, Naman Goyal, Jingfei Du, Mandar Joshi, Danqi Chen, Omer
  Levy, Mike Lewis, Luke Zettlemoyer, and Veselin Stoyanov. 2019.
\newblock \href {http://arxiv.org/abs/1907.11692} {Ro{BERT}a: A robustly
  optimized {BERT} pretraining approach}.
\newblock \emph{arXiv preprint}.

\bibitem[{Loshchilov and Hutter(2019)}]{loshchilov2018decoupled}
Ilya Loshchilov and Frank Hutter. 2019.
\newblock \href {https://openreview.net/forum?id=Bkg6RiCqY7} {Decoupled weight
  decay regularization}.
\newblock In \emph{Proceedings of the 7th International Conference on Learning
  Representations ({ICLR})}, New Orleans, LA, USA. OpenReview.net.

\bibitem[{Maas et~al.(2011)Maas, Daly, Pham, Huang, Ng, and Potts}]{maas:2011}
Andrew~L. Maas, Raymond~E. Daly, Peter~T. Pham, Dan Huang, Andrew~Y. Ng, and
  Christopher Potts. 2011.
\newblock \href {http://www.aclweb.org/anthology/P11-1015} {Learning word
  vectors for sentiment analysis}.
\newblock In \emph{Proceedings of the 49th Annual Meeting of the Association
  for Computational Linguistics: Human Language Technologies}, pages 142--150,
  Portland, Oregon, USA. Association for Computational Linguistics.

\bibitem[{Martin et~al.(2020)Martin, Muller, Ortiz~Su{\'a}rez, Dupont, Romary,
  de~la Clergerie, Seddah, and Sagot}]{martin:2020}
Louis Martin, Benjamin Muller, Pedro~Javier Ortiz~Su{\'a}rez, Yoann Dupont,
  Laurent Romary, {\'E}ric de~la Clergerie, Djam{\'e} Seddah, and Beno{\^\i}t
  Sagot. 2020.
\newblock \href {https://www.aclweb.org/anthology/2020.acl-main.645}
  {{C}amem{BERT}: {A} tasty {F}rench language model}.
\newblock In \emph{Proceedings of the 58th Annual Meeting of the Association
  for Computational Linguistics}, pages 7203--7219, Online. Association for
  Computational Linguistics.

\bibitem[{Mulcaire et~al.(2019)Mulcaire, Kasai, and
  Smith}]{mulcaire-etal-2019-polyglot}
Phoebe Mulcaire, Jungo Kasai, and Noah~A. Smith. 2019.
\newblock \href {https://doi.org/10.18653/v1/N19-1392} {Polyglot contextual
  representations improve crosslingual transfer}.
\newblock In \emph{Proceedings of the 2019 Conference of the North {A}merican
  Chapter of the Association for Computational Linguistics: Human Language
  Technologies, Volume 1 (Long and Short Papers)}, pages 3912--3918,
  Minneapolis, Minnesota. Association for Computational Linguistics.

\bibitem[{Nivre et~al.(2016)Nivre, de~Marneffe, Ginter, Goldberg, Haji{\v{c}},
  Manning, McDonald, Petrov, Pyysalo, Silveira, Tsarfaty, and
  Zeman}]{nivre:2016}
Joakim Nivre, Marie-Catherine de~Marneffe, Filip Ginter, Yoav Goldberg, Jan
  Haji{\v{c}}, Christopher~D. Manning, Ryan McDonald, Slav Petrov, Sampo
  Pyysalo, Natalia Silveira, Reut Tsarfaty, and Daniel Zeman. 2016.
\newblock \href {https://www.aclweb.org/anthology/L16-1262} {{U}niversal
  {D}ependencies v1: A multilingual treebank collection}.
\newblock In \emph{Proceedings of the Tenth International Conference on
  Language Resources and Evaluation ({LREC}'16)}, pages 1659--1666,
  Portoro{\v{z}}, Slovenia. European Language Resources Association (ELRA).

\bibitem[{Nivre et~al.(2020)Nivre, de~Marneffe, Ginter, Haji{\v{c}}, Manning,
  Pyysalo, Schuster, Tyers, and Zeman}]{nivre:2020}
Joakim Nivre, Marie-Catherine de~Marneffe, Filip Ginter, Jan Haji{\v{c}},
  Christopher~D. Manning, Sampo Pyysalo, Sebastian Schuster, Francis Tyers, and
  Daniel Zeman. 2020.
\newblock \href {https://www.aclweb.org/anthology/2020.lrec-1.497} {{U}niversal
  {D}ependencies v2: An evergrowing multilingual treebank collection}.
\newblock In \emph{Proceedings of the 12th Language Resources and Evaluation
  Conference}, pages 4034--4043, Marseille, France. European Language Resources
  Association.

\bibitem[{Nozza et~al.(2020)Nozza, Bianchi, and Hovy}]{nozza2020mask}
Debora Nozza, Federico Bianchi, and Dirk Hovy. 2020.
\newblock \href {http://arxiv.org/abs/2003.02912} {What the [{MASK}]? {M}aking
  sense of language-specific {BERT} models}.
\newblock \emph{arXiv preprint}.

\bibitem[{Pan et~al.(2017)Pan, Zhang, May, Nothman, Knight, and Ji}]{pan:2017}
Xiaoman Pan, Boliang Zhang, Jonathan May, Joel Nothman, Kevin Knight, and Heng
  Ji. 2017.
\newblock \href {https://doi.org/10.18653/v1/P17-1178} {Cross-lingual name
  tagging and linking for 282 languages}.
\newblock In \emph{Proceedings of the 55th Annual Meeting of the Association
  for Computational Linguistics (Volume 1: Long Papers)}, pages 1946--1958,
  Vancouver, Canada. Association for Computational Linguistics.

\bibitem[{Peters et~al.(2018)Peters, Neumann, Iyyer, Gardner, Clark, Lee, and
  Zettlemoyer}]{peters:2018}
Matthew Peters, Mark Neumann, Mohit Iyyer, Matt Gardner, Christopher Clark,
  Kenton Lee, and Luke Zettlemoyer. 2018.
\newblock \href {https://doi.org/10.18653/v1/N18-1202} {Deep contextualized
  word representations}.
\newblock In \emph{Proceedings of the 2018 Conference of the North {A}merican
  Chapter of the Association for Computational Linguistics: Human Language
  Technologies, Volume 1 (Long Papers)}, pages 2227--2237, New Orleans,
  Louisiana. Association for Computational Linguistics.

\bibitem[{Pfeiffer et~al.(2021)Pfeiffer, Kamath, R{\"u}ckl{\'e}, Cho, and
  Gurevych}]{pfeiffer:2020a}
Jonas Pfeiffer, Aishwarya Kamath, Andreas R{\"u}ckl{\'e}, Kyunghyun Cho, and
  Iryna Gurevych. 2021.
\newblock \href {https://www.aclweb.org/anthology/2021.eacl-main.39}
  {{A}dapter{F}usion: Non-destructive task composition for transfer learning}.
\newblock In \emph{Proceedings of the 16th Conference of the European Chapter
  of the Association for Computational Linguistics: Main Volume}, pages
  487--503, Online. Association for Computational Linguistics.

\bibitem[{Pfeiffer et~al.(2020{\natexlab{a}})Pfeiffer, R{\"u}ckl{\'e}, Poth,
  Kamath, Vuli{\'c}, Ruder, Cho, and Gurevych}]{pfeiffer:2020c}
Jonas Pfeiffer, Andreas R{\"u}ckl{\'e}, Clifton Poth, Aishwarya Kamath, Ivan
  Vuli{\'c}, Sebastian Ruder, Kyunghyun Cho, and Iryna Gurevych.
  2020{\natexlab{a}}.
\newblock \href {https://www.aclweb.org/anthology/2020.emnlp-demos.7}
  {{A}dapter{H}ub: A framework for adapting transformers}.
\newblock In \emph{Proceedings of the 2020 Conference on Empirical Methods in
  Natural Language Processing: System Demonstrations}, pages 46--54, Online.
  Association for Computational Linguistics.

\bibitem[{Pfeiffer et~al.(2020{\natexlab{b}})Pfeiffer, Vuli{\'c}, Gurevych, and
  Ruder}]{pfeiffer:2020b}
Jonas Pfeiffer, Ivan Vuli{\'c}, Iryna Gurevych, and Sebastian Ruder.
  2020{\natexlab{b}}.
\newblock \href {https://www.aclweb.org/anthology/2020.emnlp-main.617}
  {{MAD-X}: {A}n {A}dapter-{B}ased {F}ramework for {M}ulti-{T}ask
  {C}ross-{L}ingual {T}ransfer}.
\newblock In \emph{Proceedings of the 2020 Conference on Empirical Methods in
  Natural Language Processing (EMNLP)}, pages 7654--7673, Online. Association
  for Computational Linguistics.

\bibitem[{Pfeiffer et~al.(2020{\natexlab{c}})Pfeiffer, Vuli\'{c}, Gurevych, and
  Ruder}]{Pfeiffer20Gib}
Jonas Pfeiffer, Ivan Vuli\'{c}, Iryna Gurevych, and Sebastian Ruder.
  2020{\natexlab{c}}.
\newblock \href {https://arxiv.org/abs/2012.15562} {{UNKs Everywhere: Adapting
  Multilingual Language Models to New Scripts}}.
\newblock \emph{arXiv preprint}.

\bibitem[{Pires et~al.(2019)Pires, Schlinger, and Garrette}]{pires:2019}
Telmo Pires, Eva Schlinger, and Dan Garrette. 2019.
\newblock \href {https://doi.org/10.18653/v1/P19-1493} {How multilingual is
  multilingual {BERT}?}
\newblock In \emph{Proceedings of the 57th Annual Meeting of the Association
  for Computational Linguistics}, pages 4996--5001, Florence, Italy.
  Association for Computational Linguistics.

\bibitem[{Ponti et~al.(2020)Ponti, Glava{\v{s}}, Majewska, Liu, Vuli{\'c}, and
  Korhonen}]{ponti-etal-2020-xcopa}
Edoardo~Maria Ponti, Goran Glava{\v{s}}, Olga Majewska, Qianchu Liu, Ivan
  Vuli{\'c}, and Anna Korhonen. 2020.
\newblock \href {https://doi.org/10.18653/v1/2020.emnlp-main.185} {{XCOPA}: A
  multilingual dataset for causal commonsense reasoning}.
\newblock In \emph{Proceedings of the 2020 Conference on Empirical Methods in
  Natural Language Processing (EMNLP)}, pages 2362--2376, Online. Association
  for Computational Linguistics.

\bibitem[{Ponti et~al.(2019)Ponti, O{'}Horan, Berzak, Vuli{\'c}, Reichart,
  Poibeau, Shutova, and Korhonen}]{ponti-etal-2019-modeling}
Edoardo~Maria Ponti, Helen O{'}Horan, Yevgeni Berzak, Ivan Vuli{\'c}, Roi
  Reichart, Thierry Poibeau, Ekaterina Shutova, and Anna Korhonen. 2019.
\newblock \href {https://doi.org/10.1162/coli_a_00357} {Modeling language
  variation and universals: A survey on typological linguistics for natural
  language processing}.
\newblock \emph{Computational Linguistics}, 45(3):559--601.

\bibitem[{Prechelt(1998)}]{prechelt:1998}
Lutz Prechelt. 1998.
\newblock \href {https://doi.org/https://doi.org/10.1007/978-3-642-35289-8_5}
  {Early stopping-but when?}
\newblock In \emph{Neural Networks: Tricks of the Trade}, pages 55--69.
  Springer, Berlin, Germany.

\bibitem[{{Purwarianti} and {Crisdayanti}(2019)}]{purwarianti:2019}
Ayu {Purwarianti} and Ida Ayu Putu~Ari {Crisdayanti}. 2019.
\newblock \href {https://doi.org/10.1109/ICAICTA.2019.8904199} {Improving
  {B}i-{LSTM} performance for {I}ndonesian sentiment analysis using paragraph
  vector}.
\newblock In \emph{Proceedings of the 2019 International Conference of Advanced
  Informatics: Concepts, Theory and Applications (ICAICTA)}, pages 1--5,
  Yogyakarta, Indonesia. IEEE.

\bibitem[{Pyysalo et~al.(2020)Pyysalo, Kanerva, Virtanen, and
  Ginter}]{pyysalo2020wikibert}
Sampo Pyysalo, Jenna Kanerva, Antti Virtanen, and Filip Ginter. 2020.
\newblock \href {http://arxiv.org/abs/2006.01538} {Wiki{BERT} models: {D}eep
  transfer learning for many languages}.
\newblock \emph{arXiv preprint}.

\bibitem[{Raffel et~al.(2020)Raffel, Shazeer, Roberts, Lee, Narang, Matena,
  Zhou, Li, and Liu}]{Raffel:2020t5}
Colin Raffel, Noam Shazeer, Adam Roberts, Katherine Lee, Sharan Narang, Michael
  Matena, Yanqi Zhou, Wei Li, and Peter~J. Liu. 2020.
\newblock \href {http://jmlr.org/papers/v21/20-074.html} {Exploring the limits
  of transfer learning with a unified text-to-text transformer}.
\newblock \emph{Journal of Machine Learning Research}, 21(140):1--67.

\bibitem[{Rahimi et~al.(2019)Rahimi, Li, and Cohn}]{rahimi:2019}
Afshin Rahimi, Yuan Li, and Trevor Cohn. 2019.
\newblock \href {https://doi.org/10.18653/v1/P19-1015} {Massively multilingual
  transfer for {NER}}.
\newblock In \emph{Proceedings of the 57th Annual Meeting of the Association
  for Computational Linguistics}, pages 151--164, Florence, Italy. Association
  for Computational Linguistics.

\bibitem[{Rajpurkar et~al.(2016)Rajpurkar, Zhang, Lopyrev, and
  Liang}]{rajpurkar:2016}
Pranav Rajpurkar, Jian Zhang, Konstantin Lopyrev, and Percy Liang. 2016.
\newblock \href {https://doi.org/10.18653/v1/D16-1264} {{SQ}u{AD}: 100,000+
  questions for machine comprehension of text}.
\newblock In \emph{Proceedings of the 2016 Conference on Empirical Methods in
  Natural Language Processing}, pages 2383--2392, Austin, Texas. Association
  for Computational Linguistics.

\bibitem[{Ruokolainen et~al.(2020)Ruokolainen, Kauppinen, Silfverberg, and
  Lind{\'e}n}]{ruokolainen:2019}
Teemu Ruokolainen, Pekka Kauppinen, Miikka Silfverberg, and Krister Lind{\'e}n.
  2020.
\newblock \href {https://doi.org/10.1007/s10579-019-09471-7} {A {F}innish news
  corpus for named entity recognition}.
\newblock \emph{Language Resources and Evaluation}, 54(1):247--272.

\bibitem[{Rönnqvist et~al.(2019)Rönnqvist, Kanerva, Salakoski, and
  Ginter}]{ronnqvist:2019}
Samuel Rönnqvist, Jenna Kanerva, Tapio Salakoski, and Filip Ginter. 2019.
\newblock \href {https://www.aclweb.org/anthology/W19-6204} {Is multilingual
  {BERT} fluent in language generation?}
\newblock In \emph{Proceedings of the First NLPL Workshop on Deep Learning for
  Natural Language Processing}, pages 29--36, Turku, Finland. Linköping
  University Electronic Press.

\bibitem[{Schweter(2020)}]{schweter:2020}
Stefan Schweter. 2020.
\newblock \href {https://doi.org/10.5281/zenodo.3770924} {{BERT}urk - {BERT}
  models for {T}urkish}.
\newblock Zenodo.

\bibitem[{Shao et~al.(2019)Shao, Liu, Lai, Tseng, and Tsai}]{shao:2019}
Chih~Chieh Shao, Trois Liu, Yuting Lai, Yiying Tseng, and Sam Tsai. 2019.
\newblock \href {http://arxiv.org/abs/1806.00920} {{DRCD}: a {C}hinese machine
  reading comprehension dataset}.
\newblock \emph{arXiv preprint}.

\bibitem[{Smetanin and Komarov(2019)}]{smetanin:2019}
Sergey Smetanin and Michail Komarov. 2019.
\newblock \href {https://doi.org/10.1109/CBI.2019.00062} {Sentiment analysis of
  product reviews in {R}ussian using convolutional neural networks}.
\newblock In \emph{Proceedings of the 2019 IEEE 21st Conference on Business
  Informatics (CBI)}, pages 482--486, Moscow, Russia. IEEE.

\bibitem[{Straka et~al.(2016)Straka, Haji{\v{c}}, and
  Strakov{\'a}}]{straka-etal-2016-udpipe}
Milan Straka, Jan Haji{\v{c}}, and Jana Strakov{\'a}. 2016.
\newblock \href {https://www.aclweb.org/anthology/L16-1680} {{UDP}ipe:
  Trainable pipeline for processing {C}o{NLL}-{U} files performing
  tokenization, morphological analysis, {POS} tagging and parsing}.
\newblock In \emph{Proceedings of the Tenth International Conference on
  Language Resources and Evaluation ({LREC}'16)}, pages 4290--4297,
  Portoro{\v{z}}, Slovenia. European Language Resources Association (ELRA).

\bibitem[{Tjong Kim~Sang and De~Meulder(2003)}]{sang:2003}
Erik~F. Tjong Kim~Sang and Fien De~Meulder. 2003.
\newblock \href {https://www.aclweb.org/anthology/W03-0419} {Introduction to
  the {C}o{NLL}-2003 shared task: Language-independent named entity
  recognition}.
\newblock In \emph{Proceedings of the Seventh Conference on Natural Language
  Learning at {HLT}-{NAACL} 2003}, pages 142--147, Edmonton, Canada.
  Association for Computational Linguistics.

\bibitem[{{\"U}st{\"u}n et~al.(2020){\"U}st{\"u}n, Bisazza, Bouma, and van
  Noord}]{ustun-etal-2020-udapter}
Ahmet {\"U}st{\"u}n, Arianna Bisazza, Gosse Bouma, and Gertjan van Noord. 2020.
\newblock \href {https://doi.org/10.18653/v1/2020.emnlp-main.180} {{UD}apter:
  Language adaptation for truly {U}niversal {D}ependency parsing}.
\newblock In \emph{Proceedings of the 2020 Conference on Empirical Methods in
  Natural Language Processing (EMNLP)}, pages 2302--2315, Online. Association
  for Computational Linguistics.

\bibitem[{Vaswani et~al.(2017)Vaswani, Shazeer, Parmar, Uszkoreit, Jones,
  Gomez, Kaiser, and Polosukhin}]{Vaswani:2017}
Ashish Vaswani, Noam Shazeer, Niki Parmar, Jakob Uszkoreit, Llion Jones,
  Aidan~N Gomez, {\L}ukasz Kaiser, and Illia Polosukhin. 2017.
\newblock \href
  {https://proceedings.neurips.cc/paper/2017/file/3f5ee243547dee91fbd053c1c4a845aa-Paper.pdf}
  {Attention is all you need}.
\newblock In \emph{Advances in Neural Information Processing Systems}, pages
  5998--6008, Long Beach, CA, USA. Curran Associates, Inc.

\bibitem[{Virtanen et~al.(2019)Virtanen, Kanerva, Ilo, Luoma, Luotolahti,
  Salakoski, Ginter, and Pyysalo}]{virtanen:2019}
Antti Virtanen, Jenna Kanerva, Rami Ilo, Jouni Luoma, Juhani Luotolahti, Tapio
  Salakoski, Filip Ginter, and Sampo Pyysalo. 2019.
\newblock \href {http://arxiv.org/abs/1912.07076} {Multilingual is not enough:
  {BERT} for {F}innish}.
\newblock \emph{arXiv preprint}.

\bibitem[{de~Vries et~al.(2019)de~Vries, van Cranenburgh, Bisazza, Caselli, van
  Noord, and Nissim}]{vries:2019}
Wietse de~Vries, Andreas van Cranenburgh, Arianna Bisazza, Tommaso Caselli,
  Gertjan van Noord, and Malvina Nissim. 2019.
\newblock \href {http://arxiv.org/abs/1912.09582} {{BERT}je: {A} {D}utch {BERT}
  {M}odel}.
\newblock \emph{arXiv preprint}.

\bibitem[{Vuli{\'c} et~al.(2020)Vuli{\'c}, Ponti, Litschko, Glava{\v{s}}, and
  Korhonen}]{vulic-etal-2020-probing}
Ivan Vuli{\'c}, Edoardo~Maria Ponti, Robert Litschko, Goran Glava{\v{s}}, and
  Anna Korhonen. 2020.
\newblock \href {https://doi.org/10.18653/v1/2020.emnlp-main.586} {Probing
  pretrained language models for lexical semantics}.
\newblock In \emph{Proceedings of the 2020 Conference on Empirical Methods in
  Natural Language Processing (EMNLP)}, pages 7222--7240, Online. Association
  for Computational Linguistics.

\bibitem[{Wilie et~al.(2020)Wilie, Vincentio, Winata, Cahyawijaya, Li, Lim,
  Soleman, Mahendra, Fung, Bahar, and Purwarianti}]{wilie-etal-2020-indonlu}
Bryan Wilie, Karissa Vincentio, Genta~Indra Winata, Samuel Cahyawijaya,
  Xiaohong Li, Zhi~Yuan Lim, Sidik Soleman, Rahmad Mahendra, Pascale Fung,
  Syafri Bahar, and Ayu Purwarianti. 2020.
\newblock \href {https://www.aclweb.org/anthology/2020.aacl-main.85}
  {{I}ndo{NLU}: Benchmark and resources for evaluating {I}ndonesian natural
  language understanding}.
\newblock In \emph{Proceedings of the 1st Conference of the Asia-Pacific
  Chapter of the Association for Computational Linguistics and the 10th
  International Joint Conference on Natural Language Processing}, pages
  843--857, Suzhou, China. Association for Computational Linguistics.

\bibitem[{Wolf et~al.(2020)Wolf, Debut, Sanh, Chaumond, Delangue, Moi, Cistac,
  Rault, Louf, Funtowicz, Davison, Shleifer, von Platen, Ma, Jernite, Plu, Xu,
  Le~Scao, Gugger, Drame, Lhoest, and Rush}]{wolf:2020}
Thomas Wolf, Lysandre Debut, Victor Sanh, Julien Chaumond, Clement Delangue,
  Anthony Moi, Pierric Cistac, Tim Rault, Remi Louf, Morgan Funtowicz, Joe
  Davison, Sam Shleifer, Patrick von Platen, Clara Ma, Yacine Jernite, Julien
  Plu, Canwen Xu, Teven Le~Scao, Sylvain Gugger, Mariama Drame, Quentin Lhoest,
  and Alexander Rush. 2020.
\newblock \href {https://doi.org/10.18653/v1/2020.emnlp-demos.6} {Transformers:
  State-of-the-art natural language processing}.
\newblock In \emph{Proceedings of the 2020 Conference on Empirical Methods in
  Natural Language Processing: System Demonstrations}, pages 38--45, Online.
  Association for Computational Linguistics.

\bibitem[{Wu and Dredze(2019)}]{wu-dredze-2019-beto}
Shijie Wu and Mark Dredze. 2019.
\newblock \href {https://doi.org/10.18653/v1/D19-1077} {Beto, bentz, becas: The
  surprising cross-lingual effectiveness of {BERT}}.
\newblock In \emph{Proceedings of the 2019 Conference on Empirical Methods in
  Natural Language Processing and the 9th International Joint Conference on
  Natural Language Processing (EMNLP-IJCNLP)}, pages 833--844, Hong Kong,
  China. Association for Computational Linguistics.

\bibitem[{Wu and Dredze(2020)}]{wu-dredze-2020-languages}
Shijie Wu and Mark Dredze. 2020.
\newblock \href {https://doi.org/10.18653/v1/2020.repl4nlp-1.16} {Are all
  languages created equal in multilingual {BERT}?}
\newblock In \emph{Proceedings of the 5th Workshop on Representation Learning
  for NLP}, pages 120--130, Online. Association for Computational Linguistics.

\bibitem[{Wu et~al.(2016)Wu, Schuster, Chen, Le, Norouzi, Macherey, Krikun,
  Cao, Gao, Macherey, Klingner, Shah, Johnson, Liu, Kaiser, Gouws, Kato, Kudo,
  Kazawa, Stevens, Kurian, Patil, Wang, Young, Smith, Riesa, Rudnick, Vinyals,
  Corrado, Hughes, and Dean}]{Wu:2016wp}
Yonghui Wu, Mike Schuster, Zhifeng Chen, Quoc~V. Le, Mohammad Norouzi, Wolfgang
  Macherey, Maxim Krikun, Yuan Cao, Qin Gao, Klaus Macherey, Jeff Klingner,
  Apurva Shah, Melvin Johnson, Xiaobing Liu, Lukasz Kaiser, Stephan Gouws,
  Yoshikiyo Kato, Taku Kudo, Hideto Kazawa, Keith Stevens, George Kurian,
  Nishant Patil, Wei Wang, Cliff Young, Jason Smith, Jason Riesa, Alex Rudnick,
  Oriol Vinyals, Greg Corrado, Macduff Hughes, and Jeffrey Dean. 2016.
\newblock \href {http://arxiv.org/abs/1609.08144} {Google's neural machine
  translation system: {B}ridging the gap between human and machine
  translation}.
\newblock \emph{arxiv preprint}.

\bibitem[{Xu et~al.(2017)Xu, Wen, Sun, and Su}]{xu:2017}
Jingjing Xu, Ji~Wen, Xu~Sun, and Qi~Su. 2017.
\newblock \href {http://arxiv.org/abs/1711.07010} {A discourse-level named
  entity recognition and relation extraction dataset for {C}hinese literature
  text}.
\newblock \emph{arXiv preprint}.

\bibitem[{Xue et~al.(2021)Xue, Constant, Roberts, Kale, Al-Rfou, Siddhant,
  Barua, and Raffel}]{xue2020mt5}
Linting Xue, Noah Constant, Adam Roberts, Mihir Kale, Rami Al-Rfou, Aditya
  Siddhant, Aditya Barua, and Colin Raffel. 2021.
\newblock \href {https://www.aclweb.org/anthology/2021.naacl-main.41} {m{T}5: A
  massively multilingual pre-trained text-to-text transformer}.
\newblock In \emph{Proceedings of the 2021 Conference of the North American
  Chapter of the Association for Computational Linguistics: Human Language
  Technologies}, pages 483--498, Online. Association for Computational
  Linguistics.

\bibitem[{Zeman et~al.(2020)Zeman, Nivre, Abrams, Ackermann, Aepli, Agi{\'c},
  Ahrenberg, Ajede, Aleksandravi{\v c}i{\=u}t{\.e}, Antonsen, Aplonova, Aquino,
  Aranzabe, Arutie, Asahara, Ateyah, Atmaca, Attia, Atutxa, Augustinus,
  Badmaeva, Ballesteros, Banerjee, Bank, Barbu~Mititelu, Basmov, Batchelor,
  Bauer, Bengoetxea, Berzak, Bhat, Bhat, Biagetti, Bick, Bielinskien{\.e},
  Blokland, Bobicev, Boizou, Borges~V{\"o}lker, B{\"o}rstell, Bosco, Bouma,
  Bowman, Boyd, Brokait{\.e}, Burchardt, Candito, Caron, Caron, Cavalcanti,
  Cebiro{\u g}lu~Eryi{\u g}it, Cecchini, Celano, {\v C}{\'e}pl{\"o}, Cetin,
  Chalub, Chi, Choi, Cho, Chun, Cignarella, Cinkov{\'a}, Collomb, {\c
  C}{\"o}ltekin, Connor, Courtin, Davidson, de~Marneffe, de~Paiva, de~Souza,
  Diaz~de Ilarraza, Dickerson, Dione, Dirix, Dobrovoljc, Dozat, Droganova,
  Dwivedi, Eckhoff, Eli, Elkahky, Ephrem, Erina, Erjavec, Etienne, Evelyn,
  Farkas, Fernandez~Alcalde, Foster, Freitas, Fujita, Gajdo{\v s}ov{\'a},
  Galbraith, Garcia, G{\"a}rdenfors, Garza, Gerdes, Ginter, Goenaga, Gojenola,
  G{\"o}k{\i}rmak, Goldberg, G{\'o}mez~Guinovart, Gonz{\'a}lez~Saavedra,
  Grici{\=u}t{\.e}, Grioni, Grobol, Gr{\= u}z{\={\i}}tis, Guillaume,
  Guillot-Barbance, G{\"u}ng{\"o}r, Habash, Haji{\v c}, Haji{\v c}~jr.,
  H{\"a}m{\"a}l{\"a}inen, H{\`a}~M{\~y}, Han, Harris, Haug, Heinecke, Hellwig,
  Hennig, Hladk{\'a}, Hlav{\'a}{\v c}ov{\'a}, Hociung, Hohle, Hwang, Ikeda,
  Ion, Irimia, Ishola, Jel{\'{\i}}nek, Johannsen, J{\'o}nsd{\'o}ttir,
  J{\o}rgensen, Juutinen, Ka{\c s}{\i}kara, Kaasen, Kabaeva, Kahane, Kanayama,
  Kanerva, Katz, Kayadelen, Kenney, Kettnerov{\'a}, Kirchner, Klementieva,
  K{\"o}hn, K{\"o}ksal, Kopacewicz, Korkiakangas, Kotsyba, Kovalevskait{\.e},
  Krek, Kwak, Laippala, Lambertino, Lam, Lando, Larasati, Lavrentiev, Lee,
  L{\^e}~H{\`{\^o}}ng, Lenci, Lertpradit, Leung, Levina, Li, Li, Li, Lim, Li,
  Ljube{\v s}i{\'c}, Loginova, Lyashevskaya, Lynn, Macketanz, Makazhanov,
  Mandl, Manning, Manurung, M{\u a}r{\u a}nduc, Mare{\v c}ek, Marheinecke,
  Mart{\'{\i}}nez~Alonso, Martins, Ma{\v s}ek, Matsuda, Matsumoto, {McDonald},
  {McGuinness}, Mendon{\c c}a, Miekka, Misirpashayeva, Missil{\"a}, Mititelu,
  Mitrofan, Miyao, Montemagni, More, Moreno~Romero, Mori, Morioka, Mori, Moro,
  Mortensen, Moskalevskyi, Muischnek, Munro, Murawaki, M{\"u}{\"u}risep,
  Nainwani, Navarro~Hor{\~n}iacek, Nedoluzhko, Ne{\v s}pore-B{\=e}rzkalne,
  Nguy{\~{\^e}}n~Th{\d i}, Nguy{\~{\^e}}n Th{\d i}~Minh, Nikaido, Nikolaev,
  Nitisaroj, Nurmi, Ojala, Ojha, Ol{\'u}{\`o}kun, Omura, Onwuegbuzia, Osenova,
  {\"O}stling, {\O}vrelid, {\"O}zate{\c s}, {\"O}zg{\"u}r, {\"O}zt{\"u}rk~Ba{\c
  s}aran, Partanen, Pascual, Passarotti, Patejuk, Paulino-Passos,
  Peljak-{\L}api{\'n}ska, Peng, Perez, Perrier, Petrova, Petrov, Phelan,
  Piitulainen, Pirinen, Pitler, Plank, Poibeau, Ponomareva, Popel, Pretkalni{\c
  n}a, Pr{\'e}vost, Prokopidis, Przepi{\'o}rkowski, Puolakainen, Pyysalo, Qi,
  R{\"a}{\"a}bis, Rademaker, Ramasamy, Rama, Ramisch, Ravishankar, Real,
  Rebeja, Reddy, Rehm, Riabov, Rie{\ss}ler, Rimkut{\.e}, Rinaldi, Rituma,
  Rocha, Romanenko, Rosa, Roșca, Rovati, Rudina, Rueter, Sadde, Sagot, Saleh,
  Salomoni, Samard{\v z}i{\'c}, Samson, Sanguinetti, S{\"a}rg, Saul{\={\i}}te,
  Sawanakunanon, Scarlata, Schneider, Schuster, Seddah, Seeker, Seraji, Shen,
  Shimada, Shirasu, Shohibussirri, Sichinava, Silveira, Silveira, Simi,
  Simionescu, Simk{\'o}, {\v S}imkov{\'a}, Simov, Skachedubova, Smith,
  Soares-Bastos, Spadine, Stella, Straka, Strnadov{\'a}, Suhr, Sulubacak,
  Suzuki, Sz{\'a}nt{\'o}, Taji, Takahashi, Tamburini, Tanaka, Tella, Tellier,
  Thomas, Torga, Toska, Trosterud, Trukhina, Tsarfaty, T{\"u}rk, Tyers,
  Uematsu, Untilov, Ure{\v s}ov{\'a}, Uria, Uszkoreit, Utka, Vajjala, van
  Niekerk, van Noord, Varga, Villemonte de~la Clergerie, Vincze, Wakasa,
  Wallin, Walsh, Wang, Washington, Wendt, Widmer, Williams, Wir{\'e}n, Wittern,
  Woldemariam, Wong, Wr{\'o}blewska, Yako, Yamashita, Yamazaki, Yan, Yasuoka,
  Yavrumyan, Yu, {\v Z}abokrtsk{\'y}, Zeldes, Zhu, and Zhuravleva}]{zeman:2020}
Daniel Zeman, Joakim Nivre, Mitchell Abrams, Elia Ackermann, No{\"e}mi Aepli,
  {\v Z}eljko Agi{\'c}, Lars Ahrenberg, Chika~Kennedy Ajede, Gabriel{\.e}
  Aleksandravi{\v c}i{\=u}t{\.e}, Lene Antonsen, Katya Aplonova, Angelina
  Aquino, Maria~Jesus Aranzabe, Gashaw Arutie, Masayuki Asahara, Luma Ateyah,
  Furkan Atmaca, Mohammed Attia, Aitziber Atutxa, Liesbeth Augustinus, Elena
  Badmaeva, Miguel Ballesteros, Esha Banerjee, Sebastian Bank, Verginica
  Barbu~Mititelu, Victoria Basmov, Colin Batchelor, John Bauer, Kepa
  Bengoetxea, Yevgeni Berzak, Irshad~Ahmad Bhat, Riyaz~Ahmad Bhat, Erica
  Biagetti, Eckhard Bick, Agn{\.e} Bielinskien{\.e}, Rogier Blokland, Victoria
  Bobicev, Lo{\"{\i}}c Boizou, Emanuel Borges~V{\"o}lker, Carl B{\"o}rstell,
  Cristina Bosco, Gosse Bouma, Sam Bowman, Adriane Boyd, Kristina Brokait{\.e},
  Aljoscha Burchardt, Marie Candito, Bernard Caron, Gauthier Caron, Tatiana
  Cavalcanti, G{\"u}l{\c s}en Cebiro{\u g}lu~Eryi{\u g}it, Flavio~Massimiliano
  Cecchini, Giuseppe G.~A. Celano, Slavom{\'{\i}}r {\v C}{\'e}pl{\"o}, Savas
  Cetin, Fabricio Chalub, Ethan Chi, Jinho Choi, Yongseok Cho, Jayeol Chun,
  Alessandra~T. Cignarella, Silvie Cinkov{\'a}, Aur{\'e}lie Collomb, {\c C}a{\u
  g}r{\i} {\c C}{\"o}ltekin, Miriam Connor, Marine Courtin, Elizabeth Davidson,
  Marie-Catherine de~Marneffe, Valeria de~Paiva, Elvis de~Souza, Arantza
  Diaz~de Ilarraza, Carly Dickerson, Bamba Dione, Peter Dirix, Kaja Dobrovoljc,
  Timothy Dozat, Kira Droganova, Puneet Dwivedi, Hanne Eckhoff, Marhaba Eli,
  Ali Elkahky, Binyam Ephrem, Olga Erina, Toma{\v z} Erjavec, Aline Etienne,
  Wograine Evelyn, Rich{\'a}rd Farkas, Hector Fernandez~Alcalde, Jennifer
  Foster, Cl{\'a}udia Freitas, Kazunori Fujita, Katar{\'{\i}}na Gajdo{\v
  s}ov{\'a}, Daniel Galbraith, Marcos Garcia, Moa G{\"a}rdenfors, Sebastian
  Garza, Kim Gerdes, Filip Ginter, Iakes Goenaga, Koldo Gojenola, Memduh
  G{\"o}k{\i}rmak, Yoav Goldberg, Xavier G{\'o}mez~Guinovart, Berta
  Gonz{\'a}lez~Saavedra, Bernadeta Grici{\=u}t{\.e}, Matias Grioni, Lo{\"{\i}}c
  Grobol, Normunds Gr{\= u}z{\={\i}}tis, Bruno Guillaume, C{\'e}line
  Guillot-Barbance, Tunga G{\"u}ng{\"o}r, Nizar Habash, Jan Haji{\v c}, Jan
  Haji{\v c}~jr., Mika H{\"a}m{\"a}l{\"a}inen, Linh H{\`a}~M{\~y}, Na-Rae Han,
  Kim Harris, Dag Haug, Johannes Heinecke, Oliver Hellwig, Felix Hennig,
  Barbora Hladk{\'a}, Jaroslava Hlav{\'a}{\v c}ov{\'a}, Florinel Hociung,
  Petter Hohle, Jena Hwang, Takumi Ikeda, Radu Ion, Elena Irimia, {\d
  O}l{\'a}j{\'{\i}}d{\'e} Ishola, Tom{\'a}{\v s} Jel{\'{\i}}nek, Anders
  Johannsen, Hildur J{\'o}nsd{\'o}ttir, Fredrik J{\o}rgensen, Markus Juutinen,
  H{\"u}ner Ka{\c s}{\i}kara, Andre Kaasen, Nadezhda Kabaeva, Sylvain Kahane,
  Hiroshi Kanayama, Jenna Kanerva, Boris Katz, Tolga Kayadelen, Jessica Kenney,
  V{\'a}clava Kettnerov{\'a}, Jesse Kirchner, Elena Klementieva, Arne K{\"o}hn,
  Abdullatif K{\"o}ksal, Kamil Kopacewicz, Timo Korkiakangas, Natalia Kotsyba,
  Jolanta Kovalevskait{\.e}, Simon Krek, Sookyoung Kwak, Veronika Laippala,
  Lorenzo Lambertino, Lucia Lam, Tatiana Lando, Septina~Dian Larasati, Alexei
  Lavrentiev, John Lee, Phuong L{\^e}~H{\`{\^o}}ng, Alessandro Lenci, Saran
  Lertpradit, Herman Leung, Maria Levina, Cheuk~Ying Li, Josie Li, Keying Li,
  {KyungTae} Lim, Yuan Li, Nikola Ljube{\v s}i{\'c}, Olga Loginova, Olga
  Lyashevskaya, Teresa Lynn, Vivien Macketanz, Aibek Makazhanov, Michael Mandl,
  Christopher Manning, Ruli Manurung, C{\u a}t{\u a}lina M{\u a}r{\u a}nduc,
  David Mare{\v c}ek, Katrin Marheinecke, H{\'e}ctor Mart{\'{\i}}nez~Alonso,
  Andr{\'e} Martins, Jan Ma{\v s}ek, Hiroshi Matsuda, Yuji Matsumoto, Ryan
  {McDonald}, Sarah {McGuinness}, Gustavo Mendon{\c c}a, Niko Miekka, Margarita
  Misirpashayeva, Anna Missil{\"a}, C{\u a}t{\u a}lin Mititelu, Maria Mitrofan,
  Yusuke Miyao, Simonetta Montemagni, Amir More, Laura Moreno~Romero,
  Keiko~Sophie Mori, Tomohiko Morioka, Shinsuke Mori, Shigeki Moro, Bjartur
  Mortensen, Bohdan Moskalevskyi, Kadri Muischnek, Robert Munro, Yugo Murawaki,
  Kaili M{\"u}{\"u}risep, Pinkey Nainwani, Juan~Ignacio Navarro~Hor{\~n}iacek,
  Anna Nedoluzhko, Gunta Ne{\v s}pore-B{\=e}rzkalne, Luong Nguy{\~{\^e}}n~Th{\d
  i}, Huy{\`{\^e}}n Nguy{\~{\^e}}n Th{\d i}~Minh, Yoshihiro Nikaido, Vitaly
  Nikolaev, Rattima Nitisaroj, Hanna Nurmi, Stina Ojala, Atul~Kr. Ojha,
  Ad{\'e}day{\d o} Ol{\'u}{\`o}kun, Mai Omura, Emeka Onwuegbuzia, Petya
  Osenova, Robert {\"O}stling, Lilja {\O}vrelid, {\c S}aziye~Bet{\"u}l
  {\"O}zate{\c s}, Arzucan {\"O}zg{\"u}r, Balk{\i}z {\"O}zt{\"u}rk~Ba{\c
  s}aran, Niko Partanen, Elena Pascual, Marco Passarotti, Agnieszka Patejuk,
  Guilherme Paulino-Passos, Angelika Peljak-{\L}api{\'n}ska, Siyao Peng,
  Cenel-Augusto Perez, Guy Perrier, Daria Petrova, Slav Petrov, Jason Phelan,
  Jussi Piitulainen, Tommi~A Pirinen, Emily Pitler, Barbara Plank, Thierry
  Poibeau, Larisa Ponomareva, Martin Popel, Lauma Pretkalni{\c n}a, Sophie
  Pr{\'e}vost, Prokopis Prokopidis, Adam Przepi{\'o}rkowski, Tiina Puolakainen,
  Sampo Pyysalo, Peng Qi, Andriela R{\"a}{\"a}bis, Alexandre Rademaker,
  Loganathan Ramasamy, Taraka Rama, Carlos Ramisch, Vinit Ravishankar, Livy
  Real, Petru Rebeja, Siva Reddy, Georg Rehm, Ivan Riabov, Michael Rie{\ss}ler,
  Erika Rimkut{\.e}, Larissa Rinaldi, Laura Rituma, Luisa Rocha, Mykhailo
  Romanenko, Rudolf Rosa, Valentin Roșca, Davide Rovati, Olga Rudina, Jack
  Rueter, Shoval Sadde, Beno{\^{\i}}t Sagot, Shadi Saleh, Alessio Salomoni,
  Tanja Samard{\v z}i{\'c}, Stephanie Samson, Manuela Sanguinetti, Dage
  S{\"a}rg, Baiba Saul{\={\i}}te, Yanin Sawanakunanon, Salvatore Scarlata,
  Nathan Schneider, Sebastian Schuster, Djam{\'e} Seddah, Wolfgang Seeker,
  Mojgan Seraji, Mo~Shen, Atsuko Shimada, Hiroyuki Shirasu, Muh Shohibussirri,
  Dmitry Sichinava, Aline Silveira, Natalia Silveira, Maria Simi, Radu
  Simionescu, Katalin Simk{\'o}, M{\'a}ria {\v S}imkov{\'a}, Kiril Simov, Maria
  Skachedubova, Aaron Smith, Isabela Soares-Bastos, Carolyn Spadine, Antonio
  Stella, Milan Straka, Jana Strnadov{\'a}, Alane Suhr, Umut Sulubacak, Shingo
  Suzuki, Zsolt Sz{\'a}nt{\'o}, Dima Taji, Yuta Takahashi, Fabio Tamburini,
  Takaaki Tanaka, Samson Tella, Isabelle Tellier, Guillaume Thomas, Liisi
  Torga, Marsida Toska, Trond Trosterud, Anna Trukhina, Reut Tsarfaty, Utku
  T{\"u}rk, Francis Tyers, Sumire Uematsu, Roman Untilov, Zde{\v n}ka Ure{\v
  s}ov{\'a}, Larraitz Uria, Hans Uszkoreit, Andrius Utka, Sowmya Vajjala,
  Daniel van Niekerk, Gertjan van Noord, Viktor Varga, Eric Villemonte de~la
  Clergerie, Veronika Vincze, Aya Wakasa, Lars Wallin, Abigail Walsh, Jing~Xian
  Wang, Jonathan~North Washington, Maximilan Wendt, Paul Widmer, Seyi Williams,
  Mats Wir{\'e}n, Christian Wittern, Tsegay Woldemariam, Tak-sum Wong, Alina
  Wr{\'o}blewska, Mary Yako, Kayo Yamashita, Naoki Yamazaki, Chunxiao Yan,
  Koichi Yasuoka, Marat~M. Yavrumyan, Zhuoran Yu, Zden{\v e}k {\v
  Z}abokrtsk{\'y}, Amir Zeldes, Hanzhi Zhu, and Anna Zhuravleva. 2020.
\newblock \href {http://hdl.handle.net/11234/1-3226} {Universal {D}ependencies
  2.6}.
\newblock {LINDAT}/{CLARIAH}-{CZ} digital library at the Institute of Formal
  and Applied Linguistics ({{\'U}FAL}), Faculty of Mathematics and Physics,
  Charles University.

\bibitem[{Zhang et~al.(2020)Zhang, Warstadt, Li, and Bowman}]{zhang2020need}
Yian Zhang, Alex Warstadt, Haau-Sing Li, and Samuel~R. Bowman. 2020.
\newblock \href {http://arxiv.org/abs/2011.04946} {When do you need billions of
  words of pretraining data?}
\newblock \emph{arXiv preprint}.

\end{thebibliography}

\clearpage
\appendix

\newpage

\section{Reproducibility}

\subsection{Pretrained Models}
\label{a:model_selection}

All of the pretrained language models we use are available on the HuggingFace model hub\footnote{\href{https://huggingface.co/models}{https://huggingface.co/models}} and compatible with the HuggingFace transformers Python library \cite{wolf:2020}. Table~\ref{tab:vocab_sizes} displays the model hub identifiers of our selected models.

\subsection{Estimating the Pretraining Corpora Sizes}
\label{a:data_size_estimation}

Since mBERT was pretrained on the entire Wikipedia dumps of all languages it covers \cite{devlin:2019}, we estimate the language-specific shares of the mBERT pretraining corpus by word counts of the respective raw Wikipedia dumps, according to numbers obtained from Wikimedia\footnote{\label{fn:wikimedia}\href{https://meta.m.wikimedia.org/wiki/List_of_Wikipedias}{https://meta.m.wikimedia.org/wiki/List\_of\_Wikipedias}}: 327M words for \textsc{ar}, 3.7B for \textsc{en}, 134M for \textsc{fi}, 142M for \textsc{id}, 1.1B for \textsc{ja}, 125M for \textsc{ko}, 781M for \textsc{ru}, 104M for \textsc{tr}, 482M for \textsc{zh}.\footnote{We obtained the numbers for \textsc{id} and \textsc{tr} on Dec 10, 2020 and for the remaining languages on Sep 10, 2020.} \citet{devlin:2019} only included text passages from the articles, and used older Wikipedia dumps, so these numbers should serve as upper limits, yet be reasonably accurate. For the monolingual models, we rely on information provided by the authors.\footnote{For \textsc{ja}, \textsc{ru}, and \textsc{zh}, the authors do not provide exact word counts. Therefore, we estimate them using other provided information (\textsc{ru}, \textsc{zh}) or scripts for training corpus reconstruction (\textsc{ja}).} 

\subsection{Data for Tokenizer Analyses}
\label{a:ud_tokenizer_data}

We tokenize the training and development splits of the UD \cite{nivre:2016, nivre:2020} v2.6 \cite{zeman:2020} treebanks listed in Table~\ref{tab:ud_data_tokenizers}.

\subsection{Fine-Tuning Datasets}
\label{a:data_preprocessing}
We list the datasets we used, including the number of examples per dataset split, in the Table~\ref{tab:ner_data}.

\subsection{Training Procedure of New Models} 
\label{sec:app:training_new_models}
We pretrain our models on single Nvidia Tesla V100, A100, and Titan RTX GPUs with 32GB, 40GB, and 24GB of video memory, respectively. To support larger batch sizes, we train in mixed-precision (fp16) mode. Following \citet{wu-dredze-2020-languages}, we only use masked language modeling (MLM) as pretraining objective and omit the next sentence prediction task as \citet{liu:2019} find it does not yield performance gains. We otherwise mostly follow the default pretraining procedure by \citet{devlin:2019}. \\
We pretrain the new monolingual models ({\footnotesize\textsc{MonoModel-*}}) from scratch for 1M steps with batch size 64. We choose a sequence length of 128 for the first 900,000 steps and 512 for the remaining 100,000 steps. In both phases, we warm up the learning rate to $1e-4$ over the first 10,000 steps, then decay linearly. We use the Adam optimizer with weight decay (AdamW) \cite{loshchilov2018decoupled} with default hyper-parameters and a weight decay of 0.01. 
We enable whole word masking \cite{devlin:2019} for the \textsc{fi} monolingual models, following the pretraining procedure for FinBERT \cite{virtanen:2019}. To lower computational requirements for the monolingual models with mBERT tokenizers, we remove all tokens from mBERT's vocabulary that do not appear in the pretraining data. We, thereby, obtain vocabularies of size 78,193 (\textsc{ar}), 60,827 (\textsc{fi}), 72,787 (\textsc{id}), 66,268 (\textsc{ko}), and 71,007 (\textsc{tr}),
which for all languages reduces the number of parameters in the embedding layer significantly, compared to the 119,547 word piece vocabulary of mBERT. \\
For the retrained mBERT models (i.e., {\footnotesize\textsc{mBERTModel-*}}), we run MLM for 250,000 steps (similar to \citet{artetxe-etal-2020-cross}) with batch size 64 and sequence length 512, otherwise using the same hyper-parameters as for the monolingual models. In order to retrain the embedding layer, we first resize it to match the vocabulary of the respective tokenizer. For the {\footnotesize\textsc{mBERTModel-mBERTTok}} models, we use the mBERT tokenizers with reduced vocabulary as outlined above. We initialize the positional embeddings, segment embeddings, and embeddings of special tokens ({\footnotesize\texttt{[CLS]}}, {\footnotesize\texttt{[SEP]}}, {\footnotesize\texttt{[PAD]}}, {\footnotesize\texttt{[UNK]}}, {\footnotesize\texttt{[MASK]}}) from mBERT, and reinitialize the remaining embeddings randomly. We freeze all parameters outside the embedding layer. For all pretraining runs, we set the random seed to 42.

\subsection{Code}
\label{a:code}
Our code with usage instructions for fine-tuning, pretraining, data preprocessing, and calculating the tokenizer statistics is available at \href{https://github.com/Adapter-Hub/hgiyt}{https://github.com/Adapter-Hub/hgiyt}. The repository also contains further links to a collection of our new pretrained models and language adapters.

\section{Further Analyses and Discussions}
\label{a:further_analyses}

\subsection{Tokenization Analysis}
\label{a:tokenization_analysis}
In our tokenization analysis in \S\ref{s:tokenizer_analysis} of the main text, we only include the fertility and the proportion of continued words as they are sufficient to illustrate and quantify the differences between tokenizers. In support of the findings in \S\ref{s:tokenizer_analysis} and for completeness, we provide additional tokenization statistics here.

For each tokenizer, Table~\ref{tab:vocab_sizes} lists the respective vocabulary size and the proportion of its vocabulary also contained in mBERT. It shows that the tokenizers scoring lower in fertility (and accordingly performing better) than mBERT are often not adequately covered by mBERT's vocabulary. For instance, only 5.6\% of the AraBERT (\textsc{ar}) vocabulary is covered by mBERT.

\begin{figure}[ht]
    \centering
    \includegraphics[width=\columnwidth]{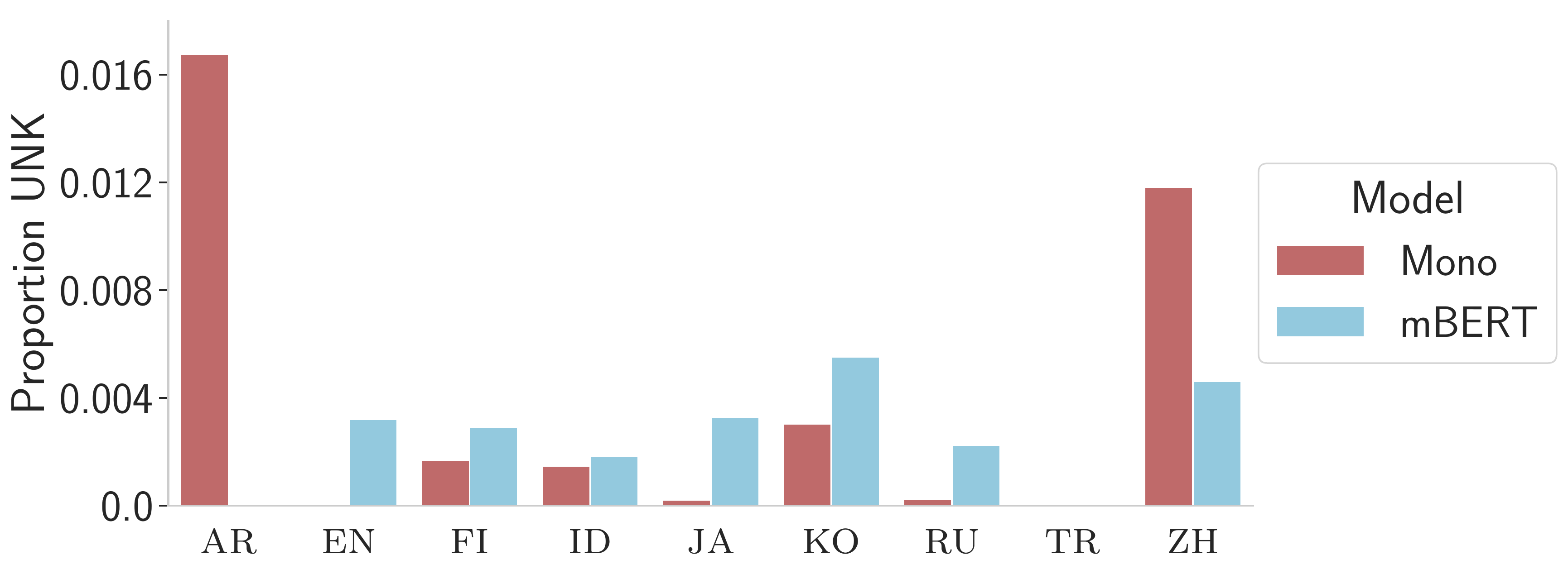}
    \caption{Proportion of unknown tokens in respective monolingual corpora tokenized by monolingual models vs. mBERT.}
    \label{fig:unk_tokens}
\end{figure}

Figure \ref{fig:unk_tokens} compares the proportion of unknown tokens ({\footnotesize\texttt{[UNK]}}) in the tokenized data. It shows that the proportion is generally extremely low, i.e., the tokenizers can typically split unknown words into known subwords. 

\begin{figure}[ht]
    \centering
    \begin{minipage}{\linewidth}
    \centering
    \vspace{-1em}
    \begin{subfigure}[b]{0.32\textwidth}
        \centering
        \includegraphics[width=\textwidth]{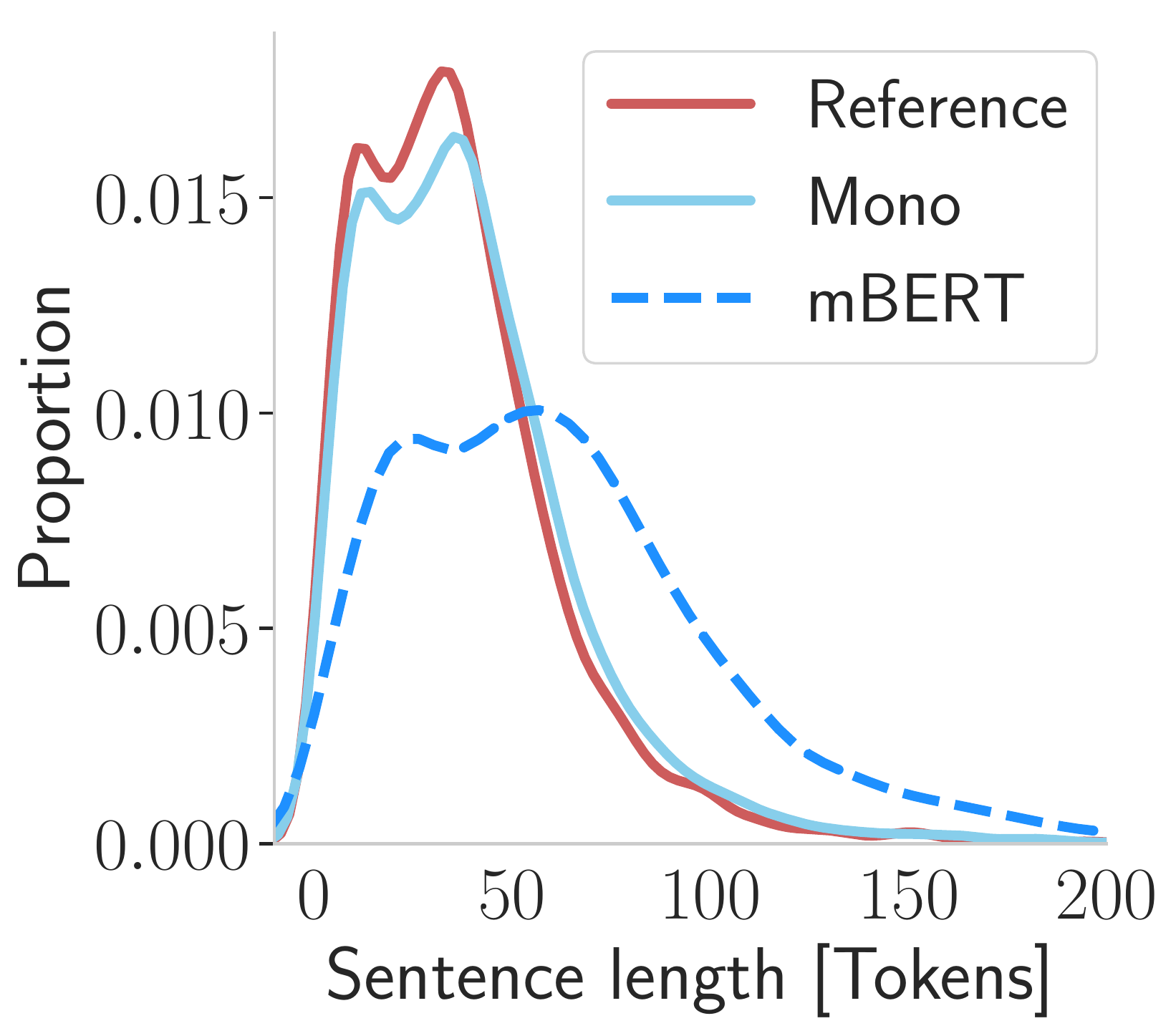}
        \caption{\textsc{ar}}
        \label{fig:ar_sent_len}
    \end{subfigure}
    \begin{subfigure}[b]{0.32\textwidth}
        \centering
        \includegraphics[width=\textwidth]{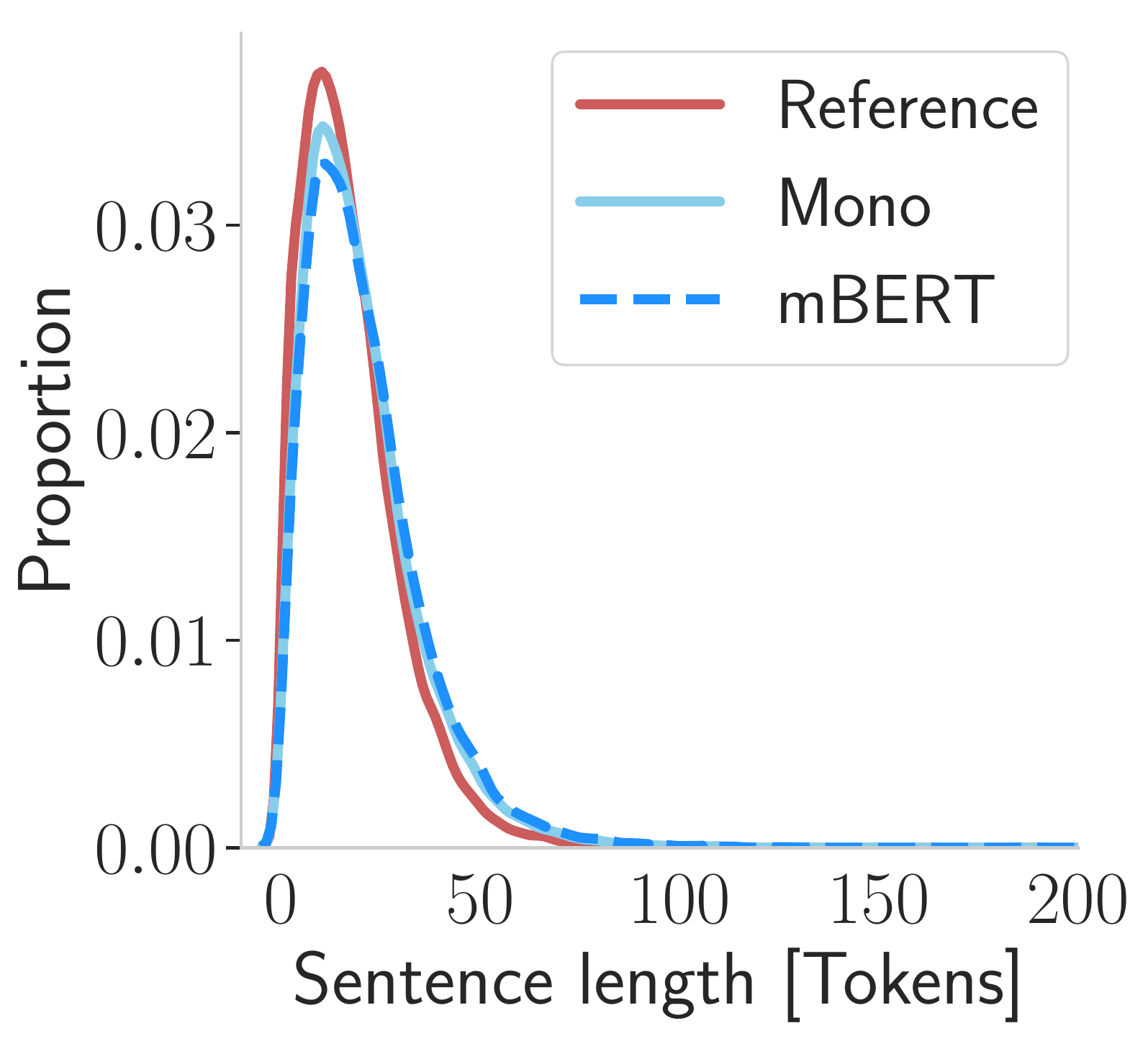}
        \caption{\textsc{en}}
        \label{fig:en_sent_len}
    \end{subfigure}
    \begin{subfigure}[b]{0.32\textwidth}
        \centering
        \includegraphics[width=\textwidth]{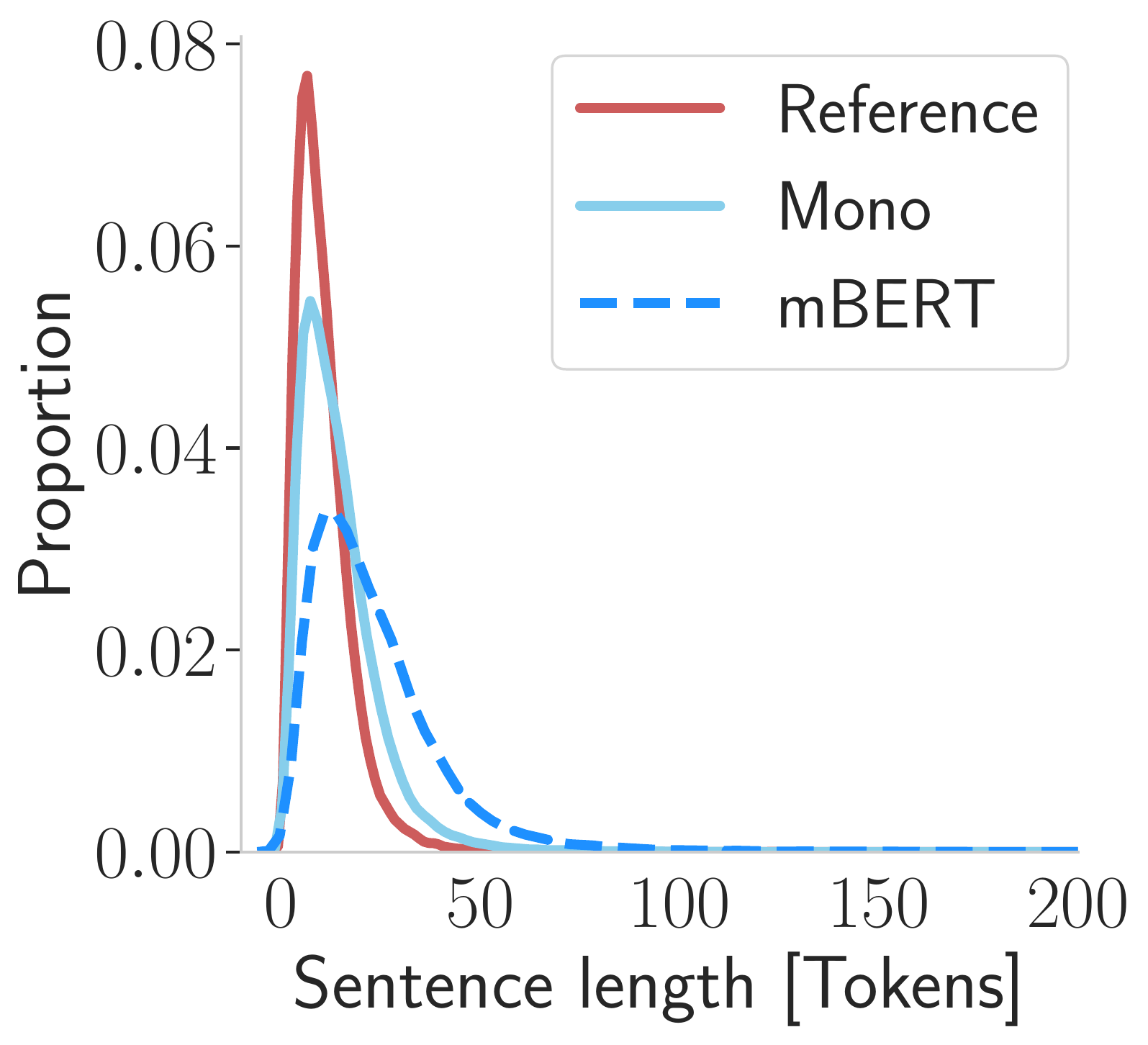}
        \caption{\textsc{fi}}
        \label{fig:fi_sent_len}
    \end{subfigure}
    \begin{subfigure}[b]{0.32\textwidth}
        \centering
        \includegraphics[width=\textwidth]{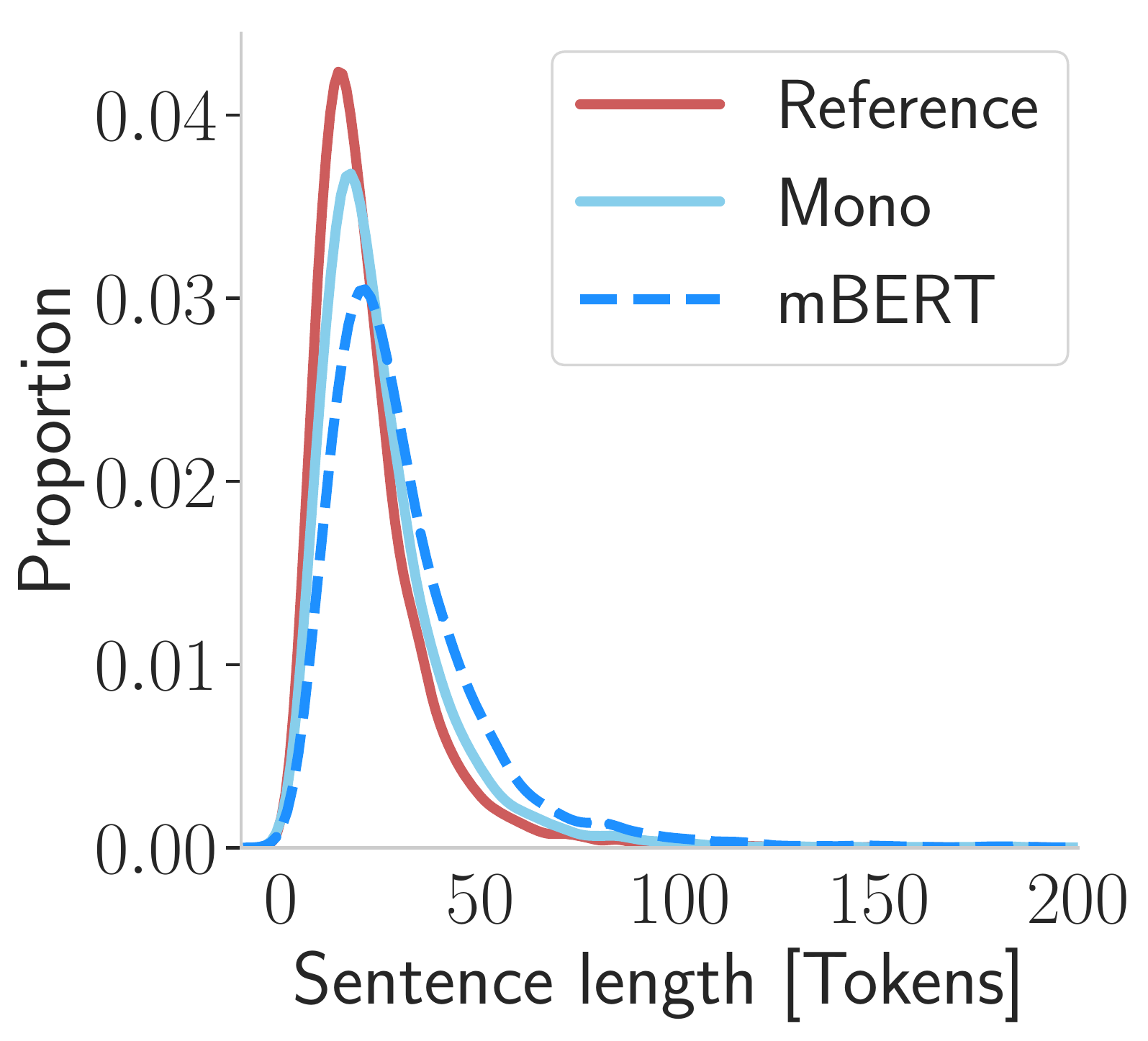}
        \caption{\textsc{id}}
        \label{fig:id_sent_len}
    \end{subfigure}
        \begin{subfigure}[b]{0.32\textwidth}
        \centering
        \includegraphics[width=\textwidth]{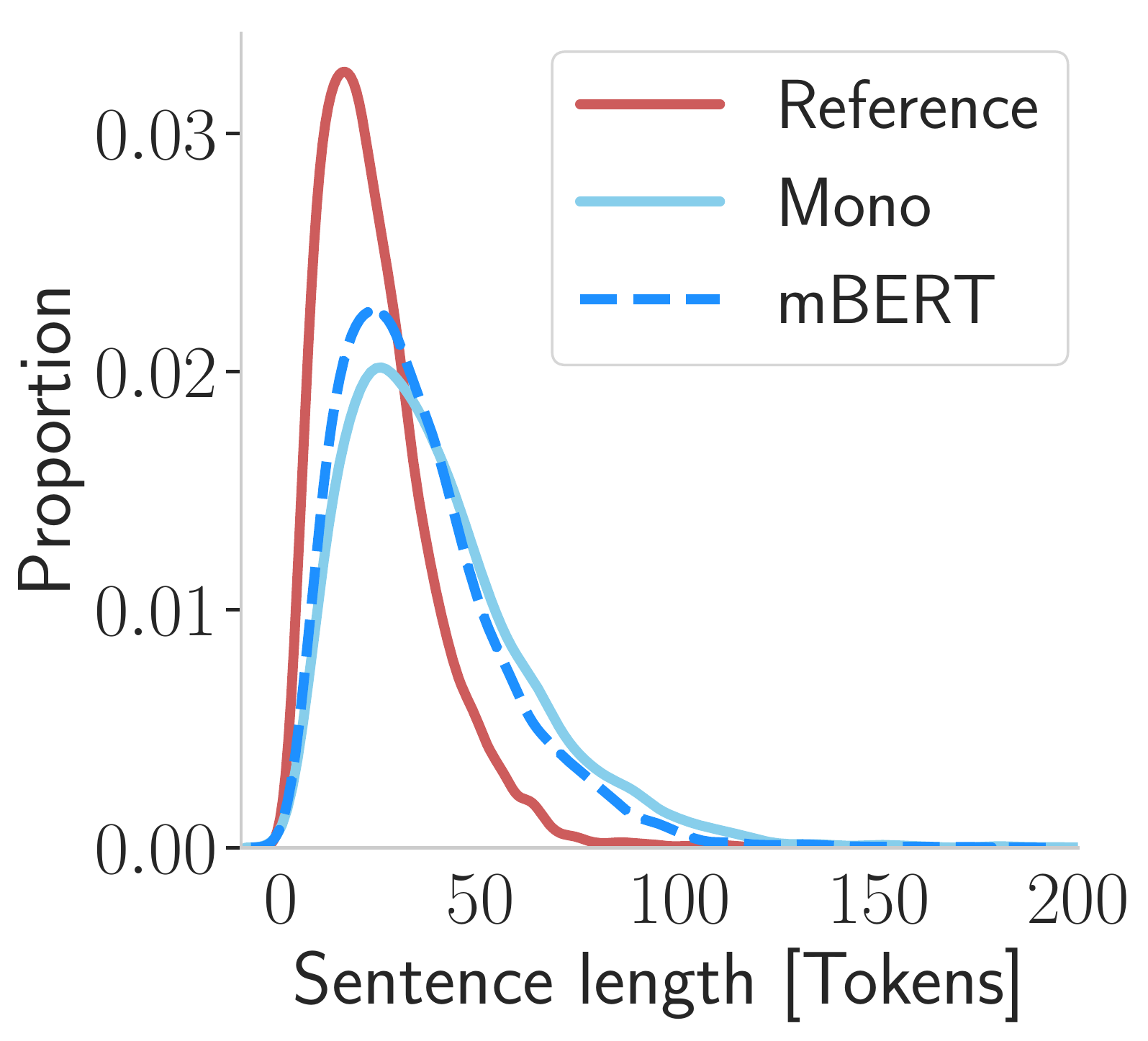}
        \caption{\textsc{ja}}
        \label{fig:ja_sent_len}
    \end{subfigure}
    \begin{subfigure}[b]{0.32\textwidth}
        \centering
        \includegraphics[width=\textwidth]{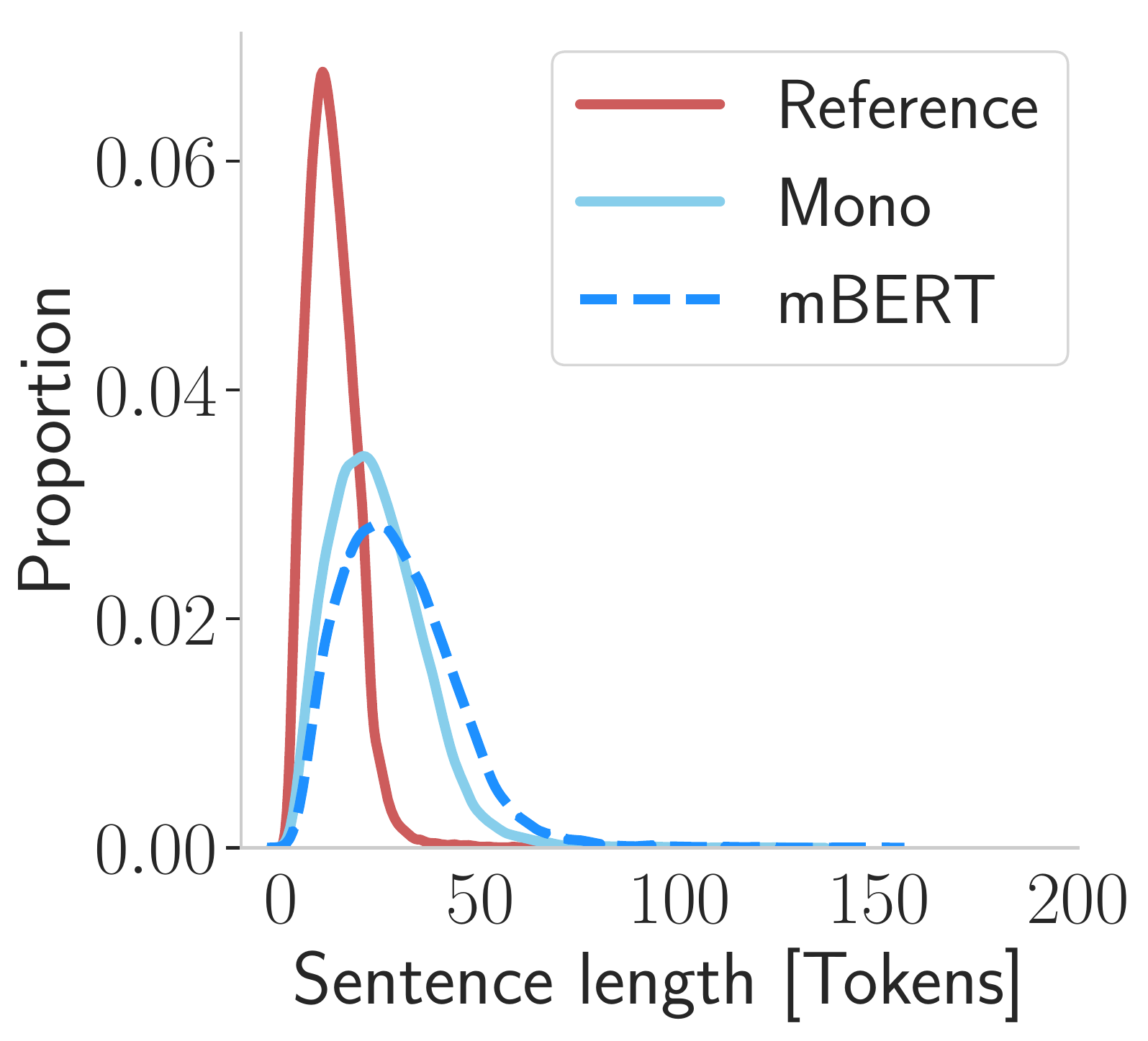}
        \caption{\textsc{ko}}
        \label{fig:ko_sent_len}
    \end{subfigure}
    \begin{subfigure}[b]{0.32\textwidth}
        \centering
        \includegraphics[width=\textwidth]{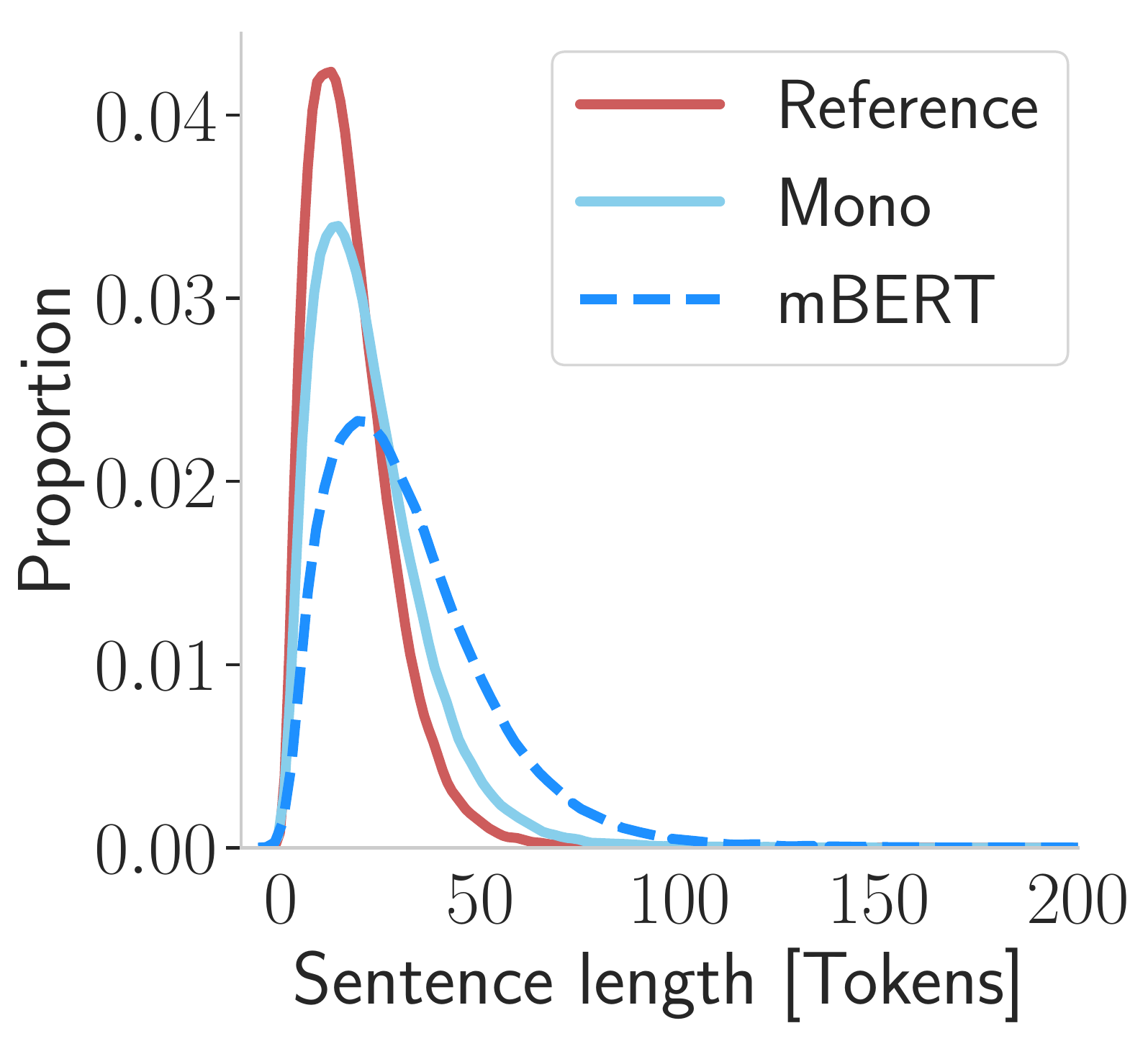}
        \caption{\textsc{ru}}
        \label{fig:ru_sent_len}
    \end{subfigure}
    \begin{subfigure}[b]{0.32\textwidth}
        \centering
        \includegraphics[width=\textwidth]{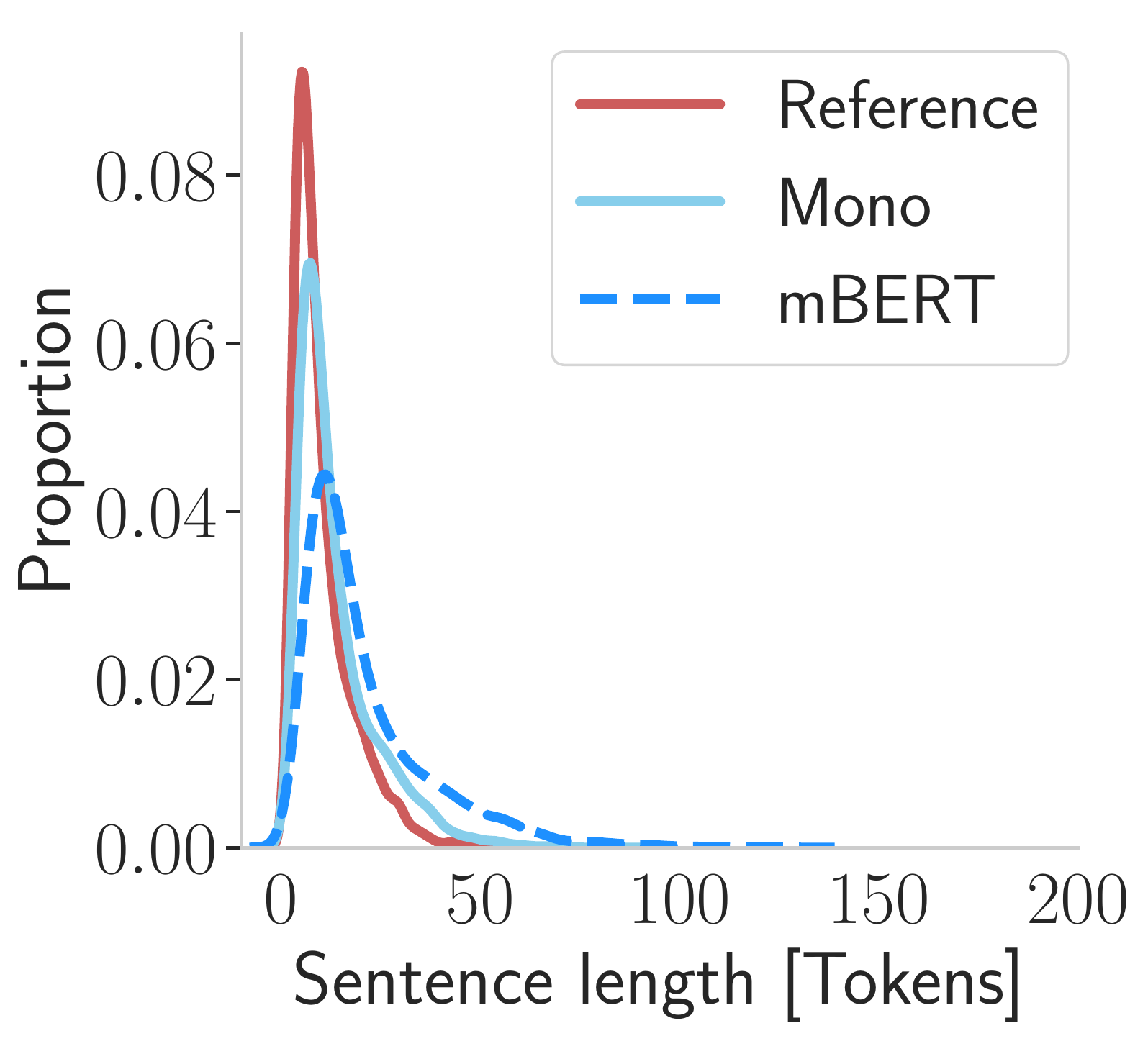}
        \caption{\textsc{tr}}
        \label{fig:tr_sent_len}
    \end{subfigure}
    \begin{subfigure}[b]{0.32\textwidth}
        \centering
        \includegraphics[width=\textwidth]{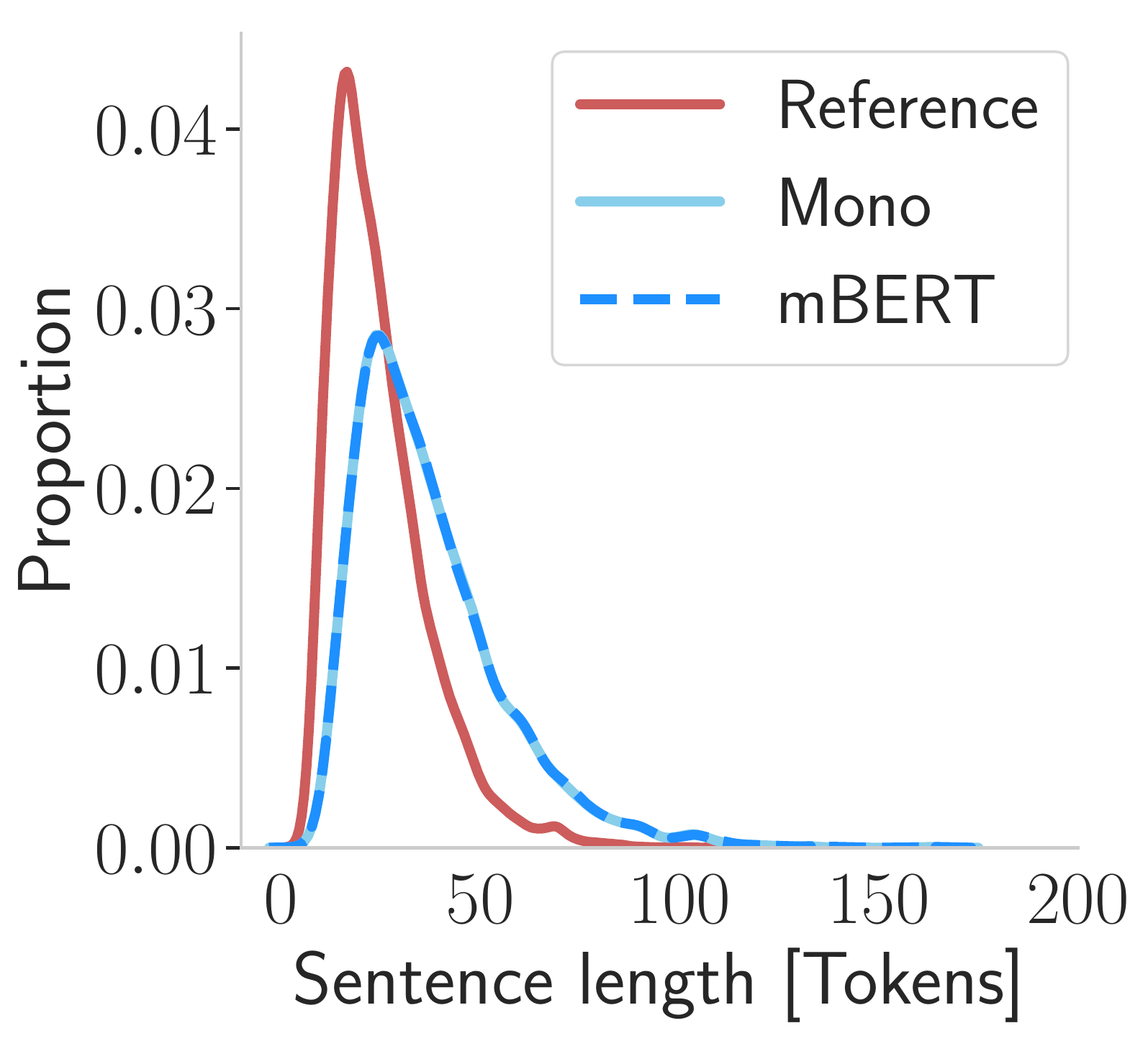}
        \caption{\textsc{zh}}
        \label{fig:zh_sent_len}
    \end{subfigure}
    \vspace{1em}
    \caption{Sentence length distributions of monolingual UD corpora tokenized by respective monolingual BERT models and mBERT, compared to the reference tokenizations by human UD treebank annotators.}
    \label{fig:sentence_distributions1}
    \end{minipage}
    \end{figure}

Similar to the work by \citet{acs:2019}, Figure~\ref{fig:sentence_distributions1} compares the tokenizations produced by the monolingual models and mBERT with the reference tokenizations provided by the human dataset annotators with respect to their sentence lengths. We find that the tokenizers scoring low in fertility and the proportion of continued words typically exhibit sentence length distributions much closer to the reference tokenizations by human UD annotators, indicating they are more capable than the mBERT tokenizer. Likewise, the monolingual models' and mBERT's sentence length distributions are closer for languages with similar fertility and proportion of continued words, such as \textsc{en}, \textsc{ja}, and \textsc{zh}. 

\subsection{Correlation Analysis}
\label{sec:app:correlation_analysis}
To uncover some of the hidden patterns in our results (Tables~\ref{tab:results_table_test}, \ref{tab:new_models_results}, \ref{tab:adapter_results_table_test}), we perform a statistical analysis assessing the correlation between the individual factors (pretraining data size, subword fertility, proportion of continued words) and the downstream performance. 

\begin{figure}
    \centering
    \begin{subfigure}[b]{\linewidth}
        \centering
        \includegraphics[width=\linewidth]{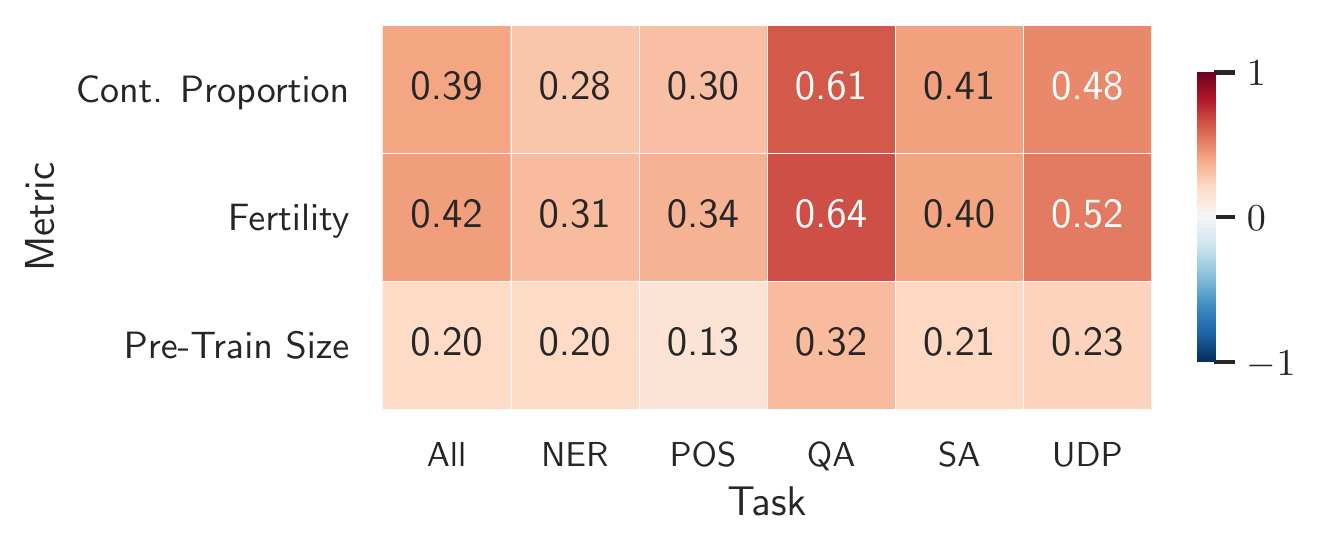}
        \caption{We consider all languages and models.}
        \label{fig:all_models_correlation}
    \end{subfigure}
    \begin{subfigure}[b]{\linewidth}
        \centering
        \includegraphics[width=\linewidth]{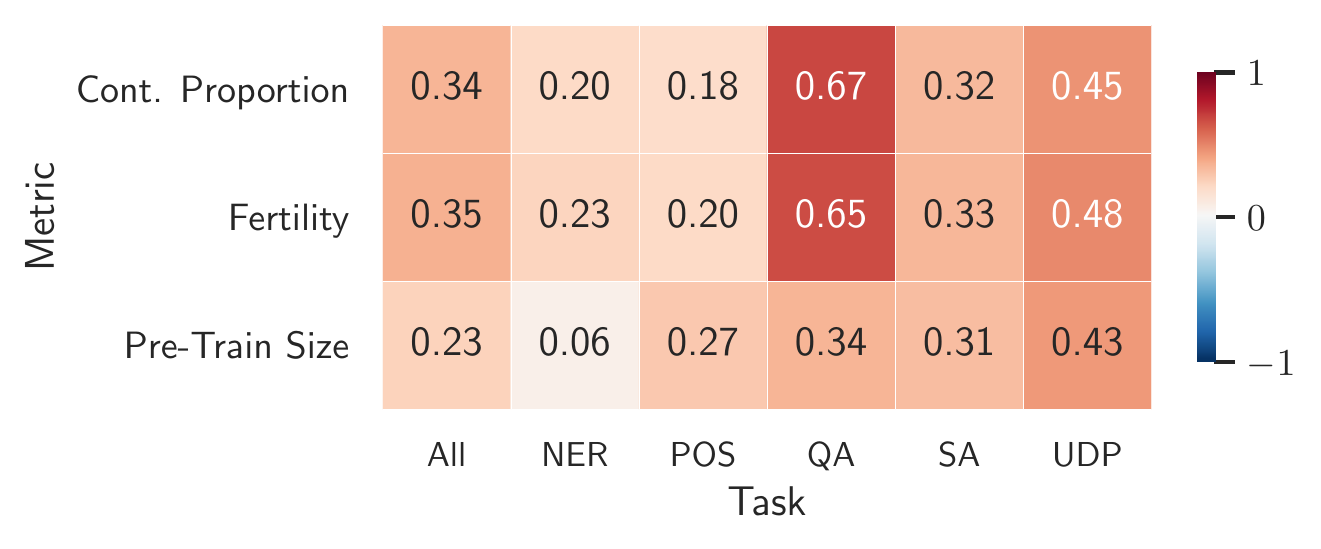}
        \caption{For the proportion of continued words and the fertility, we consider fully fine-tuned mBERT, the {\footnotesize\textsc{MonoModel-*}} models, and the {\footnotesize\textsc{mBERTModel-*}} models. For the pretraining corpus size, we consider the original monolingual models and the {\footnotesize\textsc{MonoModel-MonoTok}} models.}
        \label{fig:new_models_corr_w_id}
    \end{subfigure}
    \caption{Spearman's $\rho$ correlation of a relative decrease in the proportion of continued words (Cont. Proportion), a relative decrease in fertility, and a relative increase in pretraining corpus size with a relative increase in downstream performance over fully fine-tuned mBERT.}
    \label{fig:new_models_correlation}
\end{figure}

Figure \ref{fig:new_models_corr_w_id} shows that both decreases in the proportion of continued words and the fertility correlate with an increase in downstream performance relative to fully fine-tuned mBERT across all tasks. The correlation is stronger for UDP and QA, where we found models with monolingual tokenizers to outperform their counterparts with the mBERT tokenizer consistently. The correlation is weaker for NER and POS tagging, which is also expected, considering the inconsistency of the results.

Somewhat surprisingly, the tokenizer metrics seem to be more indicative of high downstream performance than the size of the pretraining corpus. We believe that this in parts due to the overall poor performance of the uncased IndoBERT model, which we (in this case unfairly) compare to our cased {\footnotesize\textsc{id-MonoModel-MonoTok}} model.
Therefore, we plot the same correlation matrix excluding \textsc{id} in Figure~\ref{fig:new_models_correlation_no_id}.

Compared to Figure~\ref{fig:new_models_corr_w_id}, the overall correlations for the proportion of continued words and the fertility remain mostly unaffected. In contrast, the correlation for the pretraining corpus size becomes much stronger, confirming that the subpar performance of IndoBERT is indeed an outlier in this scenario. Leaving out Indonesian also strengthens the indication that the performance in POS tagging correlates more with the data size than with the tokenizer, although we argue that this indication may be misleading. The performance gap is generally very minor in POS tagging. Therefore, the Spearman correlation coefficient, which only takes the rank into account, but not the absolute score differences, is particularly sensitive to changes in POS tagging performance.

Finally, we plot the correlation between the three metrics and the downstream performance under consideration of all languages and models, including the adapter-based fine-tuning settings, to gain an understanding of how pronounced their effects are in a more ``noisy'' setting.

As Figure~\ref{fig:all_models_correlation} shows, the three factors still correlate with the downstream performance in a similar manner even when not isolated. This correlation tells us that even when there may be other factors that could have an influence, these three factors are still highly indicative of the downstream performance. 

We also see that the correlation coefficients for the proportion of continued words and the fertility are nearly identical, which is expected based on the visual similarity of the respective plots (seen in Figures~\ref{fig:fertility} and \ref{fig:continuation}).

\newpage
\section{Full Results}
For compactness, we have only reported the performance of our models on the respective test datasets in the main text.\footnote{Except for QA, where we do not use any test data} For completeness, we also include the full tables, including development (dev) dataset performance averaged over three random initializations, as described in \S\ref{sec:experimental_setup}. Table~\ref{tab:full_results} shows the full results corresponding to Table~\ref{tab:results_table_test} (initial results), Table~\ref{tab:full_results_new_models} shows the full results corresponding to Table~\ref{tab:new_models_results} (results for our new models), and Table~\ref{tab:full_results_adapters} shows the full results corresponding to Table~\ref{tab:adapter_results_table_test} (adapter-based training).

\begin{table}[ht]
\centering
\resizebox{\columnwidth}{!}{%
\begin{tabular}{lllcc}
\toprule
\textbf{Lang} & \textbf{Model} & \textbf{Reference} & \textbf{V. Size} & \textbf{\% Voc} \\
\midrule
\textsc{multi} & bert-base-multilingual-cased & \citet{devlin:2019} & 119547 & 100 \\
\midrule
\textsc{ar} & aubmindlab/bert-base-arabertv01 & \citet{antoun:2020} & 64000 & 5.6 \\
\textsc{en} & bert-base-cased & \citet{devlin:2019} & 28996 & 66.4 \\
\textsc{fi} & TurkuNLP/bert-base-finnish-cased-v1 & \citet{virtanen:2019} & 50105 & 14.3 \\
\textsc{id} & indobenchmark/indobert-base-p2 & \citet{wilie-etal-2020-indonlu} & 30521 & 40.5 \\
\textsc{ja} & cl-tohoku/bert-base-japanese-char & \textsuperscript{\ref{fn:jbert}} & 4000 & 99.1 \\
\textsc{ko} & snunlp/KR-BERT-char16424 & \citet{lee2020krbert} & 16424 & 47.4 \\
\textsc{ru} & DeepPavlov/rubert-base-cased & \citet{kuratov:2019} & 119547 & 21.1 \\
\textsc{tr} & dbmdz/bert-base-turkish-cased & \citet{schweter:2020} & 32000 & 23.0 \\
\textsc{zh} & bert-base-chinese & \citet{devlin:2019} & 21128 & 79.4 \\
\bottomrule
\end{tabular}
}
\caption{Selection of pretrained models used in our experiments. We display the respective vocabulary sizes and the proportion of tokens that are also covered by mBERT's vocabulary.}
\label{tab:vocab_sizes}
\end{table}

\begin{table}[htp]
\centering
\resizebox{0.5\columnwidth}{!}{%
\begin{tabular}{@{}lll@{}}
\toprule
\textbf{\textbf{Lang}} & \multicolumn{1}{l}{\textbf{\textbf{Treebank}}} & \textbf{\textbf{\# Words}} \\ \midrule
\textsc{ar} & PADT & 254192 \\
\textsc{en} & LinES, EWT, GUM, ParTUT & 449977 \\
\textsc{fi} & FTB, TDT & 324680 \\
\textsc{id} & GSD & 110141 \\
\textsc{ja} & GSD & 179571 \\
\textsc{ko} & GSD & 390369 \\
\textsc{ru} & GSD, SynTagRus, Taiga & 1130482 \\
\textsc{tr} & IMST & 47830 \\
\textsc{zh} & GSD, GSDSimp & 222558 \\ \bottomrule
\end{tabular}%
}
\caption{UD v2.6 \cite{zeman:2020} treebanks used for our tokenizer analyses. We use training and development portions only and display the total number of words per language.}
\label{tab:ud_data_tokenizers}

\end{table}

\begin{table}[ht]
\centering
\resizebox{0.97\columnwidth}{!}{%
\begin{tabular}{lllll}
\toprule
\multicolumn{1}{c}{\textbf{Task}} & \multicolumn{1}{c}{\textbf{Lang}} & \textbf{Dataset} & \textbf{Reference} & \textbf{Train / Dev / Test} \\ 
\midrule
\multirow{9}{*}{\textsc{NER}} & \textsc{ar} & WikiAnn & \citet{pan:2017, rahimi:2019} & 20000 / 10000 / 10000 \\
 & \textsc{en} & CoNLL-2003 & \citet{sang:2003} & 14041 / 3250 / 3453 \\ 
 & \textsc{fi} & FiNER & \citet{ruokolainen:2019} & 13497 / 986 / 3512 \\
 & \textsc{id} & WikiAnn & \citet{pan:2017, rahimi:2019} & 20000 / 10000 / 10000 \\
 & \textsc{ja} & WikiAnn & \citet{pan:2017, rahimi:2019} & 20202 / 10100 / 10113 \\
 & \textsc{ko} & KMOU NER & \textsuperscript{\ref{fn:kmou}} & 23056 / 468 / 463 \\ 
 & \textsc{ru} & WikiAnn & \citet{pan:2017, rahimi:2019} & 20000 / 10000 / 10000 \\ 
 & \textsc{tr} & WikiAnn & \citet{pan:2017, rahimi:2019} & 20000 / 10000 / 10000 \\ 
 & \textsc{zh} & Chinese Literature & \citet{xu:2017} & 24270 / 1902 / 2844 \\ 
 \midrule
\multirow{9}{*}{\textsc{SA}} &  \textsc{ar} & HARD & \citet{elnagar:2018} & 84558 / 10570 / 10570 \\
 & \textsc{en} & IMDb Movie Reviews & \citet{maas:2011} & 20000 / 5000 / 25000 \\
 & \textsc{fi} & --- & --- & --- \\
 & \textsc{id} & Indonesian Prosa & \citet{purwarianti:2019} & 6853 / 763 / 409 \\  
 & \textsc{ja} & Yahoo Movie Reviews & \textsuperscript{\ref{fn:yahoo}} & 30545 / 3818 / 3819 \\
 & \textsc{ko} & NSMC & \textsuperscript{\ref{note:nsmc}} & 120000 / 30000 / 50000 \\
 & \textsc{ru} & RuReviews & \citet{smetanin:2019} & 48000 / 6000 / 6000 \\
 & \textsc{tr} & Movie \& Product Reviews & \citet{demirtas:2013} & 13009 / 1627 / 1629 \\
 & \textsc{zh} & ChnSentiCorp & \textsuperscript{\ref{note:chnsenticorp}} & 9600 / 1200 / 1200 \\
 \midrule
\multirow{9}{*}{\textsc{QA}}& \textsc{ar} & TyDiQA-GoldP & \citet{clark:2020} & 14805 / 921 \\
& \textsc{en} & SQuAD v1.1 & \citet{rajpurkar:2016} & 87599 / 10570 \\
& \textsc{fi} & TyDiQA-GoldP & \citet{clark:2020} & 6855 / 782 \\
& \textsc{id} & TyDiQA-GoldP & \citet{clark:2020} & 5702 / 565 \\
& \textsc{ja} & \multicolumn{1}{l}{---} & \multicolumn{1}{l}{---} & \multicolumn{1}{l}{---} \\
& \textsc{ko} & KorQuAD 1.0 & \citet{lim:2019} & 60407 / 5774 \\
& \textsc{ru} & SberQuAD & \citet{efimov:2020} & 45328 / 5036 \\
& \textsc{tr} & TQuAD & \textsuperscript{\ref{fn:tquad}} & 8308 / 892 \\
& \textsc{zh} & DRCD & \citet{shao:2019} & 26936 / 3524 \\
 \midrule
\multirow{9}{*}{\textsc{UD}} & \textsc{ar} & PADT & \cite{zeman:2020} & 6075 / 909 / 680 \\
 & \textsc{en} & EWT & \cite{zeman:2020} & 12543 / 2002 / 2077 \\
 & \textsc{fi} & FTB & \cite{zeman:2020} & 14981 / 1875 / 1867 \\
 & \textsc{id} & GSD & \cite{zeman:2020} & 4477 / 559 / 557 \\
 & \textsc{ja} & GSD & \cite{zeman:2020} & 7027 / 501 /  543 \\
 & \textsc{ko} & GSD & \cite{zeman:2020} & 4400 / 950 / 989 \\
 & \textsc{ru} & GSD & \cite{zeman:2020} & 3850 / 579 / 601 \\
 & \textsc{tr} & IMST& \cite{zeman:2020} & 3664 / 988 / 983 \\
 & \textsc{zh} & GSD & \cite{zeman:2020} & 3997 / 500 / 500 \\
\bottomrule
\end{tabular}%
}
\caption{Named entity recognition (NER), sentiment analysis (SA), question answering (QA), and universal dependencies (UD) datasets used in our experiments and the number of examples in their respective training, development, and test portions. 
UD datasets were used for both universal dependency parsing and POS tagging experiments.}
\label{tab:ner_data}
\end{table}


\begin{table}[ht]
\resizebox{\columnwidth}{!}{%
\begin{tabular}{@{}lllcccccccc@{}}
\toprule
\multirow{3}{*}{\textbf{Lg}} & \multirow{3}{*}{\textbf{Model}}                  & \multicolumn{2}{c}{\textbf{NER}} &   \multicolumn{2}{c}{\textbf{SA}} &   \textbf{QA}                   &   \multicolumn{2}{c}{\textbf{UDP}}                                & \multicolumn{2}{c}{\textbf{POS}} \\
                    &                                 & Dev             & Test           & Dev            & Test           & Dev                           & Dev                           & Test                          & Dev             & Test           \\
                    &                                 & $F_1$           & $F_1$          & Acc            & Acc            & EM / $F_1$                    & UAS / LAS                     & UAS / LAS                     & Acc             & Acc            \\ \midrule
\multirow{2}{*}{\textsc{ar}} & Monolingual                     & \textbf{91.5}   & \textbf{91.1}  & \textbf{96.1}  & \textbf{95.9}  & \textbf{68.3} / \textbf{82.4} & \textbf{89.4} / \textbf{85.0} & \textbf{90.1} / \textbf{85.6} & \textbf{97.5}   & \textbf{96.8}           \\
                    & mBERT                           & 90.3            & 90.0           & 95.8           & 95.4           & 66.1 / 80.6                   & 87.8 / 83.0                   & 88.8 / 83.8                   & 97.2            & \textbf{96.8}           \\ \midrule
\multirow{2}{*}{\textsc{en}} & Monolingual                     & 95.4            & \textbf{91.5}  & \textbf{91.6}  & \textbf{91.6}  & 80.5 / 88.0                   & \textbf{92.6} / \textbf{90.3} & \textbf{92.1} / \textbf{89.7} & \textbf{97.1}   & \textbf{97.0}  \\
                    & mBERT                           & \textbf{95.7}   & 91.2           & 90.1           & 89.8           & \textbf{80.9} / \textbf{88.4} & 92.1 / 89.6                   & 91.6 / 89.1                   & 97.0            & 96.9           \\ \midrule
\multirow{2}{*}{\textsc{fi}} & Monolingual                     & \textbf{93.3}   & \textbf{92.0}  & -----          & -----          & \textbf{69.9} / \textbf{81.6} & \textbf{95.7} / \textbf{93.9} & \textbf{95.9} / \textbf{94.4} & \textbf{98.1}   & \textbf{98.4}  \\
                    & mBERT                           & 90.9            & 88.2           & -----          & -----          & 66.6 / 77.6                   & 91.1 / 88.0                   & 91.9 / 88.7                   & 96.0            & 96.2           \\ \midrule
\multirow{2}{*}{\textsc{id}} & Monolingual                     & 90.9            & 91.0           & \textbf{94.6}  & \textbf{96.0}  & 66.8 / 78.1                   & 84.5 / 77.4                   & 85.3 / 78.1                   & 92.0            & 92.1           \\
                    & mBERT                           & \textbf{93.7}   & \textbf{93.5}  & 93.1           & 91.4           & \textbf{71.2} / \textbf{82.1}          & \textbf{85.0} / \textbf{78.4} & \textbf{85.9} / \textbf{79.3} & \textbf{93.3}            & \textbf{93.5}  \\ \midrule
\multirow{2}{*}{\textsc{ja}} & Monolingual                     & 72.1            & 72.4           & 88.7           & \textbf{88.0}           & ----- / -----                 & \textbf{96.0} / \textbf{94.7} & \textbf{94.7} / \textbf{93.0} & \textbf{98.3}   & \textbf{98.1}  \\
                    & mBERT                           & \textbf{73.4}   & \textbf{73.4}  & \textbf{88.8}           & 87.8           & ----- / -----                 & 95.5 / 94.2                   & 94.0 / 92.3                   & 98.1            & 97.8           \\ \midrule
\multirow{2}{*}{\textsc{ko}} & Monolingual                     & \textbf{88.6}   & \textbf{88.8}  & \textbf{89.8}  & \textbf{89.7}  & \textbf{74.2} / \textbf{91.1} & \textbf{88.5} / \textbf{85.0} & \textbf{90.3} / \textbf{87.2} & \textbf{96.4}   & \textbf{97.0}  \\
                    & mBERT                           & 87.3            & 86.6           & 86.7           & 86.7           & 69.7 / 89.5                   & 86.9 / 83.2                   & 89.2 / 85.7                   & 95.8            & 96.0           \\ \midrule
\multirow{2}{*}{\textsc{ru}} & Monolingual                     & \textbf{91.9}   & \textbf{91.0}  & \textbf{95.2}           & \textbf{95.2}  & \textbf{64.3} / \textbf{83.7} & \textbf{92.4} / \textbf{90.1} & \textbf{93.1} / \textbf{89.9} & \textbf{98.6}   & \textbf{98.4}  \\
                    & mBERT                           & 90.2            & 90.0           & \textbf{95.2}           & 95.0           & 63.3 / 82.6                   & 91.5 / 88.8                   & 91.9 / 88.5                   & 98.4            & 98.2           \\ \midrule
\multirow{2}{*}{\textsc{tr}} & Monolingual                     & 93.1            & 92.8           & \textbf{89.3}  & \textbf{88.8}  & \textbf{60.6} / \textbf{78.1} & \textbf{78.0} / \textbf{70.9} & \textbf{79.8} / \textbf{73.2} & \textbf{97.0}   & \textbf{96.9}  \\
                    & mBERT                           & \textbf{93.7}   & \textbf{93.8}  & 86.4           & 86.4           & 57.9 / 76.4                   & 72.6 / 65.2                   & 74.5 / 67.4                   & 95.5            & 95.7           \\ \midrule
\multirow{2}{*}{\textsc{zh}} & Monolingual                     & \textbf{77.0}   & \textbf{76.5}  & \textbf{94.8}  & \textbf{95.3}  & \textbf{82.3} / \textbf{89.3}                   & \textbf{88.1} / \textbf{84.9} & \textbf{88.6} / \textbf{85.6} & \textbf{96.6}   & \textbf{97.2}  \\
                    & mBERT                           & 76.0            & 76.1           & 93.1           & 93.8           & 82.0 / \textbf{89.3}                   & 87.1 / 83.7                   & 88.1 / 85.0                   & 96.1            & 96.7           \\
            \midrule \midrule
\multirow{2}{*}{\textsc{avg}} & Monolingual & \textbf{88.2} & \textbf{87.4} & \textbf{92.5} & \textbf{92.4} & \textbf{70.8} / \textbf{84.0} & \textbf{89.5} / \textbf{85.8} & \textbf{90.0} / \textbf{86.3} & \textbf{96.9} & \textbf{96.9} \\
                    & mBERT & 87.9 & 87.0 & 91.2 & 91.0 & 69.7 / 83.3 & 87.7 / 83.8 & 88.4 / 84.4 & 96.4 & 96.4  \\
            \bottomrule
\end{tabular}%
}
\caption{Full Results - Performance on Named Entity Recognition (NER), Sentiment Analysis (SA), Question Answering (QA), Universal Dependency Parsing (UDP), and Part-of-Speech Tagging (POS). We use development (dev) sets only for QA. Finnish (\textsc{fi}) SA and Japanese (\textsc{ja}) QA lack respective datasets.}
\label{tab:full_results}
\end{table}

\begin{table}[ht]
\centering
\def\arraystretch{0.95}
\resizebox{0.48\textwidth}{!}{%
\begin{tabular}{@{}llllcccccccc@{}}
\toprule
\multirow{3}{*}{\textbf{Lg}} & \multicolumn{2}{l}{\multirow{3}{*}{\textbf{Model}}}                 & \multicolumn{2}{c}{\textbf{NER}} &   \multicolumn{2}{c}{\textbf{SA}} &   \textbf{QA}                   &   \multicolumn{2}{c}{\textbf{UDP}}                                & \multicolumn{2}{c}{\textbf{POS}} \\
                    & &                                 & Dev             & Test           & Dev            & Test           & Dev                           & Dev                           & Test                          & Dev             & Test           \\
                    & &                                 & $F_1$           & $F_1$          & Acc            & Acc            & EM / $F_1$                    & UAS / LAS                     & UAS / LAS                     & Acc             & Acc            \\
\addlinespace[0.4em]\midrule\addlinespace[0.4em]

\multirow{6}{*}{\vspace{-2.5em}\textsc{ar}}  & 

                     \multicolumn{2}{l}{Monolingual} & 91.5   & 91.1  & \textbf{96.1}  & \textbf{95.9}  & \textbf{68.3} / \textbf{82.4} & \textbf{89.4} / \textbf{85.0} & \textbf{90.1} / \textbf{85.6} & \textbf{97.5}   & 96.8           \\ \addlinespace[0.4em]\addlinespace[0.4em]
                     & \multicolumn{2}{l}{\footnotesize\textsc{MonoModel-MonoTok}}  & 88.6 & \underline{\textbf{91.7}} & \underline{96.0} & \underline{95.6}  & \underline{67.7} / \underline{81.6} & \underline{88.4} / \underline{83.7} & \underline{89.2} / \underline{84.4} &  97.3 & 96.6          \\
                     & \multicolumn{2}{l}{\footnotesize\textsc{MonoModel-mBERTTok}} & \underline{90.1} & 90.0          & 95.9  & 95.5 & 64.1 / 79.4                   &  87.8 / 83.2  & 88.8 / 84.0                   & \underline{97.4} & \underline{\textbf{97.0}} \\ \addlinespace[0.4em]\addlinespace[0.4em]
                     & \multicolumn{2}{l}{\footnotesize\textsc{mBERTModel-MonoTok}} & \underline{\textbf{91.9}} & \underline{91.2} & \underline{95.9} & 95.4  & \underline{66.9} / \underline{81.8} & \underline{88.2} / \underline{83.5} & \underline{89.3} / \underline{84.5} & 97.2 & 96.4          \\
                     & \multicolumn{2}{l}{\footnotesize\textsc{mBERTModel-mBERTTok}} & 90.0 & 89.7          & 95.8 & \underline{95.6} & 66.3 / 80.7                   & 87.8 / 83.0  & 89.1 / 84.2                   & \underline{97.3}  & \underline{96.8} \\ \addlinespace[0.4em]\addlinespace[0.4em]
                     & \multicolumn{2}{l}{mBERT} & 90.3 & 90.0 & 95.8 & 95.4 & 66.1 / 80.6 & 87.8 / 83.0 & 88.8 / 83.8 & 97.2 & 96.8 \\\addlinespace[0.4em]\midrule\addlinespace[0.4em]

\multirow{6}{*}{\vspace{-2.5em}\textsc{fi}}  & 

                     \multicolumn{2}{l}{Monolingual} & \textbf{93.3} & \textbf{92.0} & ----- & ----- & \textbf{69.9} / \textbf{81.6} & \textbf{95.7} / \textbf{93.9} & \textbf{95.9} / \textbf{94.4} & \textbf{98.1} & \textbf{98.4} \\ \addlinespace[0.4em]\addlinespace[0.4em]
                     & \multicolumn{2}{l}{\footnotesize\textsc{MonoModel-MonoTok}} & \underline{91.9}  & 89.1          & -----         & ----- & \underline{66.9} / \underline{79.5} & \underline{93.6} / \underline{91.0} & \underline{93.7} / \underline{91.5} & \underline{97.0} & \underline{97.3} \\
                     & \multicolumn{2}{l}{\footnotesize\textsc{MonoModel-mBERTTok}} & 91.8 & \underline{90.0} &  -----         & ----- & 65.1 / 77.0                   & 93.1 / 90.6 & 93.6 / \underline{91.5}          & 96.2 & 97.0          \\ \addlinespace[0.4em]\addlinespace[0.4em]
                     & \multicolumn{2}{l}{\footnotesize\textsc{mBERTModel-MonoTok}} & 91.0 & \underline{88.1} &  -----         & ----- & \underline{66.4} / \underline{78.3} & \underline{92.2} / \underline{89.3} & \underline{92.4} / \underline{89.6} & 96.3 & 96.6          \\
                     & \multicolumn{2}{l}{\footnotesize\textsc{mBERTModel-mBERTTok}} & \underline{92.0} & \underline{88.1} & -----         & ----- & 65.9 / 77.3                   & 92.1 / 89.2  & 92.2 / 89.4                   & \underline{96.6} & \underline{96.7} \\ \addlinespace[0.4em]\addlinespace[0.4em]
                     & \multicolumn{2}{l}{mBERT} & 90.9 & 88.2 & ----- & ----- & 66.6 / 77.6 & 91.1 / 88.0 & 91.9 / 88.7 & 96.0 & 96.2 \\ \addlinespace[0.4em]\midrule\addlinespace[0.4em]

\multirow{6}{*}{\vspace{-2.5em}\textsc{id}} & 
                     \multicolumn{2}{l}{Monolingual} & 90.9 & 91.0 & \textbf{94.6}  & \textbf{96.0}  & 66.8 / 78.1 & 84.5 / 77.4 & 85.3 / 78.1 & 92.0 & 92.1 \\ \addlinespace[0.4em]\addlinespace[0.4em]
                     & \multicolumn{2}{l}{\footnotesize\textsc{MonoModel-MonoTok}}   & 93.0 & 92.5          & \underline{93.9} & \underline{\textbf{96.0}}  & \underline{73.1} / \underline{83.6} & 83.4 / 76.8 & \underline{85.0} / 78.5          & \underline{\textbf{93.6}} & \underline{\textbf{93.9}} \\
                     & \multicolumn{2}{l}{\footnotesize\textsc{MonoModel-mBERTTok}} & \underline{93.3} & \underline{93.2} & \underline{93.9} & 94.8            & 67.0 / 79.2                   & \underline{84.0} / \underline{77.4} & 84.9 / \underline{78.6}          & 93.4  & 93.6          \\ \addlinespace[0.4em]\addlinespace[0.4em]
                     & \multicolumn{2}{l}{\footnotesize\textsc{mBERTModel-MonoTok}}  & 93.8 & \underline{\textbf{93.9}} & \underline{94.4} & \underline{94.6} & \underline{\textbf{74.1}} / \underline{\textbf{83.8}} &  \underline{\textbf{85.5}} / \underline{\textbf{78.8}} & \underline{\textbf{86.4}} / \underline{\textbf{80.2}} & \underline{93.5} & \underline{93.8} \\
                     & \multicolumn{2}{l}{\footnotesize\textsc{mBERTModel-mBERTTok}} & \underline{\textbf{93.9}} & \underline{\textbf{93.9}} & 93.7 & \underline{94.6}  & 71.9 / 82.7                   & 85.3 / 78.6 & 86.2 / 79.6                   & 93.4 & 93.7          \\ \addlinespace[0.4em]\addlinespace[0.4em]
                     & \multicolumn{2}{l}{mBERT} & 93.7 & 93.5 & 93.1 & 91.4  & 71.2 / 82.1 & 85.0 / 78.4 & 85.9 / 79.3 & 93.3  & 93.5 \\ \addlinespace[0.4em]\midrule\addlinespace[0.4em]

\multirow{6}{*}{\vspace{-2.5em}\textsc{ko}}  &
                     \multicolumn{2}{l}{Monolingual} & \textbf{88.6} & \textbf{88.8} & \textbf{89.8}  & \textbf{89.7}  & \textbf{74.2} / \textbf{91.1} & \textbf{88.5} / \textbf{85.0} & \textbf{90.3} / \textbf{87.2} & \textbf{96.4} & \textbf{97.0} \\ \addlinespace[0.4em]\addlinespace[0.4em]
                     & \multicolumn{2}{l}{\footnotesize\textsc{MonoModel-MonoTok}} & \underline{87.9} & \underline{87.1} & \underline{89.0} & \underline{88.8} & \underline{72.8} / \underline{90.3} & \underline{87.9} / \underline{84.2} & \underline{89.8} / \underline{86.6} & \underline{\textbf{96.4}} & \underline{96.7} \\
                     & \multicolumn{2}{l}{\footnotesize\textsc{MonoModel-mBERTTok}} & 86.9 & 85.8          & 87.3 & 87.2          & 68.9 / 88.7                   & 86.9 / 83.2 & 88.9 / 85.6                   & 96.1  & 96.4          \\ \addlinespace[0.4em]\addlinespace[0.4em]
                     & \multicolumn{2}{l}{\footnotesize\textsc{mBERTModel-MonoTok}} & \underline{87.9} & \underline{86.6} & \underline{88.2} & \underline{88.1} & \underline{72.9} / \underline{90.2} & \underline{87.9} / \underline{83.9} & \underline{90.1} / \underline{87.0} & \underline{96.2}  & \underline{96.5} \\
                     & \multicolumn{2}{l}{\footnotesize\textsc{mBERTModel-mBERTTok}} & 86.7 & 86.2          & 86.6 & 86.6           & 69.3 / 89.3                   & 87.2 / 83.3 & 89.2 / 85.9                   & 95.9 & 96.2          \\ \addlinespace[0.4em]\addlinespace[0.4em]
                     & \multicolumn{2}{l}{mBERT} & 87.3 & 86.6 & 86.7  & 86.7 & 69.7 / 89.5 & 86.9 / 83.2 & 89.2 / 85.7 & 95.8 & 96.0 \\ \addlinespace[0.4em]\midrule\addlinespace[0.4em]

\multirow{6}{*}{\vspace{-2.5em}\textsc{tr}} &  
                     \multicolumn{2}{l}{Monolingual} & 93.1 & 92.8 & \textbf{89.3} & \textbf{88.8} & \textbf{60.6} / \textbf{78.1} & \textbf{78.0} / \textbf{70.9} & \textbf{79.8} / \textbf{73.2} & \textbf{97.0} & \textbf{96.9} \\ \addlinespace[0.4em]\addlinespace[0.4em]
                     & \multicolumn{2}{l}{\footnotesize\textsc{MonoModel-MonoTok}} & \underline{93.5} & \underline{93.4} & \underline{87.5} & \underline{87.0} & \underline{56.2} / \underline{73.7} &  \underline{74.4} / \underline{67.3} & \underline{76.1} / \underline{68.9} & 95.9 & 96.3 \\
                     & \multicolumn{2}{l}{\footnotesize\textsc{MonoModel-mBERTTok}} & 93.2 & 93.3          & 85.8 & 84.8            & 55.3 / 72.5                   & 73.2 / 66.0  & 75.3 / 68.3                   & \underline{96.4}  & \underline{96.5} \\ \addlinespace[0.4em]\addlinespace[0.4em]
                     & \multicolumn{2}{l}{\footnotesize\textsc{mBERTModel-MonoTok}}  & 93.5 & 93.7          & \underline{86.1} & 85.3  & \underline{59.4} / \underline{76.7} & \underline{74.7} / \underline{67.6} & \underline{77.1} / \underline{70.2} & \underline{96.1} & \underline{96.3} \\
                     & \multicolumn{2}{l}{\footnotesize\textsc{mBERTModel-mBERTTok}} & \underline{\textbf{93.9}} & \underline{\textbf{93.8}} & 86.0 & \underline{86.1} & 58.7 / 76.6                   & 73.2 / 66.1 & 76.2 / 69.2                   & 95.9 & \underline{96.3} \\ \addlinespace[0.4em]\addlinespace[0.4em]
                     & \multicolumn{2}{l}{mBERT} & 93.7 & \textbf{93.8} & 86.4 & 86.4 & 57.9 / 76.4 & 72.6 / 65.2 & 74.5 / 67.4 & 95.5 & 95.7 \\ \addlinespace[0.4em]\midrule\midrule\addlinespace[0.4em]

\multirow{6}{*}{\vspace{-2.5em}\textsc{avg}} &
                     \multicolumn{2}{l}{Monolingual} & 91.5 & \textbf{91.1} & \textbf{92.5} & \textbf{92.6} & \textbf{68.0} / \textbf{82.3} & \textbf{87.2} / \textbf{82.4} & \textbf{88.3} / \textbf{83.7} & \textbf{96.2} & \textbf{96.2} \\ \addlinespace[0.4em]\addlinespace[0.4em]
                     & \multicolumn{2}{l}{\footnotesize\textsc{MonoModel-MonoTok}} & 91.0 & \underline{90.8} & \underline{91.6} & \underline{91.9} & \underline{67.3} / \underline{81.7} & \underline{85.5} / \underline{80.6} & \underline{86.8} / \underline{82.0}  & \underline{96.0} & \underline{\textbf{96.2}} \\ 
                     & \multicolumn{2}{l}{\footnotesize\textsc{MonoModel-mBERTTok}}  & \underline{91.1} & 90.5 & 90.7 & 90.6 & 64.1 / 79.4 & 85.0 / 80.1 & 86.3 / 81.6 & 95.9 & 96.1    \\ \addlinespace[0.4em]\addlinespace[0.4em]
                     & \multicolumn{2}{l}{\footnotesize\textsc{mBERTModel-MonoTok}} & \underline{\textbf{91.6}} & \underline{90.7} & \underline{91.2} & \underline{90.9} & \underline{\textbf{68.0}} / \underline{82.2} & \underline{85.7} / \underline{80.6} & \underline{87.1} / \underline{82.3} & \underline{95.9} & \underline{95.9} \\
                     & \multicolumn{2}{l}{\footnotesize\textsc{mBERTModel-mBERTTok}} & 91.3 & 90.3 & 90.5 & 90.7 & 66.4 / 81.3 & 85.1 / 80.0 & 86.6 / 81.7 & 95.8 & \underline{95.9} \\ \addlinespace[0.4em]\addlinespace[0.4em]
                     & \multicolumn{2}{l}{mBERT} & 91.2 & 90.4 & 90.5 & 90.0 & 66.3 / 81.2 & 84.7 / 79.6 & 86.1 / 81.0 & 95.6 & 95.6 \\ \addlinespace[0.4em]\bottomrule

\end{tabular}
}
\vspace{-1mm}
\caption{Full Results - Performance of our new {\footnotesize\textsc{MonoModel-*}} and {\footnotesize\textsc{mBERTModel-*}} models (see \S\ref{sec:app:training_new_models}) fine-tuned for the NER, SA, QA, UDP, and POS tasks (see \S\ref{sec:tasks}), compared to the monolingual models from prior work and fully fine-tuned mBERT. We group model counterparts w.r.t. tokenizer choice to facilitate a direct comparison between respective counterparts. We use development sets only for QA. \textbf{Bold} denotes best score across all models for a given language and task. \underline{Underlined} denotes best score compared to its respective counterpart.}
\label{tab:full_results_new_models}
\vspace{-1.5mm}
\end{table}

\begin{table}[ht]
\resizebox{\columnwidth}{!}{%
\begin{tabular}{@{}lllcccccccc@{}}
\toprule
\multirow{3}{*}{\textbf{Lg}} & \multirow{3}{*}{\textbf{Model}}                  & \multicolumn{2}{c}{\textbf{NER}} &   \multicolumn{2}{c}{\textbf{SA}} &   \textbf{QA}                   &   \multicolumn{2}{c}{\textbf{UDP}}                                & \multicolumn{2}{c}{\textbf{POS}} \\
                    &                                 & Dev             & Test           & Dev            & Test           & Dev                           & Dev                           & Test                          & Dev             & Test           \\
                    &                                 & $F_1$           & $F_1$          & Acc            & Acc            & EM / $F_1$                    & UAS / LAS                     & UAS / LAS                     & Acc             & Acc            \\ \midrule
\multirow{4}{*}{\textsc{ar}} & mBERT                  & 90.3            & 90.0           & 95.8           & 95.4           & 66.1 / 80.6                   & \textbf{87.8} / \textbf{83.0} & \textbf{88.8} / \textbf{83.8} & 97.2            & \textbf{96.8}  \\
                    & \, \, + A\textsuperscript{Task}       & 90.0            & 89.6           & \textbf{96.1}  & 95.6           & 66.7 / 81.1                   & 86.7 / 81.6                   & 87.8 / 82.6                   & \textbf{97.3}    & \textbf{96.8}  \\
                    & \, \, +  A\textsuperscript{Task} + A\textsuperscript{Lang} & 90.2            & 89.7           & \textbf{96.1}  & \textbf{95.7}  & 66.9 / 81.0                   & 87.0 / 81.9                   & 88.0 / 82.8                   & \textbf{97.3}   & \textbf{96.8}  \\
                    & \, \, + A\textsuperscript{Task} + A\textsuperscript{Lang} +  {\footnotesize\textsc{MonoTok}}  & \textbf{91.5}	& \textbf{91.1}  & 96.0			  & \textbf{95.7}  & \textbf{67.7} / \textbf{82.1} & 87.7 / 82.8                   & 88.5 / 83.4				   & \textbf{97.3}	 & 96.5			  \\ \midrule
\multirow{4}{*}{\textsc{fi}} & mBERT                  & 90.9            & 88.2           & -----          & -----          & 66.6 / 77.6                   & 91.1 / 88.0                   & 91.9 / 88.7                   & 96.0            & 96.2           \\
                    & \, \, + A\textsuperscript{Task}       & 91.2            & \textbf{88.5}  & -----          & -----          & 65.2 / 77.3                   & 90.2 / 86.3                   & 90.8 / 87.0                   & 95.8            & 95.7           \\
                    & \, \, +  A\textsuperscript{Task} + A\textsuperscript{Lang} & \textbf{91.6}   & 88.4           & -----          & -----          & 65.7 / 77.1                   & 91.1 / 87.7                   & 91.8 / 88.5                   & 96.3            & 96.6           \\
                    & \, \, + A\textsuperscript{Task} + A\textsuperscript{Lang} +  {\footnotesize\textsc{MonoTok}}  & 90.8			& 88.1			 & -----		  & -----		   & \textbf{66.7} / \textbf{79.0} & \textbf{92.8} / \textbf{89.9} & \textbf{92.8} / \textbf{90.1} & \textbf{96.9}   & \textbf{97.3}  \\ \midrule
\multirow{4}{*}{\textsc{id}} & mBERT                  & \textbf{93.7}   & \textbf{93.5}  & 93.1           & 91.4           & 71.2 / 82.1         		   & \textbf{85.0} / \textbf{78.4} & \textbf{85.9} / \textbf{79.3} & 93.3            & \textbf{93.5}  \\
                    & \, \, + A\textsuperscript{Task}       & 93.3            & \textbf{93.5}  & 92.9           & 90.6           & 70.6 / 82.5		           & 83.7 / 76.5                   & 84.8 / 77.4                   & 93.5            & 93.4           \\
                    & \, \, +  A\textsuperscript{Task} + A\textsuperscript{Lang} & 93.6            & \textbf{93.5}  & 93.1           & 93.6           & 70.8 / 82.2                   & 84.3 / 77.4                   & 85.4 / 78.1                   & 93.6   		 & 93.4           \\
                    & \, \, + A\textsuperscript{Task} + A\textsuperscript{Lang} +  {\footnotesize\textsc{MonoTok}}  & 93.0			& 93.4			 & \textbf{94.5}  & \textbf{93.8}  & \textbf{74.4} / \textbf{84.4} & 84.6 / 77.6				   & 85.1 / 78.3				   & \textbf{93.7}	 & \textbf{93.5}  \\ \midrule
\multirow{4}{*}{\textsc{ko}} & mBERT                  & 87.3            & \textbf{86.6}  & 86.7           & 86.7           & 69.7 / 89.5                   & 86.9 / \textbf{83.2}          & \textbf{89.2} / \textbf{85.7} & 95.8            & 96.0           \\
                    & \, \, + A\textsuperscript{Task}       & 87.1            & 86.2           & 86.7           & 86.5           & 69.8 / 89.7                   & 85.5 / 81.1                   & 87.8 / 83.9                   & 95.9            & 96.2           \\
                    & \, \, +  A\textsuperscript{Task} + A\textsuperscript{Lang} & 87.3            & 86.2           & 86.6           & 86.3           & 70.0 / 89.8                   & 85.9 / 81.6                   & 88.3 / 84.3                   & 96.0            & 96.2           \\
                    & \, \, + A\textsuperscript{Task} + A\textsuperscript{Lang} +  {\footnotesize\textsc{MonoTok}}  & \textbf{87.7}   & 86.5			 & \textbf{87.9}  & \textbf{87.9}  & \textbf{73.1} / \textbf{90.4} & \textbf{87.0} / 82.7          & 88.9 / 85.2				   & \textbf{96.3}   & \textbf{96.5}  \\ \midrule
\multirow{4}{*}{\textsc{tr}} & mBERT                  & \textbf{93.7}   & \textbf{93.8}  & \textbf{86.4}  & \textbf{86.4}  & 57.9 / 76.4                   & 72.6 / 65.2                   & 74.5 / 67.4                   & 95.5            & 95.7           \\
                    & \, \, + A\textsuperscript{Task}       & 93.0            & 93.0           & 86.1           & 83.9           & 55.3 / 75.1                   & 70.4 / 62.0                   & 72.4 / 64.1                   & 95.5            & 95.7           \\
                    & \, \, +  A\textsuperscript{Task} + A\textsuperscript{Lang} & 93.3            & 93.5           & 86.2           & 84.8           & 56.9 / 75.8                   & 71.1 / 63.0                   & 73.0 / 64.7                   & 96.0            & 95.9           \\
                    & \, \, + A\textsuperscript{Task} + A\textsuperscript{Lang} +  {\footnotesize\textsc{MonoTok}}  & 92.7			& 92.7			 & 86.1			  & 85.3		   & \textbf{60.0} / \textbf{77.0} & \textbf{73.5} / \textbf{65.6} & \textbf{75.7} / \textbf{68.1} & \textbf{96.4}   & \textbf{96.3}  \\ \midrule \midrule
                    
\multirow{4}{*}{\textsc{avg}} & mBERT & \textbf{91.2} & \textbf{90.4} & 90.5 & 90.0 & 66.3 / 81.2 & 84.7 / 79.6 &  86.0 / \textbf{81.0} & 95.6 & 95.6 \\
                    & \, \, + A\textsuperscript{Task} & 90.9 & 90.2 & 90.5 & 89.2 & 65.5 / 81.1 & 83.3 / 77.5 & 84.7 / 79.0 & 95.6 & 95.6 \\
                    & \, \, +  A\textsuperscript{Task} + A\textsuperscript{Lang} & \textbf{91.2} & 90.3 & 90.5 & 90.1 & 66.1 / 81.2 & 83.9 / 78.3 & 85.3 / 79.7 & 95.8 & 95.8 \\
                    & \, \, + A\textsuperscript{Task} + A\textsuperscript{Lang} +  {\footnotesize\textsc{MonoTok}}  & 91.1 & \textbf{90.4} & \textbf{91.1} & \textbf{90.7} & \textbf{68.4} / \textbf{82.6} & \textbf{85.1} / \textbf{79.7} & \textbf{86.2} / \textbf{81.0} & \textbf{96.1} & \textbf{96.0} \\ \bottomrule
\end{tabular}%
}
\caption{Full Results - Performance on the different  tasks leveraging mBERT with different adapter components (see \S\ref{sec:adapter_based_training}).}
\label{tab:full_results_adapters}
\end{table}

\end{document}